\pdfoutput=1

\documentclass{article}

\usepackage{bosonai_techreport}

\usepackage[utf8]{inputenc}
\usepackage[T1]{fontenc}
\usepackage{amsmath,amssymb,amsthm}
\usepackage{mathtools}          %
\usepackage{booktabs}
\usepackage{array}
\usepackage{bm}
\usepackage{microtype}
\usepackage{needspace}          %
\usepackage{rotating}           %
\usepackage[boxed]{algorithm}   %
\usepackage{algpseudocode}
\usepackage{tikz}
\usepackage[colorlinks=true,linkcolor=bosonsignal,citecolor=bosonsignal,urlcolor=bosonsignal]{hyperref}


\graphicspath{{./}{figures/}}

\newtheorem{theorem}{Theorem}
\newtheorem{lemma}[theorem]{Lemma}
\newtheorem{proposition}[theorem]{Proposition}
\newtheorem{corollary}[theorem]{Corollary}
\newtheorem{definition}[theorem]{Definition}
\newtheorem{remark}[theorem]{Remark}

\definecolor{color_0}{rgb}{0.2980392156862745,0.4470588235294118,0.6901960784313725}
\definecolor{color_3}{rgb}{0.7686274509803922,0.3058823529411765,0.3215686274509804}
\newcommand{\inner}[2]{\left\langle #1,#2 \right\rangle}
\def\E{\mathbb{E}}

\DeclareMathOperator{\sign}{sign}
\DeclareMathOperator{\Var}{Var}
\DeclareMathOperator{\Cov}{Cov}
\newcommand{\R}{\mathbb{R}}

\newcommand{\ut}{\tilde{u}}
\newcommand{\vt}{\tilde{v}}
\newcommand{\xt}{\tilde{x}}
\newcommand{\Ufam}{\mathcal{U}}
\newcommand{\capset}{S}
\newcommand{\eps}{\varepsilon}
\newcommand{\Cat}{\operatorname{Cat}}
\newcommand{\bal}{\mathbf{1}_{\rm bal}}
\newcommand{\wU}{w_U}
\newcommand{\kU}{k_U}
\newcommand{\wmax}{w_{u\vee v}}
\newcommand{\kmax}{k_{u\vee v}}
\newcommand{\cvec}{\mathbf c}

\newcommand{\signzero}{\sign(0)=+1}

\newcolumntype{L}[1]{>{\raggedright\arraybackslash}p{#1}}
\title{One-Bit Estimators via the Balanced Fast\\
Johnson--Lindenstrauss Transform}

\author{Alexander J.\ Smola}
\affiliation{Boson AI \\ 4677 Old Ironsides Dr \\ Santa Clara, CA 95054}
\email{alex@smola.org}

\renewcommand{\bosonlogo}{boson}

\begin{document}

\maketitle

\begin{abstract}%
Vector databases and near-duplicate detection systems compress massive collections of
high-dimensional vectors so they can be rapidly compared against incoming queries. One-bit
sketches achieve extreme compression by retaining only the signs of random projections, yet
they still allow inner products to be accurately estimated---either by correlating a sign code
against a full-precision query (the asymmetric estimator) or by comparing two sign codes directly
(SimHash). Existing theoretical guarantees for these estimators rely on dense Gaussian projections,
whose computational cost dominates encoding in high dimensions. The fast Johnson--Lindenstrauss
transform (FJLT) offers a much faster alternative: after a randomized Hadamard transform spreads each
vector's energy nearly uniformly across coordinates, a sparse Gaussian matrix with roughly $r$
nonzeros per row can stand in for a dense projection. Taking the signs of an FJLT output is the
natural way to build a fast one-bit sketch, but until now this approach has lacked theoretical
guarantees. We establish nonasymptotic guarantees both for the classical FJLT and for a balanced
variant with exactly $r$ nonzeros per row. Fixing the row support size equalizes the workload across
rows, simplifies vectorized implementation, and reduces the variance of the energy that each row
captures from a vector.

Our analysis rests on a simple geometric observation: conditioned on its nonzero positions, a sparse
Gaussian row is still an exact Gaussian projection, but it views the data through a randomly distorted
metric. Taking signs converts this metric distortion into a systematic bias that cannot be averaged away
by adding more rows. We prove that this bias is bounded by an explicit constant times $w/r$, where the
cap $w$ measures how unevenly a transformed vector distributes its energy across coordinates. Consequently,
$O(\eps^{-2}\log(N/\delta))$ rows with $O(\eps^{-1}\log(nN/\delta))$ nonzeros per row suffice to estimate
$N$ cosine similarities in $\R^n$ to accuracy $\eps$ with probability at least $1-\delta$: quantization to
one bit increases the required row density rather than the sketch length. Matching lower bounds demonstrate
that the $\eps^{-1}$ factor in row sparsity is unavoidable for the asymmetric estimator at small $\eps$.
When no signs are taken, the bias vanishes and our framework recovers the known optimal sparsity of the linear
FJLT with explicit constants.
\end{abstract}

\section{Introduction}\label{sec:intro}

Modern information retrieval and machine learning systems routinely store massive collections of high-dimensional vectors and compare them against incoming queries. Vector databases index embeddings of documents and images, near-duplicate detectors maintain fingerprints of web pages, and large language models cache the attention keys of past tokens. Because memory bandwidth and capacity are scarce in all of these applications, stored vectors must be heavily compressed. The central question is how accurately inner products with new queries can be recovered from the compressed representations.

One-bit sketches are among the simplest and most effective compression schemes. By multiplying a vector by a random Gaussian matrix with $\ell$ rows and retaining only the signs of the resulting entries, a vector is reduced to an $\ell$-bit signature alongside its Euclidean norm. These sign codes are used in two main ways. Symmetric schemes such as SimHash compare two bit strings by the fraction of positions at which they disagree, which directly estimates the angle between the underlying vectors \citep{charikar2002similarity}. Asymmetric schemes quantize only the stored vector to one bit and correlate its sign code against a query kept at higher precision. For instance, QJL compresses attention keys in this manner \citep{zandieh2024qjl}---a primitive subsequently reused by TurboQuant \citep{zandieh2025turboquant}---while RaBitQ scores the sign codes of randomly rotated database vectors against quantized queries \citep{gao2024rabitq}.

Throughout this paper, we label each estimator by the precision kept on either side---one bit (B) or a full-precision Gaussian projection (G)---listing the stored vector first. Thus BB denotes SimHash, BG denotes the asymmetric estimator, and GG denotes the standard linear sketch with full precision on both sides, which serves as our baseline. Because vector norms are stored separately, inner product estimation reduces to estimating the cosine of the angle between two vectors, and we state all results for unit vectors. When constructed from a dense Gaussian matrix, all three estimators require only a small constant multiple of $\eps^{-2}\log(N/\delta)$ rows to estimate the cosines of $N$ vector pairs to within additive error $\eps$ with probability at least $1-\delta$, regardless of the geometry of the data.

The computational bottleneck of dense sketching lies in the projection step itself. Multiplying a vector in $\R^n$ by a dense $\ell\times n$ matrix requires $n\ell$ multiply--add operations, whereas scoring two compressed codes takes only $O(\ell)$ operations. Encoding a single vector is therefore as expensive as performing $n$ pairwise comparisons, and because $\ell$ scales as $\eps^{-2}$, this encoding cost grows rapidly when high accuracy is required. For linear sketches, \citet{ailon2009fast} overcame this bottleneck by introducing the fast Johnson--Lindenstrauss transform (FJLT). The FJLT first flips the signs of the vector's coordinates at random and then applies the normalized Walsh--Hadamard transform, an orthogonal map computable in $n\log_2n$ additions. Together, these two operations spread the energy $\|x\|^2$ of any fixed vector $x$ almost uniformly across all $n$ coordinates. Once no single coordinate dominates, sampling a small number of coordinates at random captures a representative share of the vector's energy, allowing a sparse Gaussian matrix with roughly $r\ll n$ nonzeros per row to stand in for a dense one. The resulting transform $\Phi=PHD$---where $D$ is a diagonal matrix of random signs, $H$ is the normalized Hadamard matrix, and $P$ is a sparse Gaussian matrix---reduces the encoding cost from $n\ell$ to $n\log_2n+r\ell$ operations, saving roughly a factor of $n/r$ for long sketches. We refer to $\ell$ as the \emph{sketch length} and $r$ as the \emph{row sparsity}.

Taking the signs of an FJLT projection is the natural way to accelerate one-bit sketching. Although \citet[Sec.~7]{oymak2015near} observed empirically that signing a moderately sparse Gaussian projection performs nearly as well as signing a dense one, no nonasymptotic theoretical guarantees have been established. There is a fundamental reason for this gap: the Euclidean guarantees of the FJLT rely crucially on linearity, which the sign function destroys. Once linearity is lost, sparse rows become much less forgiving. Indeed, the $\ell_1$ variant of the FJLT analyzed by \citet{ailon2009fast} already incurs an extra factor of $\eps^{-1}$ in row density, for reasons closely related to the phenomenon we examine in Section~\ref{sec:why}.

In this work, we provide the missing theoretical guarantees for fast one-bit sketches. We establish nonasymptotic error bounds with explicit constants for both the asymmetric (BG) and symmetric (BB) estimators, covering both the original FJLT and a balanced variant in which every row has exactly $r$ nonzeros. Crucially, our bounds hold conditionally on the realized preconditioner $D$ and depend on the dataset only through a few easily measurable summary statistics, allowing practitioners to certify the accuracy of a deployed sketch directly on their data. In the worst case, our results reveal a clean trade-off: quantization to one bit costs row sparsity rather than sketch length. Up to a small constant factor, the required sketch length $\ell$ matches that of a dense Gaussian sketch, whereas the required number of nonzeros per row $r$ scales as $1/\eps$. Complementary lower bounds confirm that for small $\eps$, the asymmetric estimator cannot achieve accuracy $\eps$ with $o(1/\eps)$ sparse rows.

\subsection{Why signs are harder}\label{sec:why}

Both the core difficulty of signing sparse projections and the mechanism that resolves it can be understood by examining a single row. Fix the random coordinate signs $D$, and let $\xt=HDx$ denote the preconditioned image of a vector $x$; because $HD$ is orthogonal, two unit vectors $u$ and $v$ with $\inner uv=\rho=\cos\theta$ map to unit vectors $\ut$ and $\vt$ with the same inner product $\rho$ and angle $\theta$. A row of the sparse matrix $P$ is nonzero only on a random subset of coordinates, its \emph{support} $\capset$, where its entries are independent $\mathcal N(0,1/q)$ random variables and $q=r/n$ is the sampling density.

Now condition on the support $\capset$ as well. Absorbing the $q^{-1/2}$ variance scaling into the vector by defining the \emph{subsampled vector} $\xt_\capset\in\R^n$ with coordinates $(\xt_\capset)_j:=q^{-1/2}\xt_j\,\mathbf 1\{j\in\capset\}$, a sparse row acts on $(u,v)$ identically in distribution to a standard isotropic Gaussian vector $g\sim\mathcal N(0,I_n)$ acting on $(\ut_\capset,\vt_\capset)$. Consequently, the scalar row projections $X$ and $Y$ remain jointly centered Gaussian given $\capset$, just as they would be for a dense row, and only their covariance matrix changes: whereas a dense row has covariance $\begin{psmallmatrix}1&\rho\\\rho&1\end{psmallmatrix}$, a sparse row with support $\capset$ has conditional covariance
\[
\Sigma_\capset:=\begin{pmatrix}\sigma_u^2&\sigma_{uv}\\\sigma_{uv}&\sigma_v^2\end{pmatrix}
=\begin{pmatrix}\|\ut_\capset\|^2&\inner{\ut_\capset}{\vt_\capset}\\\inner{\ut_\capset}{\vt_\capset}&\|\vt_\capset\|^2\end{pmatrix}.
\]
Here $\sigma_u^2$ and $\sigma_v^2$ are the rescaled energies that the row \emph{captures} from $u$ and $v$, and $\sigma_{uv}$ is their captured inner product. Over the random draw of $\capset$, all three entries match their dense counterparts in expectation ($\E\sigma_u^2=\E\sigma_v^2=1$ and $\E\sigma_{uv}=\rho$), yet each fluctuates from row to row.

\begin{figure}[t]
\centering
\begin{tikzpicture}[x=2.3cm,y=2.3cm,>=latex,line cap=round,line join=round,
    every node/.style={font=\small,inner sep=1.5pt},
    fm vec/.style={->,line width=0.8pt},
    fm seen/.style={line width=0.5pt,black!70},
    fm ref/.style={line width=0.4pt,black!40,dash pattern=on 2pt off 2pt},
    fm drop/.style={line width=0.4pt,black!60},
    fm sq/.style={line width=0.4pt,black!60},
    fm read/.style={line width=0.6pt,color_0},
    fm dim/.style={|-|,line width=0.4pt,color_0}]

  \begin{scope}
    \node[font=\small,anchor=base] at (0,1.22) {dense row};
    \draw[fm seen] (0,0) circle[radius=1];
    \draw[fm drop] (0.6428,0.7660) -- (0.6428,0);
    \draw[fm sq] (0.5828,0) -- (0.5828,0.06) -- (0.6428,0.06);
    \draw[fm dim] (0,-0.13) -- (0.6428,-0.13)
      node[midway,below=1pt,color_0] {$\rho=\cos\theta$};
    \fill[color_0!14] (0,0) -- (0.32,0) arc[start angle=0,end angle=50,radius=0.32] -- cycle;
    \draw[fm read] (0.32,0) arc[start angle=0,end angle=50,radius=0.32];
    \node[color_0] at (25:0.45) {$\theta$};
    \draw[fm vec] (0,0) -- (1,0);
    \draw[fm vec] (0,0) -- (0.6428,0.7660);
    \fill (0,0) circle[radius=0.9pt];
    \node[anchor=south west] at (1.03,0) {$\ut$};
    \node[anchor=north west] at (1.02,-0.02) {$\|\ut\|=1$};
    \node[anchor=south west] at (0.655,0.78) {$\vt$};
  \end{scope}

  \begin{scope}[xshift=7.3cm]
    \node[font=\small,anchor=base] at (0,1.22) {sparse row with support $\capset$};
    \draw[fm ref] (0,0) circle[radius=1];
    \draw[fm seen,rotate=12.3872] (0,0) ellipse[x radius=1.2435,y radius=0.7705];
    \draw[fm drop] (0.9629,0.6116) -- (0.9629,0);
    \draw[fm sq] (0.9029,0) -- (0.9029,0.06) -- (0.9629,0.06);
    \draw[fm dim] (0,-0.13) -- (0.9629,-0.13)
      node[midway,below=1pt,color_0] {$\zeta_\capset=\sigma_{uv}/\sigma_u$};
    \fill[color_0!14] (0,0) -- (0.32,0) arc[start angle=0,end angle=32.425,radius=0.32] -- cycle;
    \draw[fm read] (0.32,0) arc[start angle=0,end angle=32.425,radius=0.32];
    \node[color_0] at (16.2:0.5) {$\theta_\capset$};
    \draw[fm vec] (0,0) -- (1.2,0);
    \draw[fm vec] (0,0) -- (0.9629,0.6116);
    \fill (0,0) circle[radius=0.9pt];
    \node[anchor=south west] at (1.24,0.0) {$\ut_\capset$};
    \node[anchor=north west] at (1.22,-0.02) {$\|\ut_\capset\|=\sigma_u$};
    \node[anchor=south west] at (0.975,0.625) {$\vt_\capset$};
  \end{scope}
\end{tikzpicture}%
\caption{Conditioned on its support $\capset$, a sparse row views the pair $(u,v)$ through the subsampled pair $(\ut_\capset,\vt_\capset)$; among the three geometric features read by the estimators, only the inner product is unbiased across supports. Left: a dense row sees $(\ut,\vt)$ on the Euclidean unit circle. Right: given $\capset$, coordinate subsampling deforms the unit circle of $\mathrm{span}\{u,v\}$ (dashed) into an ellipse. The linear estimator GG reads $\sigma_{uv}=\inner{\ut_\capset}{\vt_\capset}$, the asymmetric estimator BG reads the scalar projection $\zeta_\capset=\sigma_{uv}/\sigma_u$ of $\vt_\capset$ onto the direction of $\ut_\capset$, and SimHash (BB) reads the angle $\theta_\capset$.}
\label{fig:metric}
\end{figure}
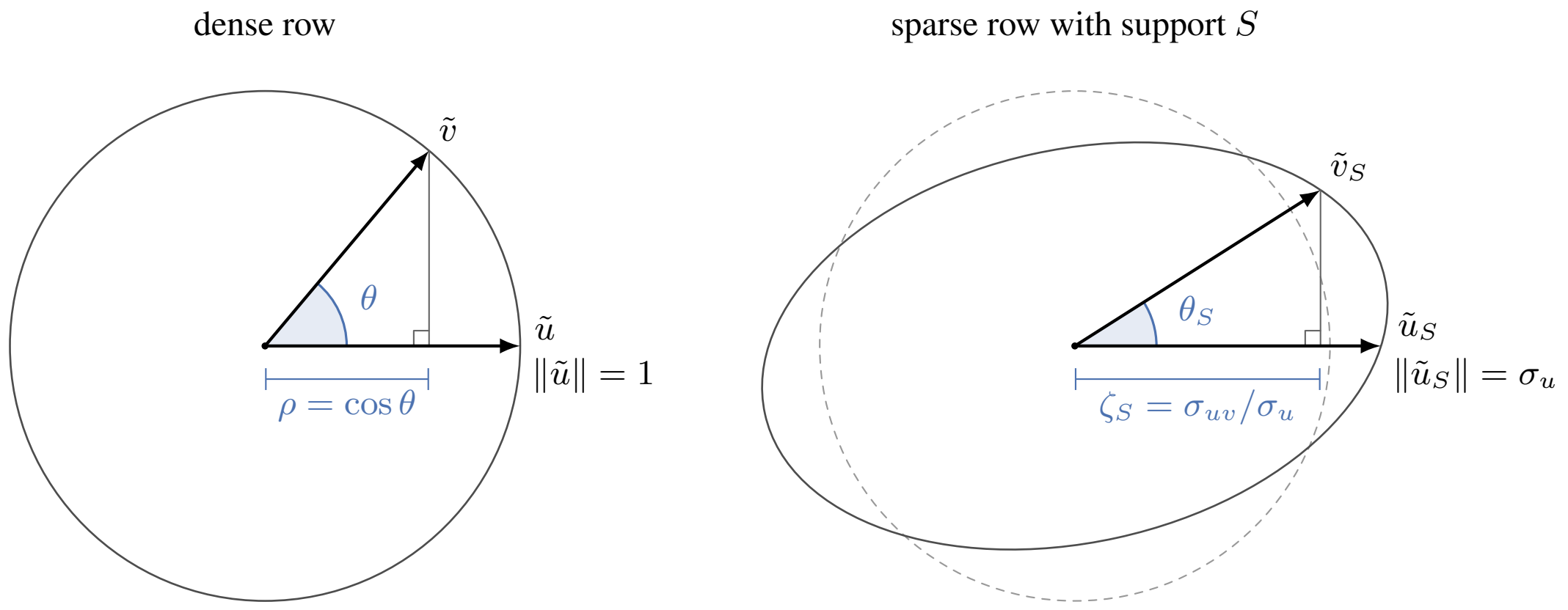

Figure~\ref{fig:metric} illustrates this effect in the two-dimensional plane spanned by $u$ and $v$. Coordinate subsampling maps the unit circle of the plane onto an ellipse that varies from row to row, sending $(\ut,\vt)$ to $(\ut_\capset,\vt_\capset)$ with squared norms $\sigma_u^2,\sigma_v^2$ and inner product $\sigma_{uv}$. Equivalently, each sparse row measures lengths and angles in $\mathrm{span}\{u,v\}$ using a random positive semidefinite $2\times2$ matrix $M_\capset$ in place of the Euclidean identity $I_2$; we call $M_\capset$ the \emph{random metric} induced by the row, and its expectation over $\capset$ is $I_2$ (Section~\ref{sec:random-metric}).

The three estimators extract three different geometric features from $(\ut_\capset,\vt_\capset)$. Each estimator averages a per-row score across the $\ell$ rows: GG averages the product $XY$, BG averages $Z:=\sqrt{\pi/2}\,\sign(X)\,Y$, and BB averages the indicator $\mathbf 1\{\sign X\ne\sign Y\}$. Conditioned on $\capset$, applying the classical dense Gaussian identities (Section~\ref{sec:dense}) directly to $(\ut_\capset,\vt_\capset)$ yields the conditional expectation of each score:
\[
\E[XY\mid\capset]=\sigma_{uv},\qquad
\E[Z\mid\capset]=\frac{\sigma_{uv}}{\sigma_u}=:\zeta_\capset,\qquad
\Pr[\sign X\ne\sign Y\mid\capset]=\frac{\theta_\capset}{\pi},
\]
where $\theta_\capset=\angle(\ut_\capset,\vt_\capset)$ satisfies $\cos\theta_\capset=\rho_\capset:=\sigma_{uv}/(\sigma_u\sigma_v)$. Notice the sharp divide between the linear and one-bit estimators: GG reads the inner product $\sigma_{uv}$, which is a linear functional of $M_\capset$. Because $\E M_\capset=I_2$, GG is completely unbiased at every sparsity level; sparse rows merely inflate its variance. By contrast, BG reads an inner product normalized by a length ($\sigma_{uv}/\sigma_u$), and BB reads an angle ($\theta_\capset$). Because neither functional is linear in $M_\capset$, Jensen's inequality and higher-order curvature convert random metric fluctuations into a systematic second-order shift in expectation, which we call the \emph{support bias}. Because every row is drawn from the same support distribution, averaging over more rows reduces the sampling variance but leaves the support bias intact; only increasing the row sparsity $r$, which dampens the metric distortion itself, can make this bias small.

A back-of-the-envelope calculation on a self-pair reveals the magnitude of the effect. Suppose we score a stored unit vector $u$ against itself using BG, so that $v=u$ and $\rho=1$. Then $\sigma_{uv}=\sigma_u^2$, the conditional mean score simplifies to $\zeta_\capset=\sigma_u$, and Jensen's inequality gives $\E\sigma_u\le\sqrt{\E\sigma_u^2}=1$. Expanding $\sigma_u=\sqrt{\sigma_u^2}$ to second order around $\sigma_u^2=1$ shows that the expected BG estimate falls short of $1$ by approximately $\Var(\sigma_u^2)/8$, where $\Var(\sigma_u^2)$ scales as $1/r$. Keeping this downward bias below $\eps$ therefore requires $r$ to grow at least as $1/\eps$. This phenomenon is already familiar from the $\ell_1$ embedding literature: when $v=u$, the BG estimate equals $\sqrt{\pi/2}\,\|\Phi u\|_1/\ell$, which is precisely Ailon and Chazelle's $\ell_1$ FJLT embedding, and the expansion of $\E\sigma_u$ is the calculation underlying \citet[Lemma~2]{ailon2009fast} that forces their row density to scale with $\eps^{-1}$.

Extending this analysis beyond self-pairs requires controlling two additional effects: for a general pair $(u,v)$, a sparse row distorts not only the length $\sigma_u$ of $\ut_\capset$ but also its correlation with the orthogonal direction in the plane, and in SimHash both vectors are quantized to signs. We derive explicit, nonasymptotic bounds on the resulting biases in terms of measurable properties of the preconditioned vectors.

Specifically, the severity of the metric distortion depends on how evenly the preconditioned vectors distribute their energy across coordinates. For any unit vector $x$, we quantify this uniformity using two summary statistics, the \emph{cap} $w_x$ and the \emph{fourth moment} $k_x$:
\[
w_x:=n\,\|\xt\|_\infty^2\qquad\text{and}\qquad k_x:=n\,\|\xt\|_4^4 .
\]
Measured relative to the uniform coordinate energy $1/n$, the cap $w_x$ records the largest squared coordinate of $\xt$, while the fourth moment $k_x$ records the average of the squared coordinate energies. These quantities satisfy $1\le k_x\le w_x\le n$, with $k_x=w_x=1$ if and only if $\xt$ is perfectly \emph{flat}, meaning that every coordinate has magnitude $n^{-1/2}$. The fourth moment governs the typical size of the metric distortion, since $\Var(\sigma_x^2)\le k_x/r$, whereas the cap governs extreme deviations, since no single sampled coordinate can contribute more than $w_x/r$ to $\sigma_x^2$.

\subsection{Balanced rows}\label{sec:balanced}

Alongside our analysis of the standard FJLT, we introduce a simple structural modification to the sparse projection matrix, which helps both computationally and statistically. In the classical FJLT of \citet{ailon2009fast}, each coordinate is included in a row's support independently with probability $q=r/n$. Under these \emph{Bernoulli supports}, rows have variable numbers of nonzeros---and may occasionally be entirely empty. Storing such a matrix requires CSR-style row-offset indices, vectorized implementations must pad or branch across ragged row lengths, and computational load balances unevenly across threads. To eliminate these drawbacks, we instead assign every row a \emph{balanced support} consisting of exactly $r$ nonzero positions chosen uniformly at random without replacement (Algorithm~\ref{alg:sketch}). We refer to the choice between Bernoulli and balanced sampling as the \emph{support law}. With balanced supports, the sparse matrix $P$ is represented simply as two rectangular $\ell\times r$ arrays (one of column indices and one of Gaussian weights, or generated on the fly from a PRNG seed), and the inner encoding loop has a fixed trip count $r$.

\algtext*{EndFunction}
\begin{algorithm}[t]
\caption{Sketching and scoring with the balanced FJLT}
\label{alg:sketch}
\begin{algorithmic}[1]
\Require shared randomness, drawn once or regenerated from a seed:
  random signs $D$; supports $\capset_1,\dots,\capset_\ell$, each a uniformly random
  subset of $\{1,\dots,n\}$ of size $r$; weights $g_{ij}\sim\mathcal N(0,n/r)$
\Function{Encode}{$x$} \Comment{$x\in\R^n\setminus\{0\}$, $n$ a power of two}
  \State $\xt\gets H(D\,x/\|x\|)$ \Comment{fast Walsh--Hadamard transform, $n\log_2n$ additions}
  \State optionally record $w_x=n\|\xt\|_\infty^2$ and $k_x=n\|\xt\|_4^4$
    \Comment{for certificates (Section~\ref{sec:practice}); $O(n)$}
  \State $X_i\gets\sum_{j\in\capset_i}g_{ij}\,\xt_j$ for $i=1,\dots,\ell$
    \Comment{$r\ell$ multiply--adds}
  \State \Return $(\|x\|,\sign X)$ for a one-bit code, $(\|x\|,X)$ otherwise
\EndFunction
\Function{Score}{$u,v$} \Comment{$X$ encodes $u$, $Y$ encodes $v$; in BG, $u$ is stored}
  \State GG: $\hat\rho_{\rm GG}=\frac1\ell\sum_iX_iY_i$; \ BG:
    $\hat\rho_{\rm BG}=\sqrt{\pi/2}\,\frac1\ell\sum_i\sign(X_i)\,Y_i$
    \Comment{or a robust mean}
  \State BB: $\hat\theta=\pi\,\mathrm{popcount}(\sign X\oplus\sign Y)/\ell$,
    \ $\hat\rho_{\rm BB}=\cos\hat\theta$
  \State \Return $\hat\rho$, or $\|u\|\,\|v\|\,\hat\rho$ to estimate $\inner uv$
\EndFunction
\end{algorithmic}
\end{algorithm}

In Algorithm~\ref{alg:sketch}, we assume that $n$ is a power of two (padding shorter vectors with zeros as needed). For GG and BG, the sample average over the $\ell$ rows may optionally be replaced by a robust mean estimator (Section~\ref{sec:dense}).

Beyond its systems benefits, balancing the rows directly improves statistical accuracy. Fixing the support size removes fluctuations in row length as a source of variance, so the captured energy $\sigma_x^2$ concentrates more tightly around $1$ (Section~\ref{sec:setting}). For Gaussian inputs, switching from Bernoulli to balanced supports reduces the support bias of BG by roughly one third (Section~\ref{sec:exp-balance}). Even more strikingly, whenever the preconditioned stored vector $\ut$ is flat, a balanced row captures its energy without error ($\sigma_u^2\equiv 1$), eliminating the support bias of BG altogether (Proposition~\ref{prop:bg-support}(iii)). For SimHash (BB), by contrast, the bias is nearly identical under both support laws because rescaling the metric of a row does not alter the angle between two vectors.

Fixed-cardinality row supports have previously been studied for linear projections by \citet{wolff2012randomness}, who combined them with Rademacher ($\pm 1$) weights to reduce randomness complexity in the FJLT (Section~\ref{sec:related}). Throughout this paper, we analyze Bernoulli and balanced supports in a unified manner, expressing every upper bound as a single formula parameterized by the support law.

\subsection{Main results}\label{sec:main-informal}

We now summarize our main contributions. Throughout, we consider $N$ pairs $(u,v)$ of unit vectors with inner product $\rho=\inner uv=\cos\theta$, where $u$ denotes the stored vector in BG. The sparse matrix $P$ is drawn with either Bernoulli or balanced supports, and we write $L:=\log(2N/\delta)$.
\begin{enumerate}\itemsep 3pt
\item[(i)] \emph{Error bounds for a realized preconditioner.} Fix the sign matrix $D$ arbitrarily, and let the $N$ vector pairs be chosen independently of the sparse matrix $P$. For each pair, let $U=[u_1\ \ u_2]$ be an orthonormal basis of $\mathrm{span}\{u,v\}$ and define the \emph{plane cap} $\wU$ as the maximum cap $w_x$ among all unit vectors $x\in\mathrm{span}\{u_1,u_2\}$. Then each of the following error bounds holds simultaneously across all $N$ pairs with probability at least $1-\delta$ over the draw of $P$, provided every pair satisfies the accompanying sparsity condition:
\begin{align*}
|\hat\rho_{\rm BG}-\rho|\ &\le\ 1.15\,\frac{w_u}{r}+1.86\sqrt{\frac L\ell}
&&\text{if}\quad r\ge64\,\wmax\ \ \text{and}\ \ \ell\ge14L,\\
|\hat\theta-\theta|\ &\le\ 1.8\,\frac{\wU}{r}+2.23\sqrt{\frac L\ell}
&&\text{if}\quad r\ge64\,\wU
\end{align*}
(Theorems~\ref{thm:bg-param} and~\ref{thm:bb-param}, where $\wmax:=\max\{w_u,w_v\}$). In each inequality, the second term matches the sampling noise of a dense sketch up to a constant factor near $1$, while the first term bounds the support bias, which is independent of $\ell$. We also establish sharper fourth-moment versions of both bounds that replace the cap in the leading bias term with a fourth moment, at the expense of an exponentially small tail term in $r$. Because fourth moments are typically much smaller than caps---for any vector or plane fixed prior to drawing $D$, the expected vector fourth moment is at most $3$ and the expected \emph{plane fourth moment} $\kU$ is at most $8$, whereas the cap of a typical vector grows as $\log n$---these refined bounds yield substantially tighter guarantees in practice.

\item[(ii)] \emph{Worst-case guarantees in classical $(n,N,\eps,\delta)$ form.} When $K$ unit vectors are fixed before drawing the random signs $D$, the Hadamard preconditioner ensures that every cap among them is bounded by the \emph{flatness level} $\mu(K):=2\log(4nK/\delta)=O(\log(nK/\delta))$ with probability at least $1-\delta/2$ (Lemma~\ref{lem:flat}). Substituting this bound shows that $\ell=O(\eps^{-2}\log(N/\delta))$ rows with $r=O(\eps^{-1}\log(nN/\delta))$ nonzeros per row suffice to estimate all $N$ cosine similarities to within $\eps$ with probability at least $1-\delta$ (Corollaries~\ref{cor:bg-classical} and~\ref{cor:bb-classical}). Encoding a vector thus requires $O(n\log n+\eps^{-3}\log(nN/\delta)\log(N/\delta))$ operations, compared to $O(n\,\eps^{-2}\log(N/\delta))$ for a dense projection. Moreover, SimHash retains its classical relative-error advantage for near-duplicate detection: achieving relative error $\eps\theta$ across all angles $\theta\ge\theta_0$ requires $O(\eps^{-2}\theta_0^{-1}\log(N/\delta))$ rows, just as in the dense case, while $\theta_0$ enters the row sparsity only logarithmically.

\item[(iii)] \emph{Matching lower bounds.} Set $\Lambda:=\log(\min\{N,n/2\}/\delta)$. When $\eps$ is small and each row samples at most half of the coordinates ($2\le r\le n/2$), the asymmetric estimator BG requires $r=\Omega(\eps^{-1}\Lambda)$ nonzeros per row on a worst-case family of at most $N$ pairs; the same sparsity lower bound holds for SimHash (BB) with balanced supports when estimating a small angle to relative accuracy $\eps$ (Theorem~\ref{thm:lower}). Whenever $N/\delta$ lies between two fixed powers of $n$, this lower bound matches the worst-case sparsity of item~(ii) for BG up to constant factors. While these lower bounds apply to the specific estimators studied here rather than to arbitrary boolean decoding rules, our BG lower bound extends to any row-aggregation rule that concentrates around the expected single-row score, including Catoni's robust mean estimator once $\ell$ is sufficiently large. Whether SimHash requires the $\eps^{-1}$ sparsity factor under additive error remains an open question.

\item[(iv)] \emph{Linear sketches at full precision.} Because GG suffers no support bias, row sparsity affects only its variance and tail behavior. Using the standard sample average, our framework recovers the optimal FJLT sparsity of \citet{fandina2023fast} with fully explicit constants (the original paper had inaccurate values), arbitrary failure probability $\delta$, and either Bernoulli or balanced supports (Corollary~\ref{cor:gg-classical}(i)). Furthermore, replacing the sample average with Catoni's robust mean estimator reduces the required sparsity to $r=3\,\mu(m)=O(\log(nm/\delta))$ nonzeros per row for estimating inner products against $m$ stored vectors, completely independent of the target accuracy $\eps\in(0,1]$ (Corollary~\ref{cor:gg-classical}(ii)). An analogous scale-free robust estimator preserves pairwise squared distances at $\eps$-independent sparsity for all $\eps\le\frac12$ (Appendix~\ref{app:gg}).
\end{enumerate}

Table~\ref{tab:compare} summarizes our worst-case bounds and their explicit constants alongside existing guarantees. Because worst-case bounds must protect against pathological alignments of the preconditioner, their constants are necessarily conservative: where both our corollaries and lower bounds apply, their required row sparsities differ by a factor of more than $5{,}000$ (Remark~\ref{rem:lb-scope}). In Section~\ref{sec:practice}, we show how evaluating our fixed-$D$ bounds directly on measured data parameters recovers most of this factor on typical inputs.

\begin{sidewaystable}\centering\footnotesize
\setlength{\tabcolsep}{4pt}
\renewcommand{\arraystretch}{1.2}
\newcommand{\tgroup}[1]{\multicolumn{5}{@{}l}{\textit{#1}}\\}
\begin{tabular}{@{}lllll@{}}
\toprule
result & decoder & sketch length $\ell$ & row sparsity $r$ & parameters\\
\midrule
\tgroup{GG: full precision on both sides}
dense JL \citep{johnson1984extensions,dasgupta2003elementary} & average & $12\eps^{-2}\log\frac{m^2}\delta$ & $n$ & ---\\
Ailon--Chazelle$^\dagger$ \citep{ailon2009fast} & average & $\Theta(\eps^{-2}\log m)$ & $\Theta(\log^2m)$ & ---\\
FHL$^\ddagger$ \citep{fandina2023fast} & average & $\Theta(\eps^{-2}\log m)$ & $\Theta\bigl(\log m\cdot\max\{1,\frac{\eps\log m}{\log(1/\eps)}\}\bigr)$ & ---\\
\textbf{Cor.~\ref{cor:gg-classical}(i)} & average & $12\eps^{-2}\log\frac{4|\Ufam|}\delta$ & $\mu(2|\Ufam|)\max\{8,\frac{2\eps\log(4\ell|\Ufam|/\delta)}{\log(1+1/\eps)}\}$ & $k_\Ufam$, $w_\Ufam$\\
\textbf{Cor.~\ref{cor:gg-classical}(ii)} & Catoni$^*$ & $8\eps^{-2}\log\frac{4N}\delta$ & $3\,\mu(m)$ & $w_u$ or $k_{uv}$\\
\addlinespace[3pt]
\tgroup{BG: one bit against full precision}
dense \citep{li2019signfull} & dot & $\pi\eps^{-2}\log\frac{2N}\delta$ & $n$ & ---\\
\textbf{Cor.~\ref{cor:bg-classical}(i)} & dot & $4\eps^{-2}\log\frac{4N}\delta$ & $16\max\{\eps^{-1},4\}\,\mu(|\Ufam|)$ & $\wmax$, $\kmax$\\
\textbf{Cor.~\ref{cor:bg-classical}(ii)} & Catoni$^*$ & $(3.9\eps^{-2}+2)\log\frac{4N}\delta$ & $16\max\{\eps^{-1},4\}\,\mu(m)$ & $w_u$ or $\kmax$\\
\addlinespace[3pt]
\tgroup{BB: one bit on both sides}
SimHash \citep{goemans1995improved,charikar2002similarity} & popcount & $\frac{\pi^2}2\eps^{-2}\log\frac{2N}\delta$ & $n$ & ---\\
\textbf{Cor.~\ref{cor:bb-classical}(i)} & popcount & $2\pi^2\eps^{-2}\log\frac{4N}\delta$ & $\mu(2N)\max\{\frac{14}\eps,\,13.5\log\frac{16\pi}\eps\}$ & $\kU$, $\wU$\\
SimHash, relative & popcount & $8.4\,\eps^{-2}\theta_0^{-1}\log\frac{2N}\delta$ & $n$ & ---\\
\textbf{Cor.~\ref{cor:bb-classical}(ii)} & popcount & $42\,\eps^{-2}\theta_0^{-1}\log\frac{4N}\delta$ & $\mu(2N)\max\{\frac{14}\eps,\,13.5\log\frac{16\pi}{\eps\theta_0}\}$ & $\kU$, $\wU$\\
\addlinespace[3pt]
\tgroup{Lower bounds}
\textbf{Thm.~\ref{thm:lower}(b)}, BG & dot, Catoni & $\Omega(\eps^{-2})$ & $>\frac\Lambda{178\eps}$ & ---\\
\textbf{Thm.~\ref{thm:lower}(c)}, BB & popcount & $\Omega(\eps^{-2}\theta^{-1})$ & $>\frac\Lambda{189\eps}$ (relative error) & ---\\
\bottomrule
\end{tabular}
\par\vspace{5pt}
\begin{minipage}{\textwidth}\footnotesize
Here $m$ is the number of points or stored vectors, $\Ufam$ is the set of unit vectors whose norms or pairs are covered, and $\mu(K):=2\log(4nK/\delta)$ is the flatness level for $K$ pre-specified vectors (Lemma~\ref{lem:flat}); notice that the Catoni corollaries ($^*$) pay $\mu(m)$ rather than $\mu(|\Ufam|)$ or $\mu(2N)$. The lower bounds apply when $2\le r\le n/2$, with $\eps\le1/256$ for BG and $\eps\le1/272$ at an angle $\theta\le\arctan(1/10)$ under balanced supports for BB; they state that once $\ell$ is of the indicated order, any sketch achieving error at most $\eps$ across all pairs of a worst-case family with probability at least $1-\delta/2$ must have row sparsity $r$ exceeding the bound shown. Every sparsity entry is capped at $n$, and we omit the small-$n$ branches $r\ge\eps n/(1+\eps)$ of Corollary~\ref{cor:gg-classical}(i) and $r=\Theta(\eps n)$ of \citet{fandina2023fast}.
\end{minipage}
\caption{Quantizing to one bit introduces a factor of $1/\eps$ in row sparsity while increasing sketch length by only a small constant factor. Entries report the parameters required to achieve accuracy $\eps$ simultaneously across all pairs with probability at least $1-\delta$; our results are shown in bold. Annotations: $^\dagger$success probability $2/3$ and $^\ddagger$success probability $1-1/m$, both assuming $n\le m$; $^*$only the $m$ stored vectors need to be fixed prior to drawing $D$, allowing queries to depend on $D$.}
\label{tab:compare}
\end{sidewaystable}

\paragraph{How to read Table~\ref{tab:compare}.}
Table~\ref{tab:compare} places our classical worst-case corollaries directly alongside the corresponding bounds for dense Gaussian projections and the linear FJLT. Comparing the sketch length column ($\ell$) vertically within each block shows that our sparse one-bit sketches require the same number of rows as dense sketches up to a modest constant factor: roughly $1.3\times$ for BG, $4\times$ for additive BB, and $5\times$ for relative-error BB. Comparing the row sparsity column ($r$) shows the fundamental difference between linear and signed projections: full-precision linear sketches require only a multiple of $\mu(K)$ that stays bounded as $\eps\to 0$, whereas one-bit sketches require $r$ to scale as $\mu(K)/\eps$, a dependence matched in order by the lower bounds in the bottom two rows. Writing the flatness level explicitly as $\mu(K)=2\log(4nK/\delta)$ also highlights how the Catoni rows ($^*$) depend only on the database size $m$ rather than the pair count $N$. The \emph{decoder} column specifies how the $\ell$ row scores of a pair are combined into a final estimate---either via the sample average (a dot product for BG and a population count of bit disagreements for BB) or via Catoni's robust mean estimator (Section~\ref{sec:dense}). Finally, the \emph{parameters} column lists the empirical caps and fourth moments that replace $\mu(K)$ when our theorems are evaluated conditionally on the realized preconditioner $D$, as defined in Sections~\ref{sec:setting} and~\ref{sec:random-metric}.

\subsection{What to expect in practice}\label{sec:practice}

What do these theoretical guarantees translate to at a realistic application scale? Table~\ref{tab:budgets} illustrates the required sketch budgets for a database of $m=10^9$ vectors in dimension $n=65{,}536$, where a single query is scored against all $10^9$ stored vectors ($N=10^9$) and we require every score to be accurate to within $\eps=0.1$ simultaneously with probability at least $1-\delta=0.9$.

\begin{table}[t]\centering\small
\setlength{\tabcolsep}{4.5pt}
\begin{tabular}{@{}lrrrrr@{}}
\toprule
 & \multicolumn{2}{c}{sketch length $\ell$} & \multicolumn{3}{c}{row sparsity $r$ ($r/n$)} \\
\cmidrule(lr){2-3}\cmidrule(l){4-6}
 & & & & \multicolumn{2}{c}{certified (realized $D$)} \\
\cmidrule(l){5-6}
guarantee & FJLT & dense & worst case & typical pair & database max. \\
\midrule
GG, average & 30{,}127 & 10{,}463 & 591 ($0.9$\%) & --- & 130 ($0.2$\%) \\
GG, Catoni & 19{,}530 & 9{,}536 & 214 ($0.3$\%) & --- & 5 ($<0.1$\%) \\
\addlinespace[2pt]
BG, dot & 9{,}765 & 7{,}452 & 11{,}361 ($17.3$\%) & 838\,/\,758 ($1.3$\,/\,$1.2$\%) & 1{,}623\,/\,1{,}545 ($2.5$\,/\,$2.4$\%) \\
BG, Catoni & 9{,}570 & 7{,}499 & 11{,}361 ($17.3$\%) & 477\,/\,398 ($0.7$\,/\,$0.6$\%) & 538\,/\,460 ($0.8$\,/\,$0.7$\%) \\
\addlinespace[2pt]
BB, popcount & 48{,}188 & 11{,}705 & 10{,}135 ($15.5$\%) & 849\,/\,842 ($1.3$\%) & 2{,}312\,/\,2{,}306 ($3.5$\%) \\
\bottomrule
\end{tabular}
\caption{Evaluating our bounds on measured data reduces the required one-bit row sparsity from $15$--$17\%$ of the ambient dimension $n$ in the worst case to roughly $1\%$ for a typical pair. Setting: $n=65{,}536$, $m=N=10^9$, $\eps=\delta=0.1$. Entries written $a\,/\,b$ denote Bernoulli\,/\,balanced supports; single entries apply to both. Note the slight \emph{improvement} for balanced supports.}
\label{tab:budgets}
\end{table}

Each row of Table~\ref{tab:budgets} corresponds to one of the corollaries in Table~\ref{tab:compare}. The \emph{worst case} sparsity column reports the a priori guarantee obtained by plugging in the union-bound flatness level $\mu(K)>70$, which forces the one-bit sketches to sample $15$--$17\%$ of the coordinates in each row. In practice, however, preconditioned vectors are far flatter than the worst-case union bound assumes, and the fixed-$D$ bounds of item~(i) allow us to take advantage of this fact. Because those theorems hold conditionally on the realized sign matrix $D$, we can plug in the actual caps and fourth moments measured on the preconditioned vectors rather than $\mu(K)$. We call the resulting guarantee a \emph{certificate}, as it certifies the accuracy of the deployed preconditioner on the dataset at hand. For BG, the certificate depends only on single-vector statistics that Algorithm~\ref{alg:sketch} records in $O(n)$ time during encoding; for BB, it uses the two-dimensional plane parameters of each pair (Section~\ref{sec:certificates}).

The last two columns of Table~\ref{tab:budgets} report the minimum row sparsity $r$ needed to certify accuracy $\eps=0.1$ at the corollary's sketch length $\ell$ when the preconditioned vectors are \emph{Gaussian-like} (meaning that $\xt=HDx$ is distributed like a normalized Gaussian vector, which holds exactly when the raw inputs $x$ are Gaussian). In the \emph{typical pair} column, evaluated at the median parameters of simulated pairs (caps around $20$ and vector fourth moments near $3$), the certified sparsity of a one-bit sketch drops to roughly $1\%$ of the coordinates---a $12$- to $29$-fold reduction relative to the worst-case bound. Even in the \emph{database max.}\ column, which plugs in the extreme parameter values expected across a collection of $10^9$ vectors, the certified sparsity remains below $4\%$ of the ambient dimension. (GG omits the typical-pair entry because its hypotheses are formulated over an entire vector family at once; see Appendix~\ref{app:k4} for a term-by-term derivation of every entry in the table.)

\begin{figure}[t]
\centering
\includegraphics[width=\textwidth]{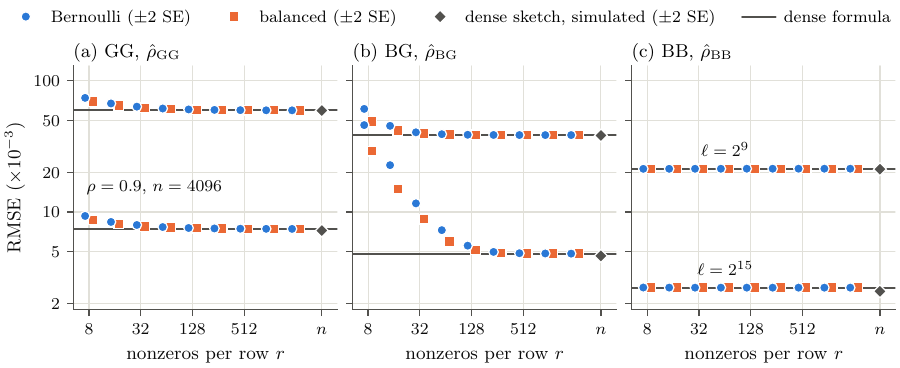}
\caption{Transition from dense to sparse rows at fixed sketch lengths: only the asymmetric estimator BG requires denser rows as the sketch length increases. Root mean squared error (RMSE) is plotted against the row sparsity $r$ for a pair of Gaussian vectors ($n=4096$, $\rho=0.9$) at sketch lengths $\ell=2^9$ and $\ell=2^{15}$, with error bars indicating $\pm2$ standard errors. Horizontal gray lines mark the theoretical RMSE of a dense Gaussian sketch, and diamonds mark end-to-end simulations at $r=n$.}
\label{fig:sparsity}
\end{figure}

Even a data-dependent certificate is an upper bound, and empirical errors on typical inputs are smaller still (Section~\ref{sec:experiments}). Figure~\ref{fig:sparsity} tracks the root mean squared error of all three estimators on a pair of Gaussian vectors in dimension $n=4096$ as the row sparsity $r$ decreases from $r=n$ (dense) down to $r=8$, at two sketch lengths ($\ell=2^9$ and $\ell=2^{15}$). SimHash (BB) suffers virtually no loss in accuracy down to $r=8$, and the unbiased linear sketch (GG) degrades by at most $7\%$ once $r\ge 32$. By contrast, BG clearly exhibits the bias--variance trade-off of item~(i): increasing the sketch length from $2^9$ to $2^{15}$ reduces the dense sampling noise by a factor of eight, so BG requires eight times as many nonzeros per row to keep its support bias within $10\%$ of the noise floor---just as the lower bound of item~(iii) dictates in the worst case. Nevertheless, sampling just one coordinate in sixteen ($r=256$) suffices to bring every estimator within $10\%$ of the dense baseline across both support laws and both sketch lengths.

Because sparse projections save computation only when $r\ll n$, the practical speedup of the FJLT grows with the ambient dimension. For a typical vector pair, our one-bit certificates require a few hundred nonzeros per row, a number that depends only weakly on $n$ (since fourth moments are dimension-free and caps grow only as $\log n$). Consequently, the computational savings are substantial when $n$ is in the thousands or tens of thousands. Conversely, at the dimension of a single transformer attention head ($n\approx 128$), the certified row sparsity exceeds the ambient dimension, and our theory appropriately recommends using dense Gaussian projections, which are already inexpensive at that scale.

\subsection{Related work}\label{sec:related}

\paragraph{The linear FJLT.}
In their foundational paper, \citet{ailon2009fast} proved that $r=\Theta(\log^2m)$ nonzeros per row suffice to preserve the Euclidean norms of $m$ points (Theorem~\ref{thm:ac09}), a bound recently sharpened by \citet{fandina2023fast} to the optimal sparsity of Theorem~\ref{thm:fhl23}. Both analyses consider Bernoulli supports and estimate norms via the sample average of squared projections. For embeddings into $\ell_1$, the row density of \citet{ailon2009fast}---as well as its exposition by \citet[Thm.~5.1(i)]{matousek2008variants}---carries an extra factor of $\eps^{-1}$. Because our lower bound for BG is constructed using self-pairs, it simultaneously proves that this $\eps^{-1}$ density factor is necessary for the standard $\ell_1$ FJLT estimator, settling a question that was, to our knowledge, previously open (Section~\ref{sec:lower}).

\paragraph{Balanced rows.}
\citet[Thm.~3.1]{wolff2012randomness} previously studied sparse projection rows whose supports are drawn uniformly among subsets of fixed size $r$, populated with random $\pm 1$ signs, obtaining an $O(\log^2m)$ sparsity guarantee analogous to Theorem~\ref{thm:ac09}. To our knowledge, Corollary~\ref{cor:gg-classical}(i) provides the first guarantee for balanced supports that achieves the optimal sparsity of \citet{fandina2023fast}, while our fourth-moment bounds (Theorem~\ref{thm:gg-param}(a)) and robust Catoni bounds (Theorem~\ref{thm:gg-param}(b), together with a scale-free counterpart for squared distances in Appendix~\ref{app:gg}) are new under both Bernoulli and balanced supports. We note that the sparse Johnson--Lindenstrauss transforms of \citet{kane2014sparser} and \citet{nelson2013osnap} instead fix the number of nonzeros per \emph{column} of the projection matrix, which introduces dependencies across rows and requires a different analytical toolkit.

\paragraph{Signs of sparse projections.}
Prior analyses of signed sparse projections have been purely asymptotic, characterizing behavior in the limit as the ambient dimension tends to infinity \citep{li2006verysparse,kim2018sparsecbe,dubey2022cssrp}. When the nonzero weights are $\pm 1$ Rademacher variables, the standard analytical tool is a Berry--Esseen normal approximation, which incurs an $O(r^{-1/2})$ approximation error even on perfectly flat vectors \citep{li2006verysparse}; a similar obstacle arises in one-bit compressed sensing with non-Gaussian measurements when signals contain spiky coordinates \citep{ai2014onebit}. Using Gaussian weights avoids the Berry--Esseen barrier entirely: conditioned on its support, each sparse row is \emph{exactly} Gaussian, isolating the $O(1/r)$ metric-distortion bias that we bound. Related architectures include \citet{zhang2021faster}, who apply the same preconditioned sparse transform $PHD$ but quantize via Sigma--Delta noise shaping rather than memoryless signs, and Fast-TurboQuant \citep{pereira2026fastturboquant}, which signs $HDx$ directly without a sparse Gaussian layer or formal error bounds for the signed estimator.

\paragraph{Fast binary embeddings with dependent rows.}
Most existing fast binary embedding schemes achieve speed by replacing the dense Gaussian matrix with a structured circulant or Toeplitz Gaussian matrix \citep{yu2014circulant,yi2015binary,dirksen2018fast}. While each individual row of a Gaussian Toeplitz matrix has the exact marginal distribution of a dense Gaussian row, all $\ell$ rows are generated by shifting a single underlying Gaussian vector, creating strong inter-row dependencies. Even for linear Toeplitz sketches with $\pm 1$ entries, this dependence inflates the required sketch length by an extra factor of $\log(1/\delta)$ \citep{freksen2020toeplitz}, and for signed Toeplitz embeddings, a gap in the variance analysis of \citet{yi2015binary} required substantial technical repair by \citet{dirksen2018fast}. \citet{choromanska2016binary} analyzed a broader class of structured Gaussian matrices preceded by a randomized Hadamard preconditioner $HD$, showing that preconditioning makes pairs of rows nearly independent on fixed inputs; however, their guarantees hold only for sufficiently large $n$ and yield a slower $\ell^{-1/3}$ convergence rate for angles in the Toeplitz case due to residual row dependencies. The signed FJLT makes the opposite design trade-off: conditioned on $D$, its rows are strictly independent, and the price of speed is paid instead through the per-row metric distortion that generates support bias.

\paragraph{Two-stage dimension reduction followed by signing.}
One can eliminate support bias entirely by composing two separate transforms: first projecting the vectors to an intermediate dimension $n'$ via a linear FJLT, and then applying a dense signed Gaussian projection in $\R^{n'}$. Originally proposed as a baseline by \citet[Alg.~3]{yi2015binary} and analyzed by \citet{dirksen2018fast} as the accelerated Gaussian binary embedding, this two-stage pipeline incurs an extra factor of $1/\eps$ in encoding complexity compared to signing sparse FJLT rows directly (Section~\ref{sec:certificates}).

\subsection{Outline and notation}\label{sec:outline}

The remainder of the paper is organized as follows. Section~\ref{sec:dense} reviews the classical guarantees for dense Gaussian projections in a form that directly parallels our sparse theorems, and Section~\ref{sec:fjlt} reviews the linear FJLT and illustrates how sign quantization breaks its standard analysis. Section~\ref{sec:results} presents our main theoretical results, Section~\ref{sec:experiments} validates the theory against numerical simulations, and Section~\ref{sec:discussion} concludes with a discussion and open problems. All mathematical proofs are deferred to Appendix~\ref{app:proofs}, and experimental details are documented in Appendix~\ref{app:experiments}.

Throughout the paper, $\|x\|$ denotes the Euclidean norm of a vector $x$, $\|x\|_p$ denotes its $\ell_p$ norm (so that $\|x\|_4^4=\sum_jx_j^4$), and $\|x\|_\infty=\max_j|x_j|$ denotes its maximum coordinate magnitude. All logarithms are natural unless a base is explicitly specified. To ensure that every sign requires exactly one bit of storage, we adopt the convention $\signzero$. We let $u,v\in\R^n$ denote unit vectors with inner product $\rho=\inner uv=\cos\theta$ and angle $\theta\in[0,\pi]$, while $\eps$ and $\delta$ denote the target accuracy and failure probability, respectively. A comprehensive summary of notation is provided in Table~\ref{tab:notation} in Appendix~\ref{app:tools}.

\section{Background}

To establish a clear baseline, we begin by reviewing both dense Gaussian projections for GG (Johnson--Lindenstrauss) and one-bit quantization for BG (asymmetric) and BB (SimHash). We state each dense guarantee in a form that directly mirrors our sparse theorems in Section~\ref{sec:results}, making the exact statistical price of sparsity easy to see by comparison.

\subsection{Dense projections}\label{sec:dense}

For dense projections the behavior of all three estimators (GG, BG and BB) is well understood. Let $\Phi\in\R^{\ell\times n}$ be a random projection matrix with rows $\varphi_1,\dots,\varphi_\ell$, mapping a pair of unit vectors $u,v\in\R^n$ with inner product $\inner uv=\rho=\cos\theta$ to the scalar row projections $X_i=\inner{\varphi_i}u$ and $Y_i=\inner{\varphi_i}v$. Each row $i$ produces a single \emph{row score}, and the three estimators aggregate these scores across the $\ell$ rows:
\begin{equation}
\begin{array}{@{}lll@{}}
\text{GG:}&\text{score } X_iY_i,&\hat\rho_{\rm GG}=\tfrac1\ell\textstyle\sum_iX_iY_i;\\[3pt]
\text{BG:}&\text{score } Z_i:=\sqrt{\pi/2}\,\sign(X_i)\,Y_i,&\hat\rho_{\rm BG}=\tfrac1\ell\textstyle\sum_iZ_i;\\[3pt]
\text{BB:}&\text{score } \mathbf 1\{\sign X_i\ne\sign Y_i\},&\hat\theta=\tfrac\pi\ell\,\#\{i:\sign X_i\ne\sign Y_i\},\quad \hat\rho_{\rm BB}=\cos\hat\theta.
\end{array}
\label{eq:estimators}
\end{equation}
In the asymmetric estimator BG, only the stored vector $u$ is quantized to one bit. Up to the scaling factor $\sqrt{\pi/2}/\ell$, the BG sample average is simply the inner product between the sign code $\sign(X)$ and the uncompressed query projection $Y$; we therefore refer to it as the \emph{dot-product estimate}. More generally, we use the term \emph{decoder} for any aggregation rule that maps the $\ell$ row scores of a pair to an inner product estimate, including both the sample average and Catoni's robust mean estimator introduced at the end of this subsection.

When $\Phi=G$ has independent standard normal entries $\mathcal N(0,1)$, three classical Gaussian identities ensure that all three estimators are unbiased or consistent. We state these identities for arbitrary nonzero vectors in any dimension $d$, as Section~\ref{sec:random-metric} will apply them directly to the subsampled vectors $(\ut_\capset,\vt_\capset)$. For any nonzero $x,y\in\R^d$ with angle $\angle(x,y)\in[0,\pi]$ and $g\sim\mathcal N(0,I_d)$,
\begin{equation}
\begin{gathered}
\E\big[\inner gx\inner gy\big]=\inner xy,\qquad
\E\big[\sign\inner gx\,\inner gy\big]=\sqrt{\frac2\pi}\,\frac{\inner xy}{\|x\|},\\
\Pr\big[\sign\inner gx\ne\sign\inner gy\big]=\frac{\angle(x,y)}\pi .
\end{gathered}
\label{eq:dense-identities}
\end{equation}
The first two identities follow from a direct Gaussian integration, while the third reflects the geometric fact that a random hyperplane through the origin separates $x$ and $y$ with probability proportional to the angle between them \citep[Lemma~3.2]{goemans1995improved}. Specializing to unit vectors $u$ and $v$ yields expectations $\rho$, $\sqrt{2/\pi}\,\rho$, and $\theta/\pi$, which explains the normalization constants $\sqrt{\pi/2}$ and $\pi$ in \eqref{eq:estimators}.

The analysis of dense Gaussian projections is simplified enormously by rotational invariance: regardless of the specific unit vectors $u$ and $v$, the projection pairs $(X_i,Y_i)$ are independent standard bivariate normal vectors with correlation $\rho$, so the error distribution depends on the data solely through $\rho$. For the linear sketch GG, the classical guarantee is stated in terms of pairwise Euclidean distances.

\begin{theorem}[Dense GG, Johnson--Lindenstrauss]\label{thm:dense-gg}
Let $\mathcal X\subset\R^n$ be a set of $m$ points fixed in advance, and let $\delta\in(0,1)$ and $\ell\ge12\log(m^2/\delta)$ \citep{johnson1984extensions,dasgupta2003elementary}. With probability at least $1-\delta$, all pairs $x,y\in\mathcal X$ satisfy
\[
\Big|\tfrac1\ell\|G(x-y)\|^2-\|x-y\|^2\Big|\ \le\ \sqrt{\frac{12\log(m^2/\delta)}{\ell}}\;\|x-y\|^2 .
\]
Equivalently, $\ell\ge12\,\eps^{-2}\log(m^2/\delta)$ rows suffice to guarantee relative squared-norm distortion at most $\eps$ for any $\eps\in(0,1]$. The sharper bound $\ell\ge4(\eps^2-\tfrac23\eps^3)^{-1}\log(m^2/\delta)$ also suffices.
\end{theorem}
Theorem~\ref{thm:dense-gg} follows immediately by applying standard $\chi^2_\ell$ tail bounds \citep[Lemma~4.1]{achlioptas2003database} to $\|Gz\|^2$ for each normalized difference vector $z=(x-y)/\|x-y\|$ and taking a union bound over all $\binom m2$ pairs. Guarantees for inner products follow by polarization: since $4\inner uv=\|u+v\|^2-\|u-v\|^2$ and $\|u+v\|^2+\|u-v\|^2=4$ for unit vectors, preserving the squared norms of $u+v$ and $u-v$ to relative accuracy $\eps$ ensures $|\hat\rho_{\rm GG}-\rho|\le\eps$. Across $N$ pre-specified pairs of unit vectors, this holds simultaneously with probability at least $1-\delta$ under the same sketch-length condition, with $\log(4N/\delta)$ replacing $\log(m^2/\delta)$. Theorem~\ref{thm:gg-param} provides the sparse FJLT counterpart.

Next, consider quantizing the stored vector to one bit (BG). Because the query projection $Y_i$ remains Gaussian, the per-row score $Z_i$ retains sub-Gaussian tails.

\begin{theorem}[Dense BG, after \citet{li2019signfull}]\label{thm:dense-bg}
For any unit vectors $u$ and $v$, the BG row score satisfies $\E Z_i=\rho$, $\Var Z_i=\pi/2-\rho^2$, and $\E e^{\lambda(Z_i-\rho)}\le e^{\pi\lambda^2/4}$ for all $\lambda\in\R$. Consequently, for any $\delta\in(0,1)$ and any $N$ pairs of unit vectors fixed in advance, with probability at least $1-\delta$ simultaneously across all pairs,
\[
|\hat\rho_{\rm BG}-\rho|\ \le\ \sqrt{\frac{\pi\log(2N/\delta)}{\ell}} .
\]
Equivalently, $\ell\ge\pi\eps^{-2}\log(2N/\delta)$ rows guarantee additive error at most $\eps$.
\end{theorem}
While the mean and variance formulas are due to \citet[Thm.~3]{li2019signfull}, the sharp moment generating function bound $\E e^{\lambda(Z_i-\rho)}\le e^{\pi\lambda^2/4}$ is new; we derive it in Appendix~\ref{app:bg} and combine it with a Chernoff and union bound. To our knowledge, no valid concentration inequality with the optimal leading constant $\pi$ has previously appeared in the literature. Theorem~\ref{thm:bg-param}(a) establishes the sparse counterpart.

Notably, the constant $\pi$ in Theorem~\ref{thm:dense-bg} cannot be improved. When $u$ and $v$ are orthogonal ($\rho=0$), the projections $X_i$ and $Y_i$ are independent, so $\sign(X_i)Y_i$ is distributed as a standard normal variable, making $Z_i\sim\mathcal N(0,\pi/2)$ and $\hat\rho_{\rm BG}\sim\mathcal N(0,\pi/(2\ell))$ exact. The tail probability $\Pr[\hat\rho_{\rm BG}>t]$ therefore decays as $e^{-\ell t^2/\pi}$ up to a polynomial prefactor in $t\sqrt\ell$. In particular, the smaller constant claimed for this estimator in the QJL analysis of \citet[Lemma~3.5]{zandieh2024qjl} is incorrect.\footnote{\citet[Lemma~3.5]{zandieh2024qjl} claim that $\ell\ge\frac43(1+\eps)\eps^{-2}\log(2/\delta)$ rows suffice for a single pair. At $\eps=0.1$ and $\delta=10^{-6}$, this formula prescribes $\ell=2128$ rows, for which the exact two-sided Gaussian failure probability at $\rho=0$ is $2.3\cdot10^{-4}$---more than $200$ times the target $\delta$; in reality, $3759$ rows are required. The error stems from their application of Bernstein's inequality, where the second moment of a row score for unit vectors is entered as $2/3$ instead of the true value $\pi/2$. Substituting $\pi/2$ into their proof yields $\ell\ge(\pi+\frac43\eps)\eps^{-2}\log(2/\delta)$, recovering the leading constant $\pi$ of Theorem~\ref{thm:dense-bg}. Their Theorem~3.6 for attention-score estimation relies on this lemma and requires the same correction.}

From a variance perspective, quantizing the stored vector to one bit is remarkably inexpensive. Whereas a linear GG row score has variance $1+\rho^2$, a BG row score has variance $\pi/2-\rho^2$, which is at most $\pi/2\approx 1.57$ times larger at $\rho=0$ and actually becomes \emph{smaller} than the GG variance whenever $\rho^2>(\pi-2)/4$ (i.e., $|\rho|\gtrsim 0.53$). As we show in Section~\ref{sec:experiments}, row sparsity shifts this variance crossover even further in favor of one-bit quantization.

Finally, when both vectors are quantized to one bit (BB), the estimator counts the fraction of disagreeing signs.

\begin{theorem}[Dense BB, SimHash]\label{thm:dense-bb}
Let $\delta\in(0,1)$, and fix $N$ pairs of unit vectors in advance \citep{goemans1995improved,charikar2002similarity}. For each pair, $\Pr[\sign X_i\ne\sign Y_i]=\theta/\pi$.
\begin{enumerate}
\item[(a)] With probability at least $1-\delta$, all pairs comply with $|\hat\theta-\theta|\le\pi\sqrt{\log(2N/\delta)/(2\ell)}$, and the same bound holds for $|\hat\rho_{\rm BB}-\rho|$. Equivalently, $\ell\ge\frac{\pi^2}2\eps^{-2}\log(2N/\delta)$ rows guarantee error at most $\eps$.
\item[(b)] Let $\eps\in(0,1]$ and $\theta_0\in(0,\pi]$. If $\ell\ge8.4\,\eps^{-2}\theta_0^{-1}\log(2N/\delta)$, then with probability at least $1-\delta$, every pair with $\theta\ge\theta_0$ satisfies $|\hat\theta-\theta|\le\eps\theta$, and every pair with $\theta=0$ satisfies $\hat\theta=0$.
\end{enumerate}
\end{theorem}
Because the empirical disagreement fraction $\hat\theta/\pi$ is the average of $\ell$ independent $\mathrm{Bernoulli}(\theta/\pi)$ indicators, each with variance at most $\theta/\pi$, part~(a) follows immediately from Hoeffding's inequality and a union bound \citep[Lemma~2]{jacques2013robust}, transferring to $\hat\rho_{\rm BB}=\cos\hat\theta$ because cosine is $1$-Lipschitz. Part~(b) follows from Bernstein's inequality: identical vectors ($\theta=0$) produce identical signs in every row, while for any pair with $\theta\ge\theta_0$,
\[
\Pr\big[|\hat\theta-\theta|>\eps\theta\big]\ \le\ 2\exp\Big\{-\frac{\ell\,\eps^2\theta}{2\pi(1+\eps/3)}\Big\}\ \le\ 2\exp\Big\{-\frac{\ell\,\eps^2\theta_0}{8.4}\Big\}\ \le\ \frac\delta N ,
\]
since $2\pi(1+\eps/3)\le8\pi/3\le8.4$ for all $\eps\in(0,1]$.

Part~(b) explains why SimHash excels at near-duplicate detection: achieving relative error $\eps$ down to a minimum angle $\theta_0$ requires only $O(\theta_0^{-1})$ rows, whereas applying the additive bound of part~(a) at accuracy $\eps\theta_0$ would require $O(\theta_0^{-2})$ rows. Theorem~\ref{thm:bb-param} establishes sparse counterparts for both regimes.

We also note that the bit complexity of part~(a) is information-theoretically optimal up to constant factors: preserving pairwise cosines among $m$ points uses $O(\eps^{-2}\log(m/\delta))$ bits per point, matching the $\Omega(\eps^{-2}\log(m/\delta))$ lower bound of \citet[Thm.~3.1]{yi2015binary} for any data-oblivious binary embedding when $n$ is at least of this order. Fast sparse one-bit sketches therefore reduce encoding time rather than code length.

\paragraph{Robust aggregation via Catoni's estimator.}
A standard sample average is vulnerable to rare outlier scores, which occur in sparse sketching whenever a row's support happens to land on several large coordinates of a vector at once. We can avoid paying for these heavy tails by aggregating row scores with Catoni's robust mean estimator \citep{catoni2012challenging}, whose high-probability concentration requires only a bound on the variance of each row score. Given scalar row scores $W_1,\dots,W_\ell$ (representing $X_iY_i$ in GG or $Z_i$ in BG) and a scale parameter $\alpha>0$, Catoni's estimator $\Cat_\alpha(W_1,\dots,W_\ell)$ is defined as the unique root $c\in\R$ of the estimating equation
\begin{equation}
\textstyle\sum_{i=1}^\ell\psi\big(\alpha(W_i-c)\big)=0,\qquad\text{where}\quad\psi(x):=\sign(x)\log\big(1+|x|+\tfrac{x^2}2\big).
\label{eq:catoni-def}
\end{equation}
Because the influence function $\psi$ grows only logarithmically for large arguments, a handful of extreme row scores cannot pull the root $c$ far from the mean.

Specifically, if the $\ell$ scores of each of $N$ pairs are independent with common mean and variance at most $\sigma^2$, and if $\ell>2L$ with $L:=\log(2N/\delta)$, then for an appropriate choice of $\alpha$, all $N$ Catoni estimates simultaneously lie within $\sqrt{2\sigma^2L/(\ell-2L)}$ of their expectations with probability at least $1-\delta$ (Appendix~\ref{app:tools}). Up to the minor degrees-of-freedom adjustment $2L$, this matches the sub-Gaussian tail of a sample average of $\ell$ truly Gaussian variables with variance $\sigma^2$. We use this robust decoder in part~(b) of Theorems~\ref{thm:gg-param} and~\ref{thm:bg-param}, and show in Remark~\ref{rem:catoni-fast} that a one-step variant can be evaluated as cheaply as a standard dot product.

\subsection{The Fast Johnson--Lindenstrauss Transform}\label{sec:fjlt}

The fast Johnson--Lindenstrauss transform (FJLT) accelerates dense Gaussian sketching by factoring the projection into random coordinate sign flips, a Walsh--Hadamard transform, and a sparse Gaussian matrix. It provides the second component of our work. We now review its formal definition, its classical norm-preservation guarantees, and the two-stage architecture of its proof. Our analysis of one-bit sketches preserves the first stage (preconditioning) verbatim, but requires a new approach for the second stage (sparse sampling), for reasons we illustrate at the end of the section.

\begin{definition}[FJLT \citep{ailon2009fast}]\label{def:fjlt}
Let $n\ge2$ be a power of two, and let $\ell\ge1$ and $q\in(0,1]$. The FJLT is the random matrix $\Phi=PHD\in\R^{\ell\times n}$, where $D$ is a diagonal matrix of independent Rademacher signs ($\pm1$ with probability $\frac12$ each), $H$ is the normalized Walsh--Hadamard matrix, and $P$ is a sparse Gaussian matrix with entries $P_{ij}=\mathbf 1\{j\in\capset_i\}\,g_{ij}$. Here $\capset_i\subseteq\{1,\dots,n\}$ denotes the \emph{support} of row~$i$, and the weights $g_{ij}\sim\mathcal N(0,1/q)$ are mutually independent. The row supports $\capset_1,\dots,\capset_\ell$ are independent and identically distributed, and $D$, $\{\capset_i\}$, and $\{g_{ij}\}$ are mutually independent. Under \emph{Bernoulli supports}---the original specification of \citet{ailon2009fast}---each coordinate $j$ is included in $\capset_i$ independently with probability $q$, so that $r:=qn$ is the expected number of nonzeros per row (the row sparsity).
\end{definition}

Applying $\Phi$ to a vector in $\R^n$ requires $O(n\log n)$ additions for the fast Walsh--Hadamard transform $HD$, followed by roughly $\ell r$ multiply--adds for the sparse matrix $P$. The central theoretical question is how small the row sparsity $r$ can be made without sacrificing accuracy. \citet{ailon2009fast} proved that $r=\Theta(\log^2m)$ nonzeros per row suffice to preserve the Euclidean norms of $m$ points.\footnote{Theorem~\ref{thm:ac09} states the Euclidean case of \citet[Lemma~1]{ailon2009fast} in our normalization. Their $\ell_1$ embedding requires row density $q=\Theta(\eps^{-1}\log m/n)$, whose extra factor of $\eps^{-1}$ is explained at the end of this section. Note also that Theorems~\ref{thm:ac09} and~\ref{thm:fhl23} bound unsquared norms $\|x\|$; guarantees for squared norms, which we use in our results, follow by replacing $\eps$ with $\sqrt{1+\eps}-1\ge\eps/3$.}

\begin{theorem}[FJLT {\citep[Lemma~1]{ailon2009fast}}]\label{thm:ac09}
Let $\mathcal X\subset\R^n$ be a set of $m$ vectors with $n\le m$, and let $\eps\in(0,1)$. Let $\Phi=PHD$ have Bernoulli supports of density $q=r/n$ with row sparsity $r=\min\{c_1\log^2m,\,n\}$ for a sufficiently large universal constant $c_1$, and sketch length $\ell=\Theta(\eps^{-2}\log m)$. With probability at least $2/3$,
\[
(1-\eps)\|x\|\le\ell^{-1/2}\|\Phi x\|\le(1+\eps)\|x\|\qquad\text{for all }x\in\mathcal X,
\]
and evaluating $\Phi x$ requires $O\big(n\log n+\min\{n\eps^{-2}\log m,\ \eps^{-2}\log^3m\}\big)$ operations.
\end{theorem}
The proof of Theorem~\ref{thm:ac09} proceeds in two stages. The first stage is \emph{flattening}: multiplying a unit vector $x$ by random signs $D$ and applying the Hadamard transform $H$ spreads its energy across all $n$ coordinates, ensuring with high probability that no single coordinate of $\xt=HDx$ carries more than a logarithmic multiple of the average energy $1/n$. As introduced in Section~\ref{sec:why}, we measure this peak coordinate energy by the cap $w_x=n\|\xt\|_\infty^2$, which ranges from $1$ when $\xt$ is perfectly flat to $n$ when $\xt$ is aligned with a standard basis vector.

\begin{lemma}[Flatness, after \citet{ailon2009fast}]\label{lem:flat}
Let $\Ufam\subset\R^n$ be a finite family of unit vectors fixed independently of $D$, and let $\delta_0\in(0,1)$. With probability at least $1-\delta_0$ over the draw of $D$, every vector $x\in\Ufam$ satisfies $w_x\le 2\log(2n|\Ufam|/\delta_0)$.
\end{lemma}
Lemma~\ref{lem:flat} is simply the flattening step of \citet[Lemma~1, inequalities (4) and~(5)]{ailon2009fast} with an explicit constant: each coordinate $\xt_j$ is a linear combination of independent Rademacher signs whose squared coefficients sum to $\|x\|^2/n=1/n$, so Hoeffding's inequality yields $\Pr[n\xt_j^2\ge t]\le2e^{-t/2}$, and setting $t=2\log(2n|\Ufam|/\delta_0)$ with a union bound over all $n|\Ufam|$ coordinates completes the proof.

To make comparisons across our corollaries uniform, for a target overall failure probability $\delta\in(0,1)$ and an effective count $K>0$ of vectors to be flattened, we define the \emph{flatness level}
\begin{equation}\label{eq:mu-def}
\mu(K):=2\log\frac{4nK}{\delta},
\end{equation}
which equals the bound of Lemma~\ref{lem:flat} on $K$ vectors at $\delta_0=\delta/2$ (and on $K/2$ vectors at $\delta_0=\delta/4$). Because $HD$ is an orthogonal transformation, it preserves all inner products and angles exactly; its sole purpose is to regularize the coordinate distribution of the inputs before sparse sampling, and we therefore call $HD$ the \emph{preconditioner}.

The second stage of the classical proof is \emph{energy capture}. Condition on a realized sign matrix $D$. Row~$i$ of $P$ inspects only the coordinates in its support $\capset_i$, capturing from $x$ the rescaled energy $\sigma_{x,i}^2:=\|\xt_{\capset_i}\|^2$, whose expectation over $\capset_i$ is $1$. Conditioned on $\capset_i$, the scalar projection $(\Phi x)_i$ is distributed as $\mathcal N(0,\sigma_{x,i}^2)$, so the squared norm estimate $\ell^{-1}\|\Phi x\|^2$ is a weighted average of independent $\chi^2_1$ variables with weights $\sigma_{x,i}^2/\ell$. When $\xt$ has a small cap $w_x$, no single coordinate can dominate $\sigma_{x,i}^2$, causing $\sigma_{x,i}^2$ to concentrate around $1$.

Specifically, \citet[Lemma~3]{ailon2009fast} showed that when $w_x\le\mu$, choosing $r=\Theta(\mu\log m)$ keeps every row's captured energy $\sigma_{x,i}^2$ within $[\frac12,2]$ and their average within $\eps$ of $1$ with high probability across all $x\in\mathcal X$. Applying a standard Chernoff bound to the weighted $\chi^2$ sum then yields Theorem~\ref{thm:ac09}. Under their assumption $n\le m$ and constant failure probability $\delta_0$, the flatness bound is $\mu=\Theta(\log m)$, resulting in a row sparsity of $r=\Theta(\log^2m)$. Recently, \citet{fandina2023fast}---hereafter FHL---sharpened this second stage to obtain the optimal sparsity achievable via a per-vector union bound.

\begin{theorem}[Sharp FJLT sparsity {\citep[Thm.~1]{fandina2023fast}}]\label{thm:fhl23}
Let $\mathcal X\subset\R^n$ be a set of $m\ge2$ points with $n\le m$, let $0<\eps\le c$ for a universal constant $c$, and let $\ell=\Theta(\eps^{-2}\log m)$ with $\ell\le n$. Under Bernoulli supports of density $q=r/n$ with row sparsity
\[
r=C\min\Big\{\eps n,\ \log m\cdot\max\Big(1,\frac{\eps\log m}{\log(1/\eps)}\Big)\Big\}
\]
for a suitable universal constant $C$, the map $\ell^{-1/2}PHD$ preserves all pairwise Euclidean distances in $\mathcal X$ to within $1\pm\eps$ with probability at least $1-1/m$.
\end{theorem}
The key refinement of FHL lies in how they control the captured energies $\{\sigma_{x,i}^2\}$. Standard Bernstein-type tail bounds for weighted sums of $\chi^2_1$ variables depend on the weights only through the sum of squares $\sum_i\sigma_{x,i}^4$ and the maximum $\max_i\sigma_{x,i}^2$. Whereas Ailon and Chazelle bounded $\sum_i\sigma_{x,i}^4$ crudely by $\ell\max_i\sigma_{x,i}^4$---forcing them to keep $\max_i\sigma_{x,i}^2=O(1)$---FHL bounded the two quantities separately, allowing rare rows to capture up to $1+1/\eps$ units of energy. Whenever $\eps n$ is not the active branch of the minimum, their required sparsity is $\Theta(\log m)$ multiplied by a factor that remains $O(1)$ for all $\eps=O(\log\log m/\log m)$. In that regime, $O(\log m)$ nonzeros per row suffice, matching the flatness level itself and shaving a factor of $\log m$ off Theorem~\ref{thm:ac09}. Corollary~\ref{cor:gg-classical}(i) recovers this sharp bound with explicit constants under both Bernoulli and balanced supports.

Moreover, \citet[Thm.~2]{fandina2023fast} proved a matching lower bound showing that for a single preconditioned vector and failure probability $m^{-\Theta(1)}$, the standard estimator $\ell^{-1}\|\Phi x\|^2$ genuinely requires the row density of Theorem~\ref{thm:fhl23}; consequently, no analysis of the sample average that relies on a union bound over individual vectors can improve upon it.

\paragraph{Why sign quantization changes the picture.}
Crucially, both linear FJLT analyses rely on the fact that the estimator is unbiased conditionally on the supports: given $\capset_i$, the squared projection $(\Phi x)_i^2$ has conditional mean $\sigma_{x,i}^2$, and $\E\sigma_{x,i}^2=1$ for every fixed $D$. Random fluctuations in $\sigma_{x,i}^2$ therefore thicken the tails of the estimate, requiring more rows or slightly denser supports to control the variance, but they never shift its expectation.

Taking signs breaks this linearity. Consider again the self-pair example from Section~\ref{sec:why}, where a stored unit vector $u$ is scored against itself using BG. Conditioned on the support $\capset_i$, the row score $Z_i=\sqrt{\pi/2}\,|X_i|$ has conditional mean $\sigma_{u,i}$, where $\sigma_{u,i}^2$ is the energy captured from $u$. Taking expectations over the support for a fixed $D$ yields
\[
\E\hat\rho_{\rm BG}\ =\ \E\sigma_{u,i}\ \le\ \sqrt{\E\sigma_{u,i}^2}\ =\ 1,
\qquad\text{with a downward bias}\qquad
1-\E\sigma_{u,i}\ \approx\ \tfrac18\Var(\sigma_{u,i}^2).
\]
By Jensen's inequality and the strict concavity of the square root, the fluctuations of $\sigma_{u,i}^2$ around $1$ induce a systematic negative bias of order $\Var(\sigma_{u,i}^2)$, which is the exact calculation underlying \citet[Lemma~2]{ailon2009fast} for their $\ell_1$ FJLT.

The magnitude of this support bias is governed by the flatness of $\ut$. Under Bernoulli supports, $\Var(\sigma_{u,i}^2)=(1-q)\,k_u/r\le w_u/r$, and Lemma~\ref{lem:flat} bounds the cap $w_u$ by $\mu(K)$. Thus, while a linear sketch only needs flatness to control the tail probabilities of the captured energy, a one-bit sketch must keep a systematic bias of order $\mu(K)/r$ below the target accuracy $\eps$. This forces the row sparsity $r$ to scale as $\mu(K)/\eps$, explaining the $\eps^{-1}$ sparsity factor that appears throughout our one-bit bounds---as well as in the $\ell_1$ FJLT, where \citet[Thm.~5.1(i)]{matousek2008variants} similarly requires $r=2w_u/\eps$ in our notation.

\section{Main results}\label{sec:results}

We analyze each of the three estimators by following the two-stage structure introduced in Section~\ref{sec:fjlt}. First, we condition on a fixed sign matrix $D$ and bound the estimation error over the random draw of the sparse matrix $P$ alone, expressing the bound in terms of the empirical flatness parameters of the preconditioned inputs. Evaluated on a specific dataset, this conditional guarantee serves directly as an accuracy certificate in the sense of Section~\ref{sec:practice}. Second, we take expectations over the random draw of $D$, invoke Lemma~\ref{lem:flat} to bound the flatness parameters of pre-specified inputs by the flatness level $\mu(K)$ of \eqref{eq:mu-def} with high probability, and state the resulting worst-case requirements on the sketch length $\ell$ and row sparsity $r$ in classical form.

\subsection{Setting}\label{sec:setting}

Whereas Bernoulli supports (Definition~\ref{def:fjlt}) allow the number of nonzeros in each row to fluctuate randomly, Algorithm~\ref{alg:sketch} fixes the support size of every row to $r$.

\needspace{5\baselineskip}%
\begin{definition}[Balanced supports]\label{def:balanced}
In the setting of Definition~\ref{def:fjlt}, let $r\in\{1,\dots,n\}$ and $q:=r/n$. Under \emph{balanced supports}, each row support $\capset_i$ is drawn uniformly at random from the $\binom nr$ subsets of $\{1,\dots,n\}$ of size $r$. To treat both support laws simultaneously, define $f_q:=(n-r)/(n-1)$ and $\bal:=1$ for balanced supports, and $f_q:=1-q$ and $\bal:=0$ for Bernoulli supports (for any $q\in(0,1]$). In both cases $f_q\in[0,1]$, with $f_q=0$ if and only if $r=n$.
\end{definition}

All of our upper bounds depend on the support law exclusively through the two constants $f_q$ and $\bal$. When $r\ll n$, the factor $f_q$ is close to $1$; under balanced supports, $f_q$ is the standard finite-population correction for sampling without replacement.

\paragraph{Data-dependent flatness parameters.}
Apart from the true inner product $\rho$, our conditional bounds depend on the input vectors solely through summary statistics that measure how uniformly the preconditioned vectors $\xt=HDx$ spread their energy across coordinates. All parameters are defined for unit vectors (general nonzero vectors are normalized prior to evaluation). Alongside the cap $w_x$ and fourth moment $k_x$ introduced in Section~\ref{sec:why}, we define the \emph{cross fourth moment} $k_{uv}$ of a pair $(u,v)$, which measures the coordinatewise overlap of their squared entries, as well as the pairwise maxima $\wmax$ and $\kmax$. Writing $x\circ y$ for the Hadamard (entrywise) product with coordinates $(x\circ y)_j=x_jy_j$, we set
\begin{equation}\label{eq:params}
\begin{gathered}
w_x:=n\,\|\xt\|_\infty^2,\qquad k_x:=n\,\|\xt\|_4^4,\qquad 1\le k_x\le w_x\le n;\\
k_{uv}:=n\,\|\ut\circ\vt\|^2\le\min\{w_u,\sqrt{k_uk_v}\},\qquad
\wmax:=\max\{w_u,w_v\},\qquad \kmax:=\max\{k_u,k_v\};\\
w_\Ufam:=\max_{x\in\Ufam}w_x,\qquad k_\Ufam:=\max_{x\in\Ufam}k_x\qquad\text{for a finite family $\Ufam$ of unit vectors}.
\end{gathered}
\end{equation}
Notice that $k_{uu}=k_u$, and that $w_x=k_x=1$ if and only if $\xt$ is perfectly flat. The inequalities in \eqref{eq:params} are proved in Appendix~\ref{app:devices}.

\paragraph{Moments of the captured energy.}
Recall from Section~\ref{sec:why} that for a row support $\capset$, the rescaled subsampled vector $\xt_\capset\in\R^n$ has coordinates $(\xt_\capset)_j:=q^{-1/2}\xt_j\,\mathbf 1\{j\in\capset\}$. Caps and fourth moments govern the behavior of the energy $\sigma_x^2:=\|\xt_\capset\|^2$ captured from a unit vector $x$. Conditioned on $D$, the random variable $\sigma_x^2$ satisfies three key properties over the draw of $\capset$ (proved in Appendix~\ref{app:devices}):
\begin{equation}\label{eq:energy-facts}
\E\sigma_x^2=1\qquad\text{and}\qquad \Var(\sigma_x^2)=\frac{f_q\,(k_x-\bal)}r,
\end{equation}
while no individual coordinate contributes more than $w_x/r$ to $\sigma_x^2$.

Crucially, the fourth moment and the cap live on very different scales. For any unit vector $x$ fixed prior to drawing $D$, the expected fourth moment over the random signs is $\E_Dk_x=3-2\|x\|_4^4\le3$, regardless of the ambient dimension $n$. By contrast, the cap $w_x$ is the maximum of $n$ squared coordinates; it typically grows as $\log n$, and Lemma~\ref{lem:flat} bounds it in the worst case by the flatness level $\mu(K)$ of \eqref{eq:mu-def}. Because $k_x$ is so much smaller than $w_x$ on typical inputs, we state several of our main bounds in two forms: a simple cap-based bound and a sharper fourth-moment bound.

\paragraph{The statistical benefit of balanced rows.}
Under Bernoulli supports, the variance of the captured energy decomposes into two distinct contributions, $\Var(\sigma_x^2)=\frac{1-q}r+\frac{(1-q)\,(k_x-1)}r$: the first term arises from random fluctuations in the support size $|\capset|$, while the second reflects the heterogeneity of the squared coordinates $\xt_j^2$. Switching to balanced supports eliminates the support-size variance entirely and scales the coordinate-heterogeneity term by $n/(n-1)$. Whenever $k_x\approx 3$ (as holds for typical preconditioned vectors), balancing removes roughly one third of $\Var(\sigma_x^2)$; when $\xt$ is completely flat ($k_x=1$), it reduces the variance to zero, making BG completely unbiased (Proposition~\ref{prop:bg-support}(iii)). For SimHash (BB), however, balancing has virtually no effect on the bias because fluctuating row lengths merely rescale the row metric without changing angles (Section~\ref{sec:bb}).

\paragraph{Computational cost.}
Encoding a vector requires $n\log_2n$ additions and $n$ sign flips to apply $HD$, followed by $r\ell$ multiply--adds to apply $P$ (in expectation under Bernoulli supports, and deterministically under balanced supports). Scoring a pair of codes takes $O(\ell)$ operations (a single bitwise XOR and population count for BB), so scoring one query against $m$ stored vectors takes $O(m\ell)$ operations for any $n$ and $r$.

\paragraph{Admissible inputs and query adaptivity.}
Our conditional theorems fix the preconditioner $D$ and take probabilities solely over the sparse matrix $P$. Consequently, they apply to any vectors or pairs chosen without knowledge of $P$, even if that choice depends on $D$. Formally, we call a collection of vectors or pairs \emph{admissible} if it is conditionally independent of $P$ given $D$. This covers standard retrieval pipelines in which a database is encoded once and queried later, though it excludes adaptive queries chosen in response to previously returned sketch scores. When we pass to our classical corollaries by drawing $D$ at random, invoking Lemma~\ref{lem:flat} requires the vectors to be fixed independently of $D$ as well. Notably, in Corollaries~\ref{cor:gg-classical}(ii) and~\ref{cor:bg-classical}(ii), only the $m$ \emph{stored} database vectors need to be fixed prior to drawing $D$: because the underlying conditional bounds depend solely on the cap $w_u$ of the stored vector, the queries $v$ may depend arbitrarily on $D$, and the flatness level is $\mu(m)$ rather than $\mu(2N)$.

\paragraph{Alternative preconditioners.}
Because our conditional theorems depend on the preconditioner $HD$ only through the transformed vectors $\xt$, they hold verbatim when $HD$ is replaced by any other orthogonal transformation---or omitted altogether. In particular, if $P$ is applied directly to raw inputs without preconditioning, then $\xt=x$ and our bounds hold with the caps and fourth moments of the original vectors.

\subsection{What a sparse row sees}\label{sec:random-metric}

We now formalize the geometric picture of Section~\ref{sec:why}: conditioned on its support, a single sparse row acts on any pair of vectors through a random $2\times2$ metric matrix, and our three estimators correspond to three distinct functionals of this matrix.

Fix $D$ and a pair of unit vectors $u,v\in\R^n$ with inner product $\rho=\cos\theta$ ($\theta\in[0,\pi]$). Only the two-dimensional subspace spanned by $u$ and $v$ plays a role. Set $u_1:=u$ and $u_2:=(v-\rho u)/\sqrt{1-\rho^2}$ (or any fixed unit vector orthogonal to $u$ when $|\rho|=1$), so that the columns of $U:=[\,u_1\ \ u_2\,]\in\R^{n\times2}$ form an orthonormal basis for the plane and $v=\cos\theta\,u_1+\sin\theta\,u_2$. In this basis, $u$ and $v$ have local coordinate vectors $e_1:=(1,0)^\top$ and $t_\theta:=(\cos\theta,\sin\theta)^\top$, meaning $u=Ue_1$ and $v=Ut_\theta$. Applying the preconditioner maps $U$ to $\tilde U:=HDU=[\,\ut_1\ \ \ut_2\,]\in\R^{n\times2}$ with rows $\cvec_j^\top=(\ut_{1,j},\ut_{2,j})$. Because $HD$ is orthogonal, $\tilde U^\top\tilde U=\sum_j\cvec_j\cvec_j^\top=I_2$.

Now examine a single row of $\Phi$ with support $\capset$ and Gaussian weights $g_j\sim\mathcal N(0,1/q)$ for $j\in\capset$, and let $\tilde U_\capset\in\R^{n\times2}$ denote the subsampled frame with rows $q^{-1/2}\mathbf 1\{j\in\capset\}\cvec_j^\top$. Any vector $x=Us$ in the plane ($s\in\R^2$) has preconditioned image $\xt=\tilde Us$, subsampled image $\xt_\capset=\tilde U_\capset s$, and scalar row projection $\sum_{j\in\capset}g_j\,\cvec_j^\top s$. Conditioned on $\capset$, the projections of any two plane vectors with local coordinates $s$ and $s'$ are therefore jointly centered Gaussian with covariance $s^\top M_\capset\,s'$, where
\begin{equation}\label{eq:random-metric}
M_\capset:=\tilde U_\capset^\top\tilde U_\capset=\frac1q\sum_{j\in\capset}\cvec_j\cvec_j^\top=\begin{pmatrix}\sigma_1^2&\sigma_{12}\\\sigma_{12}&\sigma_2^2\end{pmatrix},\qquad
\sigma_{uv}=e_1^\top M_\capset\,t_\theta,\qquad \sigma_v^2=t_\theta^\top M_\capset\,t_\theta .
\end{equation}
Whereas a dense Gaussian row induces the Euclidean covariance $s^\top s'$, a sparse row measures inner products in the plane with respect to $M_\capset$, which we call the \emph{random metric} of the row. Its diagonal entries $\sigma_1^2=\sigma_u^2=\|\ut_\capset\|^2$ and $\sigma_2^2=\|\ut_{2,\capset}\|^2$ are the energies captured from $u_1$ and $u_2$, and its off-diagonal entry $\sigma_{12}=\inner{\ut_{1,\capset}}{\ut_{2,\capset}}$ is their captured inner product. For the pair $(u,v)$, the conditional covariance matrix of $(X,Y)$ is $\Sigma_\capset=(e_1,t_\theta)^\top M_\capset(e_1,t_\theta)=\begin{psmallmatrix}\sigma_u^2&\sigma_{uv}\\\sigma_{uv}&\sigma_v^2\end{psmallmatrix}$, where $\sigma_{uv}=\rho\sigma_1^2+\sqrt{1-\rho^2}\,\sigma_{12}$ and $\sigma_v^2=\|\vt_\capset\|^2$. Finally, for any positive definite matrix $M\succ 0$, let $\vartheta_\theta(M)\in[0,\pi]$ denote the angle between $e_1$ and $t_\theta$ under the inner product $(s,s')\mapsto s^\top Ms'$:
\[
\cos\vartheta_\theta(M)=\frac{e_1^\top Mt_\theta}{\{(e_1^\top Me_1)(t_\theta^\top Mt_\theta)\}^{1/2}} ,\qquad\text{so that}\quad \vartheta_\theta(I_2)=\theta .
\]

In the lemma below, $X$ and $Y$ denote the projections of $u$ and $v$ produced by a single row, and $Z=\sqrt{\pi/2}\,\sign(X)\,Y$ denotes the corresponding BG score.

\begin{lemma}[Random-metric reduction]\label{lem:random-metric}
Fix $D$ and a pair $(u,v)$, condition on the support $\capset$ of a single row, and adopt the convention $\signzero$. Under either support law:
\begin{enumerate}
\item[(i)] $(X,Y)$ is centered bivariate normal with covariance $\Sigma_\capset=\begin{pmatrix}\sigma_u^2&\sigma_{uv}\\\sigma_{uv}&\sigma_v^2\end{pmatrix}$. Equivalently, $(X,Y)$ has the same conditional distribution as $(\inner g{\ut_\capset},\inner g{\vt_\capset})$ for $g\sim\mathcal N(0,I_n)$; that is, the sparse row acts on $(u,v)$ exactly as a dense Gaussian row acts on $(\ut_\capset,\vt_\capset)$;
\item[(ii)] taking expectations over $\capset$ yields $\E\sigma_u^2=\E\sigma_v^2=\E\sigma_2^2=1$, $\E\sigma_{uv}=\rho$, and $\E\sigma_{12}=0$, or in matrix form, $\E M_\capset=I_2$;
\item[(iii)] $\E[XY\mid\capset]=\sigma_{uv}$; $\E[Z\mid\capset]=\zeta_\capset:=\sigma_{uv}/\sigma_u$ (with $\zeta_\capset:=0$ if $\sigma_u=0$); and $\Pr[\sign X\ne\sign Y\mid\capset]=\theta_\capset/\pi$, where $\theta_\capset:=\arccos\rho_\capset$ with $\rho_\capset:=\sigma_{uv}/(\sigma_u\sigma_v)$ when $\sigma_u\sigma_v>0$, $\theta_\capset:=\pi/2$ when exactly one of $\sigma_u,\sigma_v$ vanishes, and $\theta_\capset:=0$ when both vanish; in particular, $\theta_\capset=\vartheta_\theta(M_\capset)$ whenever $M_\capset\succ0$;
\item[(iv)] (Gram identity) $\sigma_u^2\sigma_v^2-\sigma_{uv}^2=(1-\rho^2)\det M_\capset$; consequently $|\sigma_{uv}|\le\sigma_u\sigma_v$, and for every $M\succ0$, $\sin\vartheta_\theta(M)=\sin\theta\,\sqrt{\det M/\{(e_1^\top Me_1)(t_\theta^\top Mt_\theta)\}}\le\sin\theta\sqrt{\lambda_{\max}(M)/\lambda_{\min}(M)}$.
\end{enumerate}
\end{lemma}

Lemma~\ref{lem:random-metric}, proved in Appendix~\ref{app:devices}, reduces the conditional analysis of a sparse row directly to the dense Gaussian identities \eqref{eq:dense-identities} applied to the subsampled vectors $(\ut_\capset,\vt_\capset)$ of Figure~\ref{fig:metric}. When $u=v=x$, part~(i) specializes to the familiar fact $(\Phi x)_i\sim\mathcal N(0,\sigma_{x,i}^2)$ that underpins the classical FJLT analysis. Because GG reads $\sigma_{uv}$, which is linear in $M_\capset$, part~(ii) implies that GG is unbiased for all inputs and all sparsities. By contrast, BG reads $\zeta_\capset$ and BB reads $\theta_\capset$. Both quantities are nonlinear functions of $M_\capset$ that evaluate to the target values $\rho$ and $\theta$ at $M_\capset=I_2$; their first-order Taylor terms around $I_2$ therefore average to zero by part~(ii), leaving a second-order support bias governed by the variance of $M_\capset-I_2$. Moreover, part~(iv) explains why SimHash preserves small angles to high relative accuracy: the sine of the distorted angle $\theta_\capset$ is proportional to $\sin\theta$ times a metric distortion factor that stays close to $1$ whenever $M_\capset\approx I_2$, causing the angular error to vanish proportionally as $\theta\to 0$.

\paragraph{Flatness parameters of a plane.}
Because SimHash (BB) depends on the entire metric $M_\capset$ rather than the energy of a single vector, its error bounds are governed by flatness parameters of the two-dimensional plane $\mathrm{span}\{u_1,u_2\}$. The squared row norm $\|\cvec_j\|^2=\ut_{1,j}^2+\ut_{2,j}^2$ represents the share of the plane's total energy carried by coordinate $j$ (the leverage score of row $j$), with $\sum_{j=1}^n\|\cvec_j\|^2=2$. In direct analogy with $w_x$ and $k_x$ for a single vector, we define the \emph{plane cap} $\wU$ and \emph{plane fourth moment} $\kU$ as
\begin{equation}\label{eq:frame}
\wU:=n\max_j\|\cvec_j\|^2=\max\{w_x:\ x\in{\rm span}\{u_1,u_2\},\ \|x\|=1\},\qquad
\kU:=n\sum_j\|\cvec_j\|^4 .
\end{equation}
In the terminology of matrix completion, $\wU/2$ is the coherence of the preconditioned subspace, while $\frac38\kU$ equals the average of $k_x$ over the unit circle in the plane. Both quantities depend only on the subspace $\mathrm{span}\{u_1,u_2\}$ (which coincides with $\mathrm{span}\{u,v\}$ whenever $|\rho|<1$) rather than on the choice of orthonormal basis $U$, and they satisfy $\wU\ge2$, $4\le\kU\le2\wU$, $\wU\le w_{u_1}+w_{u_2}$, and $\kU\le2(k_{u_1}+k_{u_2})$ (Appendix~\ref{app:devices}). For any plane fixed independently of $D$, averaging $k_x$ around the unit circle yields $\E_D\kU\le8$, whereas the plane cap $\wU$---like a vector cap---can grow as $\log n$ and is bounded in the worst case by $2\,\mu(2N)$. Importantly, two vectors that are individually nearly flat can still span a spiky plane: Theorem~\ref{thm:lower}(a) constructs pairs with $k_u,k_v\le1.01$ whose plane has a large fourth moment $\kU$ and incurs a relative BB bias of order $\kU/r$.

\subsection{Full precision on both sides (GG)}\label{sec:gg}

We begin with the full-precision linear estimator GG, which isolates the effect of row sparsity in the absence of sign quantization. Conditioned on $D$, each row product $X_iY_i$ is unbiased at every sparsity and has variance
\begin{equation}\label{eq:gg-moments}
\E[X_iY_i\mid D]=\rho,\qquad \Var(X_iY_i\mid D)=1+\rho^2+\frac{f_q}{r}\big(3k_{uv}-\bal(1+2\rho^2)\big)\le1+\rho^2+\frac{3f_q\,k_{uv}}{r},
\end{equation}
which specializes at $v=u$ to $\Var(X_i^2\mid D)=2+3\Var(\sigma_u^2)$ (derived in Appendix~\ref{app:gg}). Because sparse sampling inflates the variance and tails without shifting the expectation, Theorem~\ref{thm:gg-param} bounds the resulting error both for the standard sample average of squared projections (the classical FJLT norm estimator) and for inner products decoded via Catoni's robust mean.

\begin{theorem}[GG for a realized preconditioner]\label{thm:gg-param}
Fix any $D$ and $\delta\in(0,1)$, and draw $P$ with balanced supports (the statement holds verbatim for Bernoulli supports). Let $h(t):=(1+t)\log(1+t)-t$.
\begin{enumerate}
\item[(a)] \emph{Norms.} Let $\Ufam$ be an admissible finite set of unit vectors and $\eps\in(0,1]$. If
\[
\ell\ \ge\ \frac{12}{\eps^{2}}\log\frac{3|\Ufam|}{\delta}
\qquad\text{and}\qquad
r\ \ge\ \max\Big\{\,8f_q\,k_\Ufam,\ \ \min\Big\{\frac{\eps n}{1+\eps},\ \frac{w_\Ufam\,\log(3\ell|\Ufam|/\delta)}{h(1/\eps)}\Big\}\Big\},
\]
then with probability at least $1-\delta$ over $P$, we have $\big|\ell^{-1}\|\Phi x\|^2-1\big|\le\eps$ for all $x\in\Ufam$.
\item[(b)] \emph{Robust inner products.} Let $N$ admissible pairs of unit vectors be given, and let $L:=\log(2N/\delta)<\ell/2$ and $\sigma_*^2\ge\max_{\rm pairs}\{1+\rho^2+3f_q\,k_{uv}/r\}$. With probability at least $1-\delta$ over $P$, Catoni's estimate $\Cat_\alpha$ of \eqref{eq:catoni-def} satisfies, for all pairs,
\[
\big|\Cat_\alpha(X_1Y_1,\dots,X_\ell Y_\ell)-\rho\big|\le s:=\sqrt{\frac{2\sigma_*^2L}{\ell-2L}},\qquad\alpha=\frac{s}{\sigma_*^2+s^2}.
\]
Since $k_{uv}\le w_u$, one may take $\sigma_*^2=2+3\max w_u/r$, which involves only the first vector $u$ of each pair $(u,v)$ and is at most $3$ once $w_u\le r/3$.
\end{enumerate}
\end{theorem}

In part~(a), the sketch-length requirement on $\ell$ matches that of a dense Gaussian projection (Theorem~\ref{thm:dense-gg}) up to the constant inside the logarithm. The sparsity requirement on $r$ takes the maximum of two conditions that play distinct roles in the proof. The first branch, $r\ge8f_q\,k_\Ufam$, is a \emph{variance condition}: by \eqref{eq:energy-facts}, it ensures that $\Var(\sigma_x^2)\le\frac18$ for every $x\in\Ufam$, which is needed to control both tails of $\|\Phi x\|^2$. The second branch is a \emph{capture condition} needed only for the upper tail: it guarantees with high probability that no row captures more than $1+1/\eps$ times its expected share of energy from any vector in $\Ufam$. This bound holds deterministically whenever $r\ge\eps n/(1+\eps)$ (since $\sigma_x^2\le n/r$ always), and otherwise follows from Bennett's inequality. Notice that $\ell$ enters the capture condition only logarithmically via a union bound over the $\ell$ rows.

Using the elementary inequality $h(t)\ge\frac t2\log(1+t)$ and $k_\Ufam\le w_\Ufam$, both sparsity conditions are satisfied whenever
\[
r\ \ge\ w_\Ufam\max\Big\{8,\ \frac{2\eps\log(3\ell|\Ufam|/\delta)}{\log(1+1/\eps)}\Big\}.
\]
As $\eps\to 0$, the second branch inside the maximum vanishes, leaving the $\eps$-independent requirement $r\ge 8w_\Ufam$. Thus, when no signs are taken, higher accuracy requires more rows rather than denser rows.

Part~(b) requires even less sparsity because Catoni's estimator depends only on the variance of a single row product $X_iY_i$, which equals $1+\rho^2$ for a dense row and increases by at most $3f_q\,k_{uv}/r$ under sparse sampling. Once $r\ge3\max w_u$, a sketch of length $\ell=(6\eps^{-2}+2)L$ achieves error at most $\eps$: the required sparsity is completely independent of $\eps$ across the entire range $\eps\in(0,1]$, and depends only on the flatness of the stored vector $u$ in each pair.

When $D$ is drawn at random, Lemma~\ref{lem:flat} bounds every cap (and hence every fourth moment) by the flatness level $\mu(K)=2\log(4nK/\delta)$ of \eqref{eq:mu-def} with high probability, converting Theorem~\ref{thm:gg-param} into its classical worst-case form.

\needspace{6\baselineskip}%
\begin{corollary}[GG, classical form]\label{cor:gg-classical}
Use either support law, let $\delta\in(0,1)$, and recall $\mu(K)=2\log(4nK/\delta)$ from \eqref{eq:mu-def}.
\begin{enumerate}
\item[(i)] \emph{Norms, embeddings and inner products.} Let $\Ufam$ be a finite set of unit vectors fixed independently of $(D,P)$ and let $\eps\in(0,1]$. If
\[
\begin{gathered}
\ell\ \ge\ \frac{12}{\eps^{2}}\log\frac{4|\Ufam|}{\delta},\\
r\ \ge\ \max\Big\{\min\{n,\,8\,\mu(2|\Ufam|)\},\ \ \min\Big\{\frac{\eps n}{1+\eps},\ \frac{2\eps\,\mu(2|\Ufam|)\log(4\ell|\Ufam|/\delta)}{\log(1+1/\eps)}\Big\}\Big\},
\end{gathered}
\]
then with probability at least $1-\delta$, we have $|\ell^{-1}\|\Phi x\|^2-1|\le\eps$ for all $x\in\Ufam$. Taking $\Ufam$ to be the normalized differences of $m$ points ($|\Ufam|<m^2/2$) gives a $(1\pm\eps)$ embedding of their squared distances; taking $\Ufam$ to be the normalized $u\pm v$ of $N$ pairs of unit vectors ($|\Ufam|\le2N$) gives $|\hat\rho_{\rm GG}-\rho|\le\eps$ for all pairs.
\item[(ii)] \emph{Robust inner products, $D$-adaptive queries.} Let $\eps\in(0,1]$, let $m$ unit vectors be fixed independently of $(D,P)$, and let $N$ admissible pairs $(u,v)$ of unit vectors have $u$ among the fixed vectors. If $r\ge\min\{n,3\,\mu(m)\}$ and $\ell\ge8\eps^{-2}\log(4N/\delta)$, then with probability at least $1-\delta$, we have $|\Cat_\alpha(X_1Y_1,\dots,X_\ell Y_\ell)-\rho|\le\eps$ for all pairs, with $\alpha=\eps/(3+\eps^2)$.
\end{enumerate}
At the smallest $(\ell,r)$ allowed, encoding costs $O\big(n\log n+\allowbreak\eps^{-2}\log\tfrac{4|\Ufam|}\delta\cdot\allowbreak\mu(2|\Ufam|)\allowbreak\max\{1,\allowbreak\eps\,\mu(2|\Ufam|)/\log\tfrac2\eps\}\big)$ in (i) and $O\big(n\log n+\eps^{-2}\log\tfrac{4N}\delta\cdot\mu(m)\big)$ in (ii).
\end{corollary}

In part~(i), the two branches of the maximum correspond directly to the variance and capture conditions of Theorem~\ref{thm:gg-param}(a) with $\mu(2|\Ufam|)$ replacing the empirical parameters; both hold whenever $r\ge\min\big\{n,\ \mu(2|\Ufam|)\max\big\{8,\ \frac{2\eps\log(4\ell|\Ufam|/\delta)}{\log(1+1/\eps)}\big\}\big\}$, which is the expression reported in Table~\ref{tab:compare}.

\begin{remark}[Comparison with Theorems~\ref{thm:ac09} and~\ref{thm:fhl23}]\label{rem:gg-compare}
Theorem~\ref{thm:fhl23} of \citet{fandina2023fast} takes the minimum of two sparsity regimes: $\Theta(\eps n)$ and $\Theta(\log m\cdot\max\{1,\allowbreak\eps\log m/\log(1/\eps)\})$. Corollary~\ref{cor:gg-classical}(i) provides the exact same two-branch structure with explicit constants, arbitrary failure probability $\delta$, and either support law. Specifically, writing $\mu_1:=\mu(2|\Ufam|)$, at the minimum allowed sketch length $\ell$ our capture condition is satisfied either by $r\ge\eps n/(1+\eps)$ or by $r\ge4\eps\mu_1+8\eps\mu_1^2/\log(2/\eps)$. Under the standing assumptions of FHL ($n\le m$, $\delta=1/m$, and $\ell\le n$), we have $\mu_1=\Theta(\log m)$ and $\log(2/\eps)\le\mu_1$ (since $\eps^{-2}\le\ell\le n$), so our second capture branch scales as $\eps\mu_1^2/\log(2/\eps)=O\big(\eps\log^2m/\log(1/\eps)\big)$, matching the second branch of FHL up to a constant factor. Meanwhile, our variance condition $r\ge 8\mu_1=\Theta(\log m)$ is dominated both by the second FHL branch and---since $\eps n\ge\eps\ell=\Omega(\eps^{-1}\log m)$---by a constant multiple of the first. Part~(i) therefore recovers the sharp sparsity of FHL up to constant factors, improving upon the $\Theta(\log^2m)$ bound of Theorem~\ref{thm:ac09} by a factor of $\min\{\log m,\log(2/\eps)/\eps\}$. In the general regime where $n$ or $1/\delta$ may exceed $\mathrm{poly}(m)$, the flatness level naturally scales as $\log(nm/\delta)$ rather than $\log m$.

While balanced row supports with Rademacher entries were previously analyzed by \citet[Thm.~3.1]{wolff2012randomness} at sparsity $O(\log(\ell/\delta)\log(n/\delta))$---which reduces to $O(\log^2m)$ for $m$ points---Corollary~\ref{cor:gg-classical}(i) is, to our knowledge, the first result for balanced supports to achieve the optimal sparsity of Theorem~\ref{thm:fhl23} (see Section~\ref{sec:related}).

Furthermore, switching from the sample average to Catoni's robust decoder bypasses the FHL capture condition altogether. Corollary~\ref{cor:gg-classical}(ii) requires only $r\ge\min\{n,3\,\mu(m)\}$ for inner products regardless of $\eps$, and an analogous scale-free Catoni estimator in Appendix~\ref{app:gg} achieves the $\eps$-independent sparsity $r\ge\min\{n,3\,\mu(m^2/2)\}$ for preserving all pairwise squared distances among $m$ points whenever $\eps\le\frac12$. In the moderate-$\eps$ regime where $\eps\log m\gg\log(1/\eps)$, this sparsity is strictly smaller than the bound of Theorem~\ref{thm:fhl23}---without contradicting the single-vector lower bound of \citet[Thm.~2]{fandina2023fast}, which applies specifically to the non-robust sample average $\ell^{-1}\|\Phi x\|^2$.
\end{remark}

\subsection{One bit against full precision (BG)}\label{sec:bg}

We now turn to the asymmetric estimator BG, in which the stored vector $u$ is quantized to one bit while the query $v$ is kept at full precision. Conditioned on the support $\capset$, a single row score $Z$ has expectation $\zeta_\capset=\sigma_{uv}/\sigma_u$ (Lemma~\ref{lem:random-metric}(iii)), whose expectation over $\capset$ deviates systematically from $\rho$. Expanding $\zeta_\capset-\rho=\rho(\sigma_u-1)+(\sigma_{uv}-\rho\sigma_u^2)/\sigma_u$ to second order around $\sigma_u^2=1$ yields the leading-order approximation
\begin{equation}\label{eq:bg-second-order}
b_{uv}\ \approx\ -\frac\rho8\,\Var(\sigma_u^2)\ -\ \frac12\,\Cov(\sigma_{uv}-\rho\sigma_u^2,\,\sigma_u^2),
\qquad
\Cov(\sigma_{uv}-\rho\sigma_u^2,\,\sigma_u^2)=\frac{f_q}{q}\sum_{j=1}^n\ut_j^3(\vt_j-\rho\ut_j),
\end{equation}
which separates the self-pair energy deficit $-\frac\rho8\Var(\sigma_u^2)$ from the cross-correlation between energy fluctuations and angular distortion (Appendix~\ref{app:bg}). To obtain rigorous finite-sample guarantees, we first bound $b_{uv}$ non-asymptotically---depending on $D$ and $(u,v)$ but not on $\ell$---and then combine it with concentration bounds for the sampling noise across the $\ell$ rows.

\begin{proposition}[BG support bias]\label{prop:bg-support}
Fix $D$ and unit vectors $u$ (stored) and $v$ (query), under either support law. Let $b_{uv}:=\E[\hat\rho_{\rm BG}\mid D]-\rho$. Then
\begin{enumerate}
\item[(i)] $|b_{uv}|\le1.15\,w_u/r$ if $w_u\le r/64$ (and $|b_{uv}|\le4.3\,w_u/r$ unconditionally);
\item[(ii)] $|b_{uv}|\le1.83\,f_q\sqrt{(k_u-\bal)\,\kmax}\,/r+1.12\,e^{-r/(16k_u)}$, which is at most $2.83\,\kmax/r$ if $\kmax\le r/70$;
\item[(iii)] under balanced supports, $b_{uv}=0$ whenever $k_u=1$.
\end{enumerate}
\end{proposition}

Each part of Proposition~\ref{prop:bg-support} serves a distinct purpose. Part~(i) depends solely on the cap $w_u$ of the stored vector, making it valid uniformly across all unit queries $v$, including queries chosen adaptively after seeing $D$. Part~(ii) replaces caps with fourth moments at the expense of an exponentially small tail term $1.12\,e^{-r/(16k_u)}$ that accounts for rare supports capturing very little energy from $u$. Part~(iii) highlights the exact unbiasedness of balanced supports on flat stored vectors, since $k_u=1$ forces $\sigma_u^2\equiv 1$ on every balanced support. We refer to bounds such as parts~(i) and~(ii), which depend on the data only through summary caps and fourth moments, as \emph{parameter forms}, and to their single-term specializations $1.15\,w_u/r$ and $2.83\,\kmax/r$ as \emph{closed forms}. Both are derived in Appendix~\ref{app:bg} from a sharper coordinatewise bound---the \emph{pair certificate}---which retains the exact support variances $V_u=\Var(\sigma_u^2)$ and $V_{uv}=\Var(\sigma_{uv}-\rho\sigma_u^2)=\frac{f_q}{q}\|\ut\circ(\vt-\rho\ut)\|^2$ before relaxing them to caps or fourth moments.

Controlling the row-to-row sampling noise is straightforward: conditioned on its support, a BG row score $Z$ is sub-Gaussian around its conditional mean $\zeta_\capset$, and because $\sign(X)^2=1$ gives $Z^2=\frac\pi2Y^2$ identically, its unconditional second moment is $\E[Z^2\mid D]=\frac\pi2\E\sigma_v^2=\pi/2$, matching that of a dense Gaussian row (Appendix~\ref{app:bg}). Combining Proposition~\ref{prop:bg-support} with tail bounds for the row average and for Catoni's estimator yields our main conditional guarantee for BG.

\begin{theorem}[BG for a realized preconditioner]\label{thm:bg-param}
Fix any $D$ and $\delta\in(0,1)$, and draw $P$ with balanced supports (the statement holds verbatim for Bernoulli supports). Let $N$ admissible pairs $(u,v)$ of unit vectors be given ($u$ stored, $v$ the query) and $L:=\log(2N/\delta)$.
\begin{enumerate}
\item[(a)] \emph{Dot product.} If $\wmax\le r/64$ and $\ell\ge4(\pi+18.5\,\wmax/r)\,L$ for every pair, then with probability at least $1-\delta$ over $P$, for all pairs,
\[
|\hat\rho_{\rm BG}-\rho|\ \le\ 1.15\,\frac{w_u}r+\sqrt{\Big(\pi+18.5\,\frac{\wmax}r\Big)\frac{L}{\ell}} .
\]
\item[(b)] \emph{Catoni decoder.} If $w_u\le r/64$ for every pair and $\ell>2L$, then with probability at least $1-\delta$ over $P$, for all pairs,
\[
|\Cat_\alpha(Z_1,\dots,Z_\ell)-\rho|\ \le\ 1.15\,\frac{w_u}r+\sqrt{\frac{\pi L}{\ell-2L}},\qquad
\alpha=\frac{s}{\pi/2+s^2},\ \ s=\sqrt{\frac{\pi L}{\ell-2L}} .
\]
\end{enumerate}
\end{theorem}

The central feature of Theorem~\ref{thm:bg-param} is the decomposition of the error into two parts. The first term in both bounds is the support bias from Proposition~\ref{prop:bg-support}(i); because it does not decay with $\ell$, driving the total error below $\eps$ requires the row sparsity $r$ to grow as $1/\eps$, a dependence that Theorem~\ref{thm:lower}(b) proves is unavoidable. Whenever fourth moments are available, one may substitute the sharper bias bound of Proposition~\ref{prop:bg-support}(ii) (such as $2.83\,\kmax/r$ for $\kmax\le r/70$) directly into the first term. With that substitution, part~(b) requires no cap condition at all, while part~(a) requires only the milder condition $\wmax\le r/16$, which increases the sub-Gaussian variance constant $\pi$ in the second term by at most $1.16$ (Appendix~\ref{app:bg}).

The second term in each bound captures the sampling noise across the $\ell$ rows, which remains very close to that of a dense sketch. In part~(a), the dense variance constant $\pi$ of Theorem~\ref{thm:dense-bg} is inflated by $18.5\,\wmax/r$ (at most $0.29$ when $\wmax\le r/64$), which is the only place where the query cap $w_v$ enters; the mild length condition $\ell\ge4(\pi+18.5\,\wmax/r)\,L$ (satisfied once $\ell\ge 14L$) simply ensures that the deviation stays within the sub-Gaussian regime ($\le 1/2$) of the moment generating function bound. In part~(b), the noise term is identical to that of Catoni's estimator on a dense sketch (Section~\ref{sec:dense}) and is completely independent of the query $v$.

Bounding the caps via Lemma~\ref{lem:flat} when $D$ is drawn at random and setting $r=\Theta(\mu(K)/\eps)$---so that the support bias consumes at most one fourteenth of the error budget $\eps$---yields our worst-case corollary for BG.

\begin{corollary}[BG, classical form]\label{cor:bg-classical}
Let $\eps\in(0,\tfrac12]$ and $\delta\in(0,1)$, use either support law, and recall $\mu(K)=2\log(4nK/\delta)$ from \eqref{eq:mu-def}.
\begin{enumerate}
\item[(i)] \emph{Dot product.} Let $\Ufam$ be a finite set of unit vectors and fix $N$ pairs in $\Ufam\times\Ufam$, all independently of $(D,P)$. If $r\ge\min\{n,\ 16\max\{\eps^{-1},4\}\,\mu(|\Ufam|)\}$ and $\ell\ge4\eps^{-2}\log(4N/\delta)$, then with probability at least $1-\delta$, we have $|\hat\rho_{\rm BG}-\rho|\le\eps$ for all pairs.
\item[(ii)] \emph{Catoni decoder, $D$-adaptive queries.} Let $m$ stored unit vectors be fixed independently of $(D,P)$, and let $N$ admissible pairs $(u,v)$ have $u$ stored and $v$ any unit vector. If $r\ge\min\{n,\ 16\max\{\eps^{-1},4\}\,\mu(m)\}$ and $\ell\ge(3.9\,\eps^{-2}+2)\log(4N/\delta)$, then with probability at least $1-\delta$, we have $|\Cat_\alpha(Z_1,\dots,Z_\ell)-\rho|\le\eps$ for all pairs, where $\alpha=0.9\eps/(\pi/2+0.81\eps^2)$.
\end{enumerate}
With $K=|\Ufam|$ in (i) and $K=m$ in (ii), the smallest sparsity allowed is at most $\lceil32\,\mu(K)/\eps\rceil$, so encoding at the smallest $(\ell,r)$ allowed costs $O\big(n\log n+\eps^{-3}\mu(K)\log\frac{4N}\delta\big)$.
\end{corollary}

\begin{remark}[The price of one bit]\label{rem:bg-main}
Quantizing the stored vector to one bit demands denser rows, but scarcely increases the sketch length: part~(i) uses roughly $1.3$ times the $\pi\eps^{-2}\log(2N/\delta)$ rows of dense BG (Theorem~\ref{thm:dense-bg}), and part~(ii) uses slightly fewer still. Unlike GG products $X_iY_i$, which are products of two Gaussians and can exhibit heavier tails when supports fluctuate, BG scores remain conditionally Gaussian because only one side is signed; consequently, robust aggregation saves very few rows in the worst case.

The true advantage of Catoni's decoder in BG is its complete insensitivity to the query vector. Because its noise bound relies solely on the unconditional second moment $\E[Z_i^2\mid D]=\pi/2$, which holds for every unit query $v$ regardless of $w_v$, part~(ii) imposes no flatness requirement on the query, pays $\mu(m)$ rather than $\mu(|\Ufam|)$, and seamlessly accommodates queries chosen adaptively after seeing $D$. Moreover, when we compute data-dependent certificates in Table~\ref{tab:budgets}, the certified sparsity for Catoni's decoder on a typical pair is roughly half that of the dot-product decoder. We emphasize that this gap is an artifact of our moment generating function bound rather than an intrinsic difference between the two decoders: both decoders experience the exact same per-row variance $\pi/2-(\E[Z\mid D])^2$, and the extra term $18.5\,\wmax/r$ in Theorem~\ref{thm:bg-param}(a) arises solely from the cap-based moment generating function bound in Appendix~\ref{app:bg}, which we did not attempt to optimize in terms of fourth moments (Appendix~\ref{app:k4}).

Finally, one can freely trade sketch length against row sparsity by partitioning the error budget as $\eps=\eps_s+\eps_b$ between sampling noise and support bias. Because Proposition~\ref{prop:bg-support}(i) guarantees $|b_{uv}|\le4.3\,w_u/r$ unconditionally (without requiring $w_u\le r/64$), part~(ii) generalizes to any split with $r\ge\min\{n,\ 4.3\,\mu(m)/\eps_b\}$, $\ell\ge(\pi\eps_s^{-2}+2)\log(4N/\delta)$, and $\alpha=\eps_s/(\pi/2+\eps_s^2)$. Choosing $\eps_b\ll\eps$ pushes the sketch length arbitrarily close to the dense BG bound at the expense of a proportionally larger row sparsity $r$.
\end{remark}

\begin{remark}[Fast evaluation of the Catoni decoder]\label{rem:catoni-fast}
Solving the root-finding equation \eqref{eq:catoni-def} exactly requires a few iterative passes over the $\ell$ row scores of each pair. However, a much simpler non-iterative variant suggested by \citet{minsker2018subgaussian} evaluates the influence function just once at the origin, yielding the closed-form estimator $(\alpha\ell)^{-1}\sum_i\psi(\alpha Z_i)$ (or, if desired, uses a single Newton step from the origin to approximate the exact root). Because this one-step estimator concentrates at a rate governed by the raw second moment $\E[Z_i^2\mid D]=\pi/2$ in place of the variance, setting $\alpha=\sqrt{4L/(\pi\ell)}$ achieves an error radius of $\sqrt{\pi L/\ell}$ rather than $\sqrt{\pi L/(\ell-2L)}$, eliminating the additive $+2$ term in Corollary~\ref{cor:bg-classical}(ii) (where $L=\log(4N/\delta)$). Better still, because $\psi$ is an odd function, the one-step estimate simplifies to $\ell^{-1}\sum_i\sign(X_i)\,\phi(Y_i)$ with $\phi(y):=\alpha^{-1}\psi(\alpha\sqrt{\pi/2}\,y)$. In other words, it is simply the standard BG dot product computed against query projections $Y_i$ that have been softly clipped once upon arrival! Full details are given in Appendix~\ref{app:tools}.
\end{remark}

\subsection{One bit on both sides (BB)}\label{sec:bb}

In SimHash (BB), both $u$ and $v$ are quantized to one bit. Conditioned on its support $\capset$, a sparse row produces a sign disagreement with probability $\theta_\capset/\pi$ (Lemma~\ref{lem:random-metric}(iii)), so the support bias of the angle estimate $\hat\theta$ is simply $\E\theta_\capset-\theta$. Theorem~\ref{thm:bb-param} bounds the total estimation error in terms of the plane parameters $(\kU,\wU)$ of \eqref{eq:frame}, both for additive error in part~(a) and for relative error in part~(b).

\begin{theorem}[BB for a realized preconditioner]\label{thm:bb-param}
Fix any $D$ and $\delta\in(0,1)$, and draw $P$ with balanced supports (the statement holds verbatim for Bernoulli supports). Let $N$ admissible pairs of unit vectors be given, with plane parameters $\kU,\wU$ as in \eqref{eq:frame}.
\begin{enumerate}
\item[(a)] \emph{Additive error.} With probability at least $1-\delta$ over $P$, for all pairs,
\[
|\hat\theta-\theta|\ \le\ \frac78\,\sin\theta\,\frac{f_q\,(\kU-2\bal)}{r}+4\pi\exp\Big(-\frac{(1-\log2)\,r}{2\wU}\Big)+\pi\sqrt{\frac{\log(2N/\delta)}{2\ell}},
\]
and the same bound holds for $|\hat\rho_{\rm BB}-\rho|$. If $\wU\le r/64$, the first two terms together are at most $1.8\,\wU/r$.
\item[(b)] \emph{Relative error.} Let $\eps\in(0,1]$ and $\theta_0\in(0,\pi]$. If $\ell\ge42\,\eps^{-2}\theta_0^{-1}\log(2N/\delta)$ and $r\ge\max\{3.5\,\kU/\eps,\ 6.52\,\wU\log(16\pi/(\eps\theta_0))\}$ for every pair, then with probability at least $1-\delta$ over $P$, every pair with $\theta\ge\theta_0$ has $|\hat\theta-\theta|\le\eps\theta$, and every pair with $\theta=0$ has $\hat\theta=0$.
\end{enumerate}
\end{theorem}

The additive error bound in part~(a) consists of three terms: the third is the exact Hoeffding sampling noise of dense SimHash (Theorem~\ref{thm:dense-bb}(a)), since conditioned on $D$ the $\ell$ rows still produce independent Bernoulli disagreements, while the first two terms bound the support bias.

Specifically, the first term bounds the bias contributed by \emph{typical supports}---those for which the random metric $M_\capset$ has eigenvalues between $\frac12$ and $2$, so that every unit vector in the plane captures between half and twice its expected energy. As predicted by the Gram identity (Lemma~\ref{lem:random-metric}(iv)), this leading bias term carries the factor $\sin\theta$ and therefore vanishes proportionally for nearly collinear vectors. Moreover, it depends on the plane solely through its fourth moment $\kU$, scaling as $O(1/r)$ rather than $O(\mu(2N)/r)$ for a typical draw of $D$. The factor $f_q(\kU-2\bal)/r$ is precisely the expected squared Frobenius distance $\E\|M_\capset-I_2\|_F^2$: to second order, the exact angular bias is a signed sum over the $n$ coordinates weighted by $\|\cvec_j\|^4$ whose summands oscillate with the orientation of $\cvec_j\in\R^2$ (causing near-complete cancellation on isotropic planes; Appendix~\ref{app:bb}), whereas bounding each directional derivative in Frobenius norm sums their magnitudes. Under Bernoulli supports, fluctuations in the row length $|\capset|$ add $2f_q/r$ to $\E\|M_\capset-I_2\|_F^2$; although a scalar multiple of $M_\capset$ does not change the angle $\theta_\capset$, our second-derivative bound charges for the full Frobenius norm, giving balanced supports a slightly smaller constant in the theorem.

The second term, which we call the \emph{cap tail}, accounts for atypical supports whose metric eigenvalues stray outside $[\frac12,2]$. It equals $\pi$ times a matrix Chernoff bound on the probability of such an event (Appendix~\ref{app:bb}, where we also derive a sharper coordinatewise pair certificate via matrix Bernstein) and is the only place where the plane cap $\wU$ enters. Because it decays exponentially in $r/\wU$, the cap tail acts as a sharp threshold: once $r$ exceeds a modest logarithmic multiple of $\wU$, its contribution becomes negligible, whereas for smaller $r$ it dominates the bound. When only $\wU$ is known, the final sentence of part~(a) combines both bias terms into the simple closed form $1.8\,\wU/r$, paralleling the bound $1.15\,w_u/r$ for BG.

To achieve a target additive accuracy $\eps$ in part~(a), we allocate a quarter of the error budget $\eps$ to each of the two bias terms and the remaining half to the sampling noise. This yields $|\hat\theta-\theta|\le\eps$ across all pairs whenever $r\ge\max\{3.5\,\kU/\eps,\ 6.52\,\wU\log(16\pi/\eps)\}$ and $\ell\ge2\pi^2\eps^{-2}\log(2N/\delta)$. As $\eps\to 0$, the first branch dominates and the required sparsity scales as $\kU/\eps$. At moderate accuracies, however, the logarithmic threshold of the cap tail is often the active constraint: in the practical regime of Table~\ref{tab:budgets} ($\eps=0.1$), the cap tail dictates the certified typical-pair sparsity $r=842$ under balanced supports, whereas the fourth-moment term alone would be satisfied at $r=104$ (Appendix~\ref{app:k4}). In the relative-error setting of part~(b), because the leading bias term already scales with $\sin\theta\le\theta$, the minimum angle $\theta_0$ enters the sparsity requirement only inside the logarithm of the cap-tail threshold.

Passing to the unconditional setting by drawing $D$ at random, Lemma~\ref{lem:flat} bounds the caps of all $2N$ basis vectors $(u_1,u_2)$ by $\mu(2N)=2\log(8nN/\delta)$ with probability at least $1-\delta/2$, which implies $\wU\le w_{u_1}+w_{u_2}\le 2\,\mu(2N)$ and $\kU\le 2\wU\le 4\,\mu(2N)$. (Bounding $\kU$ by $4\,\mu(2N)$ is conservative since $\E_D\kU\le 8$; establishing sharp high-probability tail bounds for fourth moments directly is an open problem discussed in Section~\ref{sec:limitations}.)

\begin{corollary}[BB, classical form]\label{cor:bb-classical}
Fix $N$ pairs of unit vectors independently of $(D,P)$, let $\delta\in(0,1)$, use either support law, and recall $\mu(2N)=2\log(8nN/\delta)$ from \eqref{eq:mu-def}.
\begin{enumerate}
\item[(i)] Let $\eps\in(0,\pi]$. If $r\ge\min\{n,\,\mu(2N)\max\{14/\eps,\,13.5\log(16\pi/\eps)\}\}$ and $\ell\ge2\pi^2\eps^{-2}\log(4N/\delta)$, then with probability at least $1-\delta$, we have $|\hat\theta-\theta|\le\eps$ and $|\hat\rho_{\rm BB}-\rho|\le\eps$ for all pairs.
\item[(ii)] Let $\eps\in(0,1]$ and $\theta_0\in(0,\pi]$. If $r\ge\min\{n,\,\mu(2N)\max\{14/\eps,\,13.5\log(16\pi/(\eps\theta_0))\}\}$ and $\ell\ge42\,\eps^{-2}\theta_0^{-1}\log(4N/\delta)$, then with probability at least $1-\delta$, every pair with $\theta\ge\theta_0$ has $|\hat\theta-\theta|\le\eps\theta$, and every pair with $\theta=0$ has $\hat\theta=0$.
\end{enumerate}
At the smallest $(\ell,r)$ allowed in (i), encoding costs $O\big(n\log n+\eps^{-3}\mu(2N)\log\frac{4N}\delta\big)$.
\end{corollary}

Here the two branches inside the maximum come directly from substituting $\kU\le 4\,\mu(2N)$ and $\wU\le 2\,\mu(2N)$ into the two bias terms of Theorem~\ref{thm:bb-param} (with $4/(1-\log2)\approx13.04$ rounded up to $13.5$); in part~(i), the $14/\eps$ branch dominates whenever $\eps\le0.185$.

\begin{remark}[Balancing sketch length, sparsity, and near-duplicate sensitivity]\label{rem:bb-main}
Because Corollary~\ref{cor:bb-classical}(i) assigns half of the error budget $\eps$ to the support bias and half to the sampling noise, its sketch length is four times that of dense SimHash (Theorem~\ref{thm:dense-bb}(a)). Just as in Remark~\ref{rem:bg-main}, we can trade row sparsity for a shorter sketch by allocating an arbitrary fraction $\eta\in(0,1)$ of the error budget $\eps$ to the two bias terms ($\eta\eps/2$ each) and $(1-\eta)\eps$ to the noise, which generalizes part~(i) to
\[
r\ge\min\Big\{n,\ \mu(2N)\max\Big\{\frac7{\eta\eps},\ 13.5\log\frac{8\pi}{\eta\eps}\Big\}\Big\}\qquad\text{and}\qquad
\ell\ge\frac{\pi^2}{2(1-\eta)^2\eps^2}\log\frac{4N}\delta .
\]
Setting $\eta=\frac12$ recovers Corollary~\ref{cor:bb-classical}(i), while letting $\eta\to0$ brings the sketch length arbitrarily close to the dense SimHash bound.

In part~(ii), the sketch length retains the favorable $\theta_0^{-1}$ scaling of Theorem~\ref{thm:dense-bb}(b) with a constant factor five times larger. Detecting near-duplicates at small angles therefore requires longer sketches just as in the dense case, while demanding almost no additional row sparsity.
\end{remark}

\subsection{Lower bounds}\label{sec:lower}

Whereas our linear Catoni guarantee (Corollary~\ref{cor:gg-classical}(ii)) requires only $r=3\,\mu(m)$ nonzeros per row, both one-bit corollaries require $r$ to scale as $\mu(K)/\eps$. Theorem~\ref{thm:lower} proves that this $1/\eps$ sparsity factor is a genuine property of the estimators rather than an artifact of our analysis whenever each row samples at most half of the coordinates ($2\le r\le n/2$). In addition, part~(d) confirms that neither vector caps nor fourth moments can be guaranteed to remain below $\Omega(\Lambda)$ across a family of $N$ vectors, showing that the logarithmic flatness level $\mu(K)$ of Lemma~\ref{lem:flat} is tight up to constant factors whenever $N/\delta$ lies between two fixed powers of $n$.

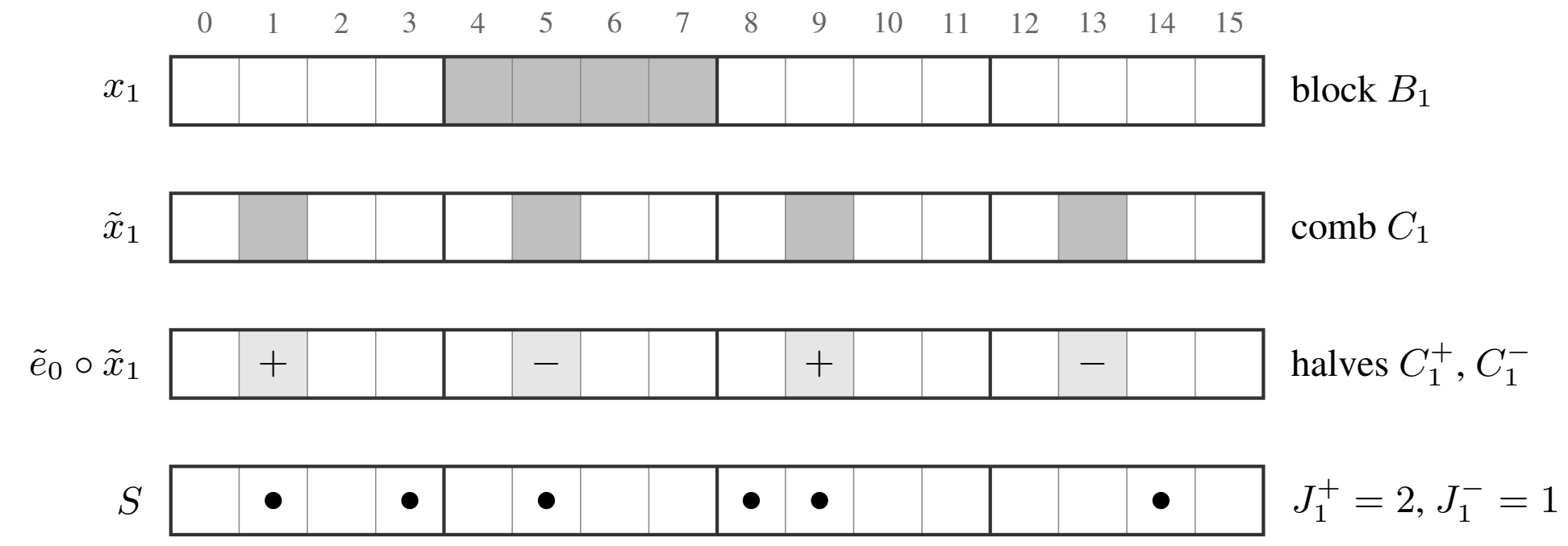
\begin{figure}[t]
\centering
\begin{tikzpicture}[x=0.56cm,y=0.56cm,every node/.style={font=\small,inner sep=1.5pt}]
  \foreach \i in {0,...,15} \node[font=\scriptsize,black!60] at (\i+0.5,1.5) {\i};
  \foreach \i in {4,...,7} \fill[black!25] (\i,0) rectangle ++(1,1);
  \foreach \i in {1,5,9,13} \fill[black!25] (\i,-2) rectangle ++(1,1);
  \foreach \i/\s in {1/+,5/-,9/+,13/-} {
    \fill[black!10] (\i,-4) rectangle ++(1,1);
    \node at (\i+0.5,-3.5) {$\s$};
  }
  \foreach \i in {1,3,5,8,9,14} \fill (\i+0.5,-5.5) circle[radius=2pt];
  \foreach \y in {0,-2,-4,-6} {
    \foreach \i in {1,...,15} \draw[line width=0.3pt,black!45] (\i,\y) -- (\i,\y+1);
    \foreach \i in {4,8,12} \draw[line width=0.8pt,black!80] (\i,\y) -- (\i,\y+1);
    \draw[line width=0.8pt,black!80] (0,\y) rectangle (16,\y+1);
  }
  \node[anchor=east] at (-0.3,0.5) {$x_1$};
  \node[anchor=east] at (-0.3,-1.5) {$\xt_1$};
  \node[anchor=east] at (-0.3,-3.5) {$\tilde e_0\circ\xt_1$};
  \node[anchor=east] at (-0.3,-5.5) {$\capset$};
  \node[anchor=west] at (16.3,0.5) {block $B_1$};
  \node[anchor=west] at (16.3,-1.5) {comb $C_1$};
  \node[anchor=west] at (16.3,-3.5) {halves $C_1^+$, $C_1^-$};
  \node[anchor=west] at (16.3,-5.5) {$J_1^+=2$, $J_1^-=1$};
\end{tikzpicture}
\caption{Geometry of a ``hit'' Walsh block ($n=16$, block size $s=4$; thick vertical lines delineate the $n/s=4$ dyadic blocks, indexed $0,\dots,15$). When the signs of $D$ on block $B_1$ align with a Walsh function, $HD$ maps the contiguous block vector $x_1$ into a sparse comb $C_1$ of $n/s$ coordinates of magnitude $\sqrt{s/n}$, giving $w_{x_1}=k_{x_1}=s$. Pairing $x_1$ with a basis vector $e_0$ outside $B_1$ splits $C_1$ into equal sign halves $C_1^+$ and $C_1^-$ according to $\tilde e_0\circ\xt_1$. A row support $\capset$ of size $r=6$ meeting $C_1^\pm$ in $J_1^+=2$ and $J_1^-=1$ coordinates captures energy $\sigma_{x_1}^2=s(J_1^++J_1^-)/r$, while its cross term $\sigma_{12}\propto J_1^+-J_1^-$ changes sign under swapping $C_1^+$ and $C_1^-$.}
\label{fig:lb-comb}
\end{figure}

Our lower-bound constructions are based on \emph{normalized Walsh blocks} $x_B:=s^{-1/2}\mathbf 1_B$, where the block size $s$ is a power of two and $B\subset\{0,\dots,n-1\}$ is one of the $n/s$ dyadic blocks of $s$ consecutive coordinates (Figure~\ref{fig:lb-comb}). With probability $2s\,2^{-s}$ over the random draw of $D$, the coordinate signs on $B$ align with a Walsh function, causing the Hadamard preconditioner $HD$ to fail to spread $x_B$: instead, $\xt_B=HDx_B$ is supported on a regular ``comb'' $C$ of just $n/s$ coordinates of equal magnitude $\sqrt{s/n}$ and vanishes everywhere else, yielding $w_{x_B}=k_{x_B}=s$. Pairing $x_B$ with a standard basis vector $e_j$ outside $B$---whose preconditioned image $\tilde e_j=HDe_j$ has entries $\pm n^{-1/2}$ everywhere---partitions the comb into equal sign halves $C^+$ and $C^-$ according to the entrywise product $\tilde e_j\circ\xt_B$. On this event, a row support $\capset$ meeting $C^+$ and $C^-$ in $J^+$ and $J^-$ coordinates captures energy $\sigma_{x_B}^2=s(J^++J^-)/r$ of variance $\Theta(s/r)$, while the cross term $\sigma_{12}\propto J^+-J^-$ cancels by symmetry under swapping $C^+$ and $C^-$, isolating a downward support bias of order $s/r$ that we bound from below explicitly.

\begin{theorem}[Lower bounds]\label{thm:lower}
Let $n\ge4$, $2\le r\le n/2$ (so $q\le\tfrac12$), $N\ge1$, $\delta\in(0,\tfrac12]$, and $\Lambda:=\log(\min\{N,n/2\}/\delta)$. The vectors below are built from normalized Walsh blocks and coordinate vectors; all families are fixed independently of $(D,P)$ and have at most $N$ members.
\begin{enumerate}
\item[(a)] \emph{The leading bias terms are attained.} For every power of two $s$ with $2\le s\le r$ there are a unit vector $x$ and an event $\mathcal E$ over $D$ of probability $2s\,2^{-s}$ on which $w_x=k_x=s$ and, for every $\rho\in[-1,1]$ and $v=\rho x+\sqrt{1-\rho^2}\,e_j$ with $j$ outside the block of $x$,
\[
\E[\hat\rho_{\rm BG}(x,v)\mid D]-\rho=-\rho\,\Delta,\qquad\text{where}\quad
\Delta:=1-\E\big[\sigma_x\mid D\big]\ \ge\ \frac{f_q\,(s-\bal)}{16\,r}
\]
under either law, where $\sigma_x^2=\|\xt_\capset\|^2$ is the energy that a row captures from $x$. Under balanced supports and with $\tan\theta=1/(33\sqrt s)$, the pair $u=e_j$, $v=\cos\theta\,e_j+\sin\theta\,x$, again with $j$ outside the block of $x$, satisfies on $\mathcal E$
\[
\E[\hat\theta\mid D]-\theta\le-\theta\,\frac{f_q(s-1)}{17\,r},\quad\text{although}\quad k_u,k_v\le1.01;\quad\text{here } \wU=s+1,\ \kU=s+3 .
\]
\item[(b)] \emph{BG.} Let $\eps\le1/256$. There is a family of at most $N$ pairs with the following property, under either law. Let the estimate be $\hat\rho_{\rm BG}$ with $\ell\ge\pi/(2\eps^2)$, or any decoder $\tilde\rho$ that concentrates around the mean row score, in the sense that $\Pr[|\tilde\rho-\E[Z_1\mid D]|\le\eps\mid D]\ge\frac12$ for every pair of the family and every $D$, where $Z_1$ is the first row score of the pair. If, with probability at least $1-\delta/2$, the estimate has error at most $\eps$ on every pair of the family, then $r>\Lambda/(178\,\eps)$.
\item[(c)] \emph{BB, relative error.} Use balanced supports, let $\eps\le1/272$, and fix $\theta\in(0,\arctan(1/10)]$. There is a family of at most $N$ pairs at angle $\theta$ such that, if $\ell\ge\pi/(\eps^2\theta)$ and with probability at least $1-\delta/2$ every pair has $|\hat\theta-\theta|\le\eps\theta$, then $r>\Lambda/(189\,\eps)$.
\item[(d)] \emph{Flatness and fourth moments are tight.} Some family of at most $N$ unit vectors, fixed independently of $D$, contains with probability larger than $\delta$ a member $x$ with $w_x=k_x>\tfrac12\min\{\Lambda,n/2\}$.
\end{enumerate}
\end{theorem}

Part~(a) demonstrates that our conditional bias upper bounds are tight up to constant factors in their dependence on the data parameters. For BG, the relative downward bias $\Delta$ on the family of pairs $(x,v)$ equals the self-pair deficit of Section~\ref{sec:fjlt}, which is at least $\Var(\sigma_x^2)/16$ (and asymptotically $\Var(\sigma_x^2)/8$). For BB, the first bias term of Theorem~\ref{thm:bb-param}(a) evaluates to $\frac78\sin\theta\,f_q(s+1)/r$ on the constructed pair, confirming that a relative angular bias of order $\kU/r$ is genuinely attained---and proving that the plane parameters $(\kU,\wU)$ cannot be replaced by the individual vector moments $(k_u,k_v)$, which are both at most $1.01$ despite their span containing the spiky Walsh block $x$.

Parts~(b) and~(c) follow from part~(a) by packing $\min\{N,n/s-1\}$ disjoint Walsh blocks of size $s$ into $\{1,\dots,n\}$: whenever $s\lesssim\Lambda$, the random signs $D$ fail to flatten at least one block in the family with probability exceeding $\delta$ (Appendix~\ref{app:lower}). On that event, the hit block suffers a bias of order $\Lambda/r$, which exceeds $2\eps$ (or $2\eps\theta$) unless $r=\Omega(\Lambda/\eps)$; once the sketch length $\ell$ is large enough for the estimate to concentrate within $\eps$ of its conditional mean with probability at least $\frac12$, this bias forces the estimator to fail with overall probability greater than $\delta/2$.

Notice also that robust aggregation cannot rescue BG from this lower bound, because the failure is driven by a shift in the conditional expectation $\E[Z_1\mid D]$ rather than by heavy tails. Accordingly, part~(b) applies to \emph{every} decoder that concentrates around $\E[Z_1\mid D]$, including Catoni's estimator once $\ell=\Omega(\eps^{-2})$. By contrast, the linear estimator GG has zero support bias, so no analogous sparsity lower bound applies to it (though the per-row variance bound of Theorem~\ref{thm:gg-param}(b) is attained up to $3f_q\bal/r$ on a hit Walsh block; see Appendix~\ref{app:gg}).

Finally, because the proof of Theorem~\ref{thm:lower}(b) uses only self-pairs $(x_b,x_b)$---for which $\hat\rho_{\rm BG}(u,u)=\sqrt{\pi/2}\,\|\Phi u\|_1/\ell$ coincides with the $\ell_1$ FJLT of \citet{ailon2009fast} (Section~\ref{sec:why})---it simultaneously establishes a lower bound for $\ell_1$ dimension reduction. While \citet[Lemma~2]{ailon2009fast} chose a row density proportional to $\eps^{-1}$ to ensure $1-\eps\le\E\sigma_u\le1$, part~(b) proves that this $\eps^{-1}$ factor is in fact necessary for the standard $\ell_1$ FJLT estimator whenever $\eps\le1/256$ and $r\le n/2$.

\begin{remark}[Scope and constants of the lower bounds]\label{rem:lb-scope}
We emphasize that Theorem~\ref{thm:lower} bounds the performance of the specific estimators studied here rather than providing an information-theoretic lower bound over arbitrary decoding algorithms. For the standard BG dot-product decoder and any concentration-based decoder (such as Catoni's estimator), part~(b) proves that the worst-case sparsity $r=O(\eps^{-1}\log(nN/\delta))$ of Corollary~\ref{cor:bg-classical} is optimal up to constant factors whenever $n^c\le N/\delta\le(n/\delta)^C$ for fixed constants $0<c\le C$. For SimHash (BB), part~(c) establishes the same optimality for relative error at a small angle $\theta$ under balanced supports, matching Corollary~\ref{cor:bb-classical}(ii) with $\theta_0=\theta$ in the regime where $\log(16\pi/(\eps\theta))=O(1/\eps)$.

At the same time, the numerical constants in our worst-case corollaries and lower bounds are separated by more than three orders of magnitude: in the regime $\eps\le1/256$ (BG) and $\eps\le1/272$ (BB) where Theorem~\ref{thm:lower} applies, the corollaries ask for a sparsity more than $5{,}000$ times the lower bound (whenever $r<n$). Very little of this gap comes from our analysis of the conditional support bias for a fixed $D$: on the hard Walsh-block pairs of part~(a), our conditional upper and lower bounds sandwich the true bias to within a factor of $3$ for BG and $6$ for BB at large $r$ (Section~\ref{sec:experiments}, Figure~\ref{fig:bias-r}). Instead, the gap accumulates when passing to the worst-case corollaries and proving the unconditional lower bounds: the corollaries use simplified closed-form bias bounds, replace fourth moments with worst-case union-bound caps $\mu(K)$, and allocate only a small fraction of the error budget $\eps$ to the bias ($\approx\eps/14$ for BG and $\eps/4$ per bias term for BB), while the lower-bound proofs sacrifice constant factors in the block-hitting and Cantelli arguments.
\end{remark}

\subsection{Certificates and cost}\label{sec:compare}\label{sec:certificates}

Whereas the classical guarantees in Table~\ref{tab:compare} must bound every vector parameter by the worst-case flatness level $\mu(K)$, our conditional theorems for a fixed $D$ adapt directly to the empirical geometry of the dataset.

\paragraph{Data-dependent certificates.}
As introduced in Section~\ref{sec:practice}, a \emph{certificate} is obtained by evaluating one of our fixed-$D$ bounds (Theorems~\ref{thm:gg-param}, \ref{thm:bg-param}, and~\ref{thm:bb-param}, together with Proposition~\ref{prop:bg-support}) at the empirical flatness parameters measured after applying the preconditioner $HD$. Because these theorems hold for every realization of $D$, a practitioner can draw $D$ once, compute the preconditioned vectors $\xt=HDx$ and their summary statistics $(w_x,k_x)$ in $O(n)$ time per vector during encoding (Algorithm~\ref{alg:sketch}), select the required sparsity $r$ and sketch length $\ell$, and only then draw the sparse matrix $P$---without ever needing to reject or redraw $D$. In particular, the caps $w_u$ of the stored database vectors suffice to bound the BG bias uniformly over all queries (Proposition~\ref{prop:bg-support}(i)) as well as the GG row variance for Catoni's decoder (Theorem~\ref{thm:gg-param}(b)), allowing $r$ to be certified offline before any query arrives. At query time, computing $\vt=HDv$ immediately yields the query's parameters $(w_v,k_v)$, which provide a refined per-pair error bar: $k_v$ and the stored $k_u$ enter the tighter fourth-moment bias bound of Proposition~\ref{prop:bg-support}(ii), while $w_v$ enters the dot-product noise term of Theorem~\ref{thm:bg-param}(a). Certifying BB requires the two-dimensional plane parameters $(\wU,\kU)$, which can be computed offline for known pairs or at query time if the preconditioned vectors $\ut$ are retained.

Two practical caveats are worth noting. First, selecting the encoding sparsity $r$ offline based on typical pair parameters---as done in the last two columns of Table~\ref{tab:budgets} for BG and BB---implicitly assumes that incoming queries will exhibit flatness parameters comparable to those of the stored database; while this can be verified for each query upon arrival, only the stored caps $w_u$ provide an unconditional guarantee over arbitrary queries. Second, the certificates reported in Table~\ref{tab:budgets} use the summary parameter forms $(w_x,k_x,\wU,\kU)$; evaluating the full coordinatewise pair certificates derived in Appendices~\ref{app:bg} and~\ref{app:bb}---which retain the exact coordinate variances of $\ut$, $\vt$, and the plane vectors $\cvec_j$---yields even tighter bounds (as shown in Figure~\ref{fig:bias-r}), at the cost of inspecting both $\ut$ and $\vt$.

\paragraph{Comparison with two-stage dimension reduction.}
As noted in Section~\ref{sec:related}, one can bypass the support bias of sparse signed rows by composing two transformations: first reducing the dimension from $n$ to an intermediate dimension $n'$ via a linear FJLT, and then applying a dense signed Gaussian projection from $\R^{n'}$ to $\{ \pm 1\}^\ell$. Proposed as a baseline by \citet[Alg.~3]{yi2015binary} and analyzed by \citet{dirksen2018fast} (with the cross-polytope hash of \citet{kennedy2017fast} serving as a related variant), this two-stage architecture incurs $\Theta(\eps^{-4}\log^2(N/\delta))$ operations per vector beyond the initial Hadamard transform. The reason for the $\eps^{-4}$ scaling is that any distortion introduced by the first-stage projection into $\R^{n'}$ is frozen into the reduced vectors and shared across all $\ell$ rows of the second stage; because the second stage cannot average this distortion away, both stages must independently achieve accuracy of order $\eps/2$. Consequently, the intermediate dimension must scale as $n'=\Theta(\eps^{-2}\log(N/\delta))$, the second stage requires $\ell=\Theta(\eps^{-2}\log(N/\delta))$ rows, and multiplying each reduced vector by the dense $\ell\times n'$ second-stage matrix costs $n'\ell=\Theta(\eps^{-4}\log^2(N/\delta))$ operations.

Signing sparse FJLT rows directly in a single stage saves a full factor of $1/\eps$. Under Corollaries~\ref{cor:bg-classical} and~\ref{cor:bb-classical}, the sparse projection step costs $r\ell=O(\eps^{-3}\mu(K)\log(N/\delta))$ operations per vector, improving upon the two-stage baseline by a factor of $\Theta(\log(N/\delta)/(\eps\,\mu(K)))$---which is $\Theta(1/\eps)$ whenever $n$ and the database size are polynomial in $N/\delta$. In the practical regime of Table~\ref{tab:budgets} ($n=65{,}536$, $N=10^9$, $\eps=0.1$), even an idealized, zero-cost first stage that preserved angles as accurately as a dense Gaussian projection would require an intermediate dimension of $n'\approx17{,}250$ (over a quarter of $n$), while the second stage would require roughly four times as many rows as a single-stage dense sketch at accuracy $\eps$. As a result, the two-stage composition requires slightly \emph{more} multiply--adds per vector than a naive dense projection of the original $n$-dimensional vectors! By contrast, signing sparse FJLT rows directly reduces the projection cost by a factor of $2$ to $5$ in the worst case, and by a factor of $20$ to $70$ when certified with typical pair parameters (see Table~\ref{tab:composite} in Appendix~\ref{app:k4}). While replacing the dense second stage with a structured circulant matrix reduces the cost of the two-stage pipeline \citep{yi2015binary,dirksen2018fast}, doing so reintroduces inter-row dependencies and requires median-of-blocks decoding rather than a simple Hamming popcount or dot product.

\section{Experiments}\label{sec:experiments}

Because the primary contribution of this paper is theoretical, our numerical experiments are designed to illustrate and test the sharpness of our theorems rather than to benchmark end-to-end retrieval systems. After describing our test-pair families and variance-reduced measurement protocol in Section~\ref{sec:exp-pairs}, we investigate three core questions: how sparse the projection rows can be made before estimation accuracy degrades (Section~\ref{sec:exp-sparse}, Figures~\ref{fig:sparsity} and~\ref{fig:rmse-panels}), how closely our upper and lower bounds track the empirical support bias across input geometries (Section~\ref{sec:exp-tight}, Figure~\ref{fig:bias-r}), and how switching from Bernoulli to balanced supports affects estimator bias and error (Section~\ref{sec:exp-balance}, Table~\ref{tab:bias-grid}). Section~\ref{sec:exp-further} concludes with two further experiments on doubly balanced matrices and per-row variance crossovers (Figure~\ref{fig:variance-crossover}), and Appendix~\ref{app:experiments} documents additional simulation details and tables (Tables~\ref{tab:k4-params}--\ref{tab:e2e-rmse}).

\subsection{Test pairs and measurements}\label{sec:exp-pairs}

Every experiment fixes a realization of the sign matrix $D$ and a pair of unit vectors $u=u_1$ and $v=\rho\,u_1+\sqrt{1-\rho^2}\,u_2$, where $u_2$ is a unit vector orthogonal to the stored vector $u_1$. In keeping with our fixed-$D$ theorems, all reported biases and errors are conditioned on $D$. To span the spectrum from benign to worst-case geometry, we evaluate three representative families of input pairs:
\begin{itemize}\itemsep 1pt
\item \emph{Gaussian pairs} represent typical well-spread inputs: $u_1$ and $u_2$ are drawn as random orthonormal vectors alongside a random sign matrix $D$. After preconditioning, the pair is Gaussian-like in the sense of Section~\ref{sec:practice}, with fourth moment $k_u\approx 3$ and cap $w_u\approx 2\log n$ (ranging between $17$ and $20$ at the dimensions $n$ tested here).
\item \emph{One-hot pairs} represent the most favorable geometry: $u_1$ and $u_2$ are distinct standard basis vectors, which the Hadamard transform $HD$ maps to perfectly flat vectors ($\ut_{1,j}^2=\ut_{2,j}^2=1/n$) for every realization of $D$, yielding $k_u=w_u=1$.
\item \emph{Hard pairs} are the worst-case constructions of Theorem~\ref{thm:lower}(a) with block size $s=16$: for BG, $u$ is a normalized Walsh block with $k_u=w_u=16$; for BB, $u$ and $v$ are two nearly flat vectors ($k_u,k_v\le 1.01$) whose span has plane parameters $\kU=19$ and $\wU=17$. We fix $D$ to lie on the block-hitting event $\mathcal E$ of Theorem~\ref{thm:lower}(a) (which occurs under random signs with probability $2s\,2^{-s}\approx 5\cdot10^{-4}$), thereby probing the exact worst-case regime of our conditional bounds.
\end{itemize}
The wide-range bias sweep in Figure~\ref{fig:bias-r} (Section~\ref{sec:exp-tight}) uses ambient dimension $n=2^{14}=16{,}384$ so that the row sparsity $r$ can vary over nearly three orders of magnitude up to $n/2$, while all other experiments use $n=4096$. For Gaussian pairs, a single realization of $D$ is already highly representative because the empirical fourth moments concentrate tightly around their expectations, just as in the $n=65{,}536$ certificate benchmarks of Table~\ref{tab:budgets} (see Table~\ref{tab:k4-params} in Appendix~\ref{app:k4}).

We define the empirical \emph{bias} of an estimator as its conditional expectation given $D$ minus the ground-truth parameter. To measure this bias free of Gaussian Monte Carlo noise, we sample between $2\cdot10^5$ and $4\cdot10^5$ row supports $\capset$ and average the exact conditional score expectations given $\capset$---namely the geometric features $\zeta_\capset=\sigma_{uv}/\sigma_u$ for BG and $\theta_\capset/\pi$ for BB illustrated in Figure~\ref{fig:metric} (Lemma~\ref{lem:random-metric}(iii)). We measure overall estimation \emph{error} by the root mean squared error (RMSE) of the cosine estimate: Figure~\ref{fig:sparsity} evaluates this RMSE directly from exact conditional single-row moments, whereas Figure~\ref{fig:rmse-panels} and Table~\ref{tab:e2e-rmse} run end-to-end sketch simulations following Algorithm~\ref{alg:sketch} (Appendix~\ref{app:experiments}). For SimHash (BB), we report the relative angular bias $|\E[\hat\theta\mid D]-\theta|/\theta$ (matching the normalization of Theorem~\ref{thm:lower}(a)) when evaluating bias in Figure~\ref{fig:bias-r}, and the RMSE of $\hat\rho_{\rm BB}=\cos\hat\theta$ when evaluating total error.

\subsection{How sparse can the rows be?}\label{sec:exp-sparse}

Our conditional theorems bound the error of a one-bit sketch by the sampling noise of a dense sketch (scaling as $1/\sqrt\ell$) plus a support bias (scaling as $1/r$ independently of $\ell$). Accordingly, the empirical RMSE should closely follow the two-term decomposition $\sqrt{\sigma^2/\ell+b^2}$, where $\sigma^2$ is the per-row score variance and $b$ is the support bias. For both BG and BB, $\sigma^2$ is virtually indistinguishable from its dense value: $\pi/2-\rho^2$ for BG (Section~\ref{sec:dense}) and, by the delta method, $\theta(\pi-\theta)\sin^2\theta$ for $\hat\rho_{\rm BB}$. For the linear estimator GG, the bias $b$ is identically zero, while by \eqref{eq:gg-moments} in Section~\ref{sec:gg} the dense row variance $1+\rho^2$ increases by the $O(1/r)$ fourth-moment term $\frac{f_q}{r}(3k_{uv}-\bal(1+2\rho^2))$. This decomposition predicts two distinct behaviors: a sparse one-bit sketch should match the accuracy of a dense sketch so long as its support bias $b$ remains small relative to the dense noise floor $\sigma/\sqrt\ell$, whereas a sparse GG sketch should incur a fixed multiplicative variance overhead across all sketch lengths $\ell$. We test both predictions on the Gaussian pair at $\rho=0.9$.

\paragraph{Varying row sparsity at fixed sketch length.}
Returning to Figure~\ref{fig:sparsity} in Section~\ref{sec:practice}, we sweep the row sparsity from $r=n$ down to $r=8$ at sketch lengths $\ell=2^9$ and $\ell=2^{15}$. In Figure~\ref{fig:sparsity}(a), the linear estimator GG incurs the same relative RMSE inflation at both sketch lengths---roughly $25\%$ under Bernoulli supports and $17\%$ under balanced supports at $r=8$, dropping below $7.2\%$ once $r\ge 32$. In Figure~\ref{fig:sparsity}(b), the BG RMSE at $\ell=2^9$ remains within $10\%$ of the dense baseline down to $r=32$ (Bernoulli) and $r=16$ (balanced); when the sketch length increases to $\ell=2^{15}$, the dense noise floor $\sigma/\sqrt\ell$ drops by a factor of eight, requiring an eightfold increase in row sparsity ($r=256$ for Bernoulli and $r=128$ for balanced supports) to keep the $O(1/r)$ support bias negligible. Meanwhile, in Figure~\ref{fig:sparsity}(c), BB stays within $0.5\%$ of the dense RMSE all the way down to $r=8$ at both sketch lengths because coordinate cancellations make its support bias exceptionally small on generic Gaussian planes (Section~\ref{sec:exp-tight}).

\begin{figure}[t]
\centering
\includegraphics[width=\textwidth]{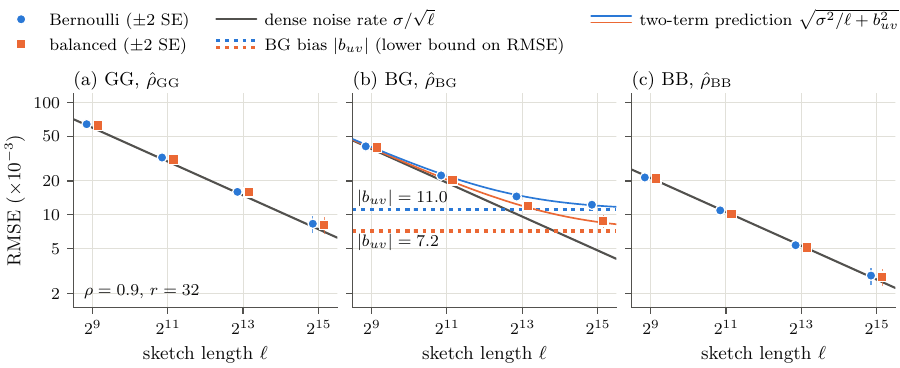}
\caption{Total sketch error decomposes into dense sampling noise plus support bias: as sketch length $\ell$ grows at fixed sparsity $r=32$, GG (panel~(a)) and BB (panel~(c)) continue to decay as $\ell^{-1/2}$, whereas BG (panel~(b)) saturates at its bias floor. Points show end-to-end RMSE on the Gaussian pair ($n=4096$, $\rho=0.9$, $r=32$) with $\pm2$ standard errors. Solid gray lines show the dense rate $\sigma/\sqrt\ell$; in panel~(b), dotted lines mark the independently measured BG bias $|b_{uv}|$ from Table~\ref{tab:bias-grid}, and thin solid curves show $\sqrt{\sigma^2/\ell+b_{uv}^2}$.}
\label{fig:rmse-panels}
\end{figure}

\paragraph{Varying sketch length at fixed row sparsity.}
Figure~\ref{fig:rmse-panels} fixes the row sparsity at $r=32$ and sweeps the sketch length from $\ell=2^9$ to $2^{15}$. Both GG in Figure~\ref{fig:rmse-panels}(a) and BB in Figure~\ref{fig:rmse-panels}(c) track the $\ell^{-1/2}$ dense convergence rate across the entire range. By contrast, BG in Figure~\ref{fig:rmse-panels}(b) follows the $\ell^{-1/2}$ curve initially before leveling off at an error floor equal to its support bias $|b_{uv}|$, which we measure independently in Table~\ref{tab:bias-grid} (Section~\ref{sec:exp-balance}) as $11.0\cdot10^{-3}$ under Bernoulli supports and $7.2\cdot10^{-3}$ under balanced supports. Across all sketch lengths, the empirical BG error lies squarely on the theoretical curve $\sqrt{\sigma^2/\ell+b_{uv}^2}$.

\paragraph{SimHash also hits a bias floor when coordinates do not cancel.}
Although BB exhibits negligible bias on the Gaussian pair of Figure~\ref{fig:rmse-panels}(c), it is not immune to support bias. Repeating the same sketch-length sweep on the one-hot pair ($r=32$, $\rho=0.9$) in Table~\ref{tab:e2e-rmse} (Appendix~\ref{app:e2e}), the bias of $\hat\rho_{\rm BB}$ is $+1.6\cdot10^{-3}$ under Bernoulli supports and $+1.8\cdot10^{-3}$ under balanced supports---five to six standard errors away from zero and comparable to the dense noise floor of $2.7\cdot10^{-3}$ at $\ell=2^{15}$. Thus SimHash, too, eventually reaches a sketch length beyond which adding more rows without increasing $r$ yields diminishing returns.

In summary, the target sampling noise dictates the sketch length $\ell$, and for one-bit estimators the support bias then dictates the required row sparsity $r$. A few dozen nonzeros per row suffice to keep GG within $7\%$ of a dense projection (and suffice for BB on typical Gaussian inputs), whereas BG requires $r$ to scale inversely with the target error $\eps$ on benign Gaussian pairs just as it does in the worst case of Theorem~\ref{thm:lower}(b).

\subsection{How tight are the bias bounds?}\label{sec:exp-tight}

\begin{figure}[t]
\centering
\includegraphics[width=\textwidth]{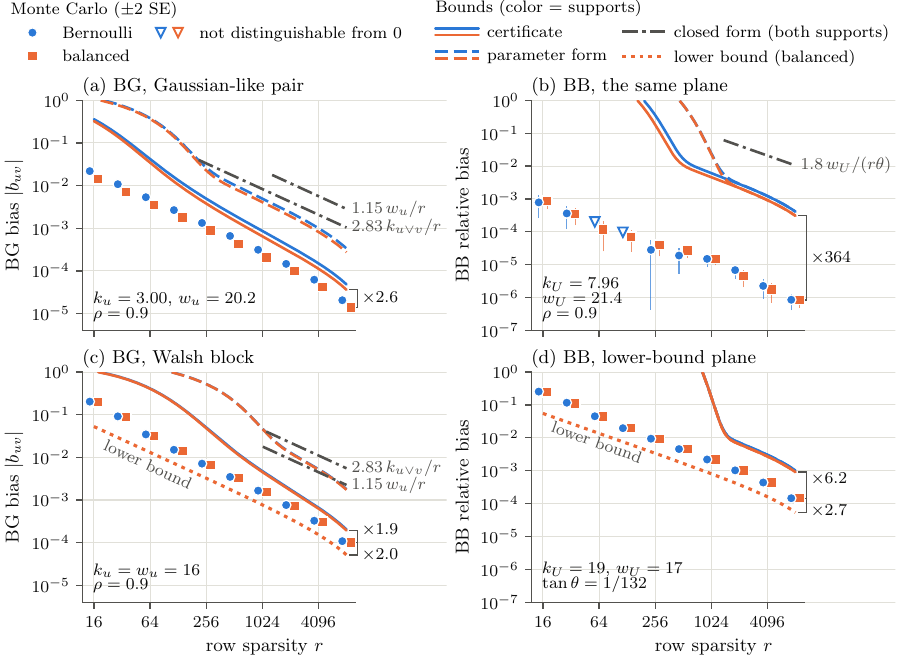}
\caption{On the worst-case pairs of Theorem~\ref{thm:lower}(a) (panels~(c) and~(d)), our upper and lower bounds sandwich the true support bias to within a factor of roughly $3$ for BG and $6$ for BB once $r$ is sufficiently large; on a typical Gaussian pair (panels~(a) and~(b)), only the coordinatewise BG pair certificate achieves comparable tightness. Markers show empirical bias with $\pm2$ standard errors ($n=2^{14}$), normalized by $\theta$ for BB. Curves plot the theoretical bounds across the regime where their hypotheses hold and their value is below $1$; brackets annotate the ratios at $r=8192$ under balanced supports.}
\label{fig:bias-r}
\end{figure}

Figure~\ref{fig:bias-r} compares the empirical support bias against three hierarchy levels of our theoretical upper bounds as $r$ varies from $16$ to $n/2$:
\begin{enumerate}\itemsep 1pt
\item the coordinatewise \emph{pair certificate} computed directly from $\ut$ and $\vt$, which retains the exact coordinate variances $V_u,V_{uv}$ for BG and uses a matrix Bernstein metric-tail bound over the row vectors $\cvec_j$ for BB (Appendices~\ref{app:bg} and~\ref{app:bb});
\item the \emph{parameter form} depending only on summary caps and fourth moments (Proposition~\ref{prop:bg-support}(ii) for BG and the two bias terms of Theorem~\ref{thm:bb-param}(a) for BB---the same bounds evaluated at $n=65{,}536$ to produce the certified budgets in Table~\ref{tab:budgets}, Table~\ref{tab:budget-split}, and Table~\ref{tab:composite}); and
\item the single-term \emph{closed forms} ($1.15\,w_u/r$ and $2.83\,\kmax/r$ for BG; $1.8\,\wU/r$ for BB).
\end{enumerate}
On the hard pairs in Figures~\ref{fig:bias-r}(c) and~\ref{fig:bias-r}(d), we also plot the theoretical lower bounds of Theorem~\ref{thm:lower}(a). Because all empirical biases and bounds decay at the same $1/r$ rate once $r$ is past the tail threshold, comparing them reduces to comparing their prefactors.

\paragraph{Typical Gaussian pair.}
For BG on the Gaussian pair at $\rho=0.9$ (Figure~\ref{fig:bias-r}(a)), the coordinatewise pair certificate tracks the empirical bias closely, exceeding it by only a factor of $2.4$ to $3.5$ for $r\ge 512$ (and a factor of $5$ to $6$ at $r=128$). By contrast, the summary parameter form and closed-form bounds are $16$ to $205$ times larger than the empirical bias; the cap bound $1.15\,w_u/r$ is the most conservative because it substitutes a cap of $w_u\approx 20$ in place of a fourth moment of $k_u\approx 3$, though it has the virtue of depending solely on the stored vector $u$.

For SimHash (BB) on the same Gaussian plane (Figure~\ref{fig:bias-r}(b)), every theoretical upper bound exceeds the empirical bias by more than two orders of magnitude. As noted in Section~\ref{sec:bb} (and derived explicitly in Appendix~\ref{app:bb}), the leading second-order term of the BB bias is a signed sum over the $n$ coordinates weighted by $\|\cvec_j\|^4$ whose terms oscillate with the orientation of the plane's row vectors $\cvec_j\in\R^2$. On an isotropic Gaussian plane, these orientations are distributed uniformly around the circle, causing the summands to cancel almost completely, whereas our Frobenius-norm bounds sum the magnitude of each coordinate's contribution.

\paragraph{Worst-case Walsh-block pairs.}
On the hard pairs of Theorem~\ref{thm:lower}(a) (Figures~\ref{fig:bias-r}(c) and~\ref{fig:bias-r}(d)), the coordinate signs align instead of canceling, and the empirical bias is tightly bracketed between our upper and lower bounds. For BG in Figure~\ref{fig:bias-r}(c), Theorem~\ref{thm:lower}(a) expresses the exact bias as $-\rho\Delta$ and proves the lower bound $\Delta\ge f_q(s-\bal)/(16r)=\Var(\sigma_u^2)/16$. At large $r$, the measured bias is almost exactly twice this lower bound, matching the asymptotic expansion $\Delta=(1+o(1))\Var(\sigma_u^2)/8$ as $s/r\to0$ (consistent with \eqref{eq:bg-second-order}; see Appendix~\ref{app:lower}), while the upper pair certificate sits within a factor of $1.9$ to $2.9$ of the true bias for all $r\ge1024$.
Similarly, for BB under balanced supports in Figure~\ref{fig:bias-r}(d), once $r\ge2048$ (so that the exponential cap tail has decayed), the empirical bias sits $2.7$ times above the theoretical lower bound while the upper bound of Theorem~\ref{thm:bb-param}(a) sits roughly $6$ times above the empirical bias. Recall that both vectors of this hard BB pair are individually nearly flat ($k_u,k_v\le1.01$), confirming empirically that the two-dimensional plane parameters $(\kU,\wU)$---not the single-vector moments---govern the support bias of SimHash.

In short, our bias bounds are tight to within small constant factors ($2\times$ to $6\times$) on the worst-case inputs they must cover, while on benign Gaussian inputs only the coordinatewise BG pair certificate captures the cancellation between coordinates.

\subsection{What does balancing change?}\label{sec:exp-balance}

\begin{table}[t]\centering\small
\setlength{\tabcolsep}{4.2pt}
\begin{tabular}{@{}lr@{\hspace{10pt}}rrrrrr@{\hspace{10pt}}r@{\,/\,}l@{}}
\toprule
 & & \multicolumn{8}{c}{all entries $\times10^{-3}$} \\
\cmidrule(l){3-10}
 & & \multicolumn{6}{c}{measured bias $b_{uv}$} & \multicolumn{2}{c}{bound on $|b_{uv}|$} \\
\cmidrule(lr){3-8}\cmidrule(l){9-10}
 & & \multicolumn{2}{c}{$\rho=0.5$} & \multicolumn{2}{c}{$\rho=0.9$} & \multicolumn{2}{c}{$\rho=0.99$} & \multicolumn{2}{c}{$\rho=0.99$} \\
\cmidrule(lr){3-4}\cmidrule(lr){5-6}\cmidrule(lr){7-8}\cmidrule(l){9-10}
input ($k_u$) & $r$ & Bern. & bal. & Bern. & bal. & Bern. & bal. & \multicolumn{1}{r@{\,/\,}}{Bern.} & bal. \\
\midrule
Gaussian ($3.03$) & $32$ & $-6.8$ & $-4.2$ & $-11.0$ & $-7.2$ & $-11.8$ & $-7.8$ & $68.6$ & $56.5$ \\
 & $128$ & $-1.6$ & $-1.1$ & $-2.6$ & $-1.8$ & $-2.8$ & $-1.9$ & $7.8$ & $5.7$ \\
\addlinespace[3pt]
one-hot ($1.00$) & $32$ & $-1.8$ & $-0.1$ & $-3.4$ & $-0.1$ & $-3.8$ & $0.0$ & $14.3$ & $<0.1$ \\
 & $128$ & $-0.4$ & $0.0$ & $-0.8$ & $0.0$ & $-0.9$ & $0.0$ & $2.2$ & $<0.1$ \\
\bottomrule
\end{tabular}
\caption{Switching from Bernoulli to balanced supports scales the BG support bias by approximately $(k_u-1)/k_u$, eliminating roughly one third of the bias on Gaussian inputs ($k_u\approx 3$) and all of the bias on flat one-hot inputs ($k_u=1$). Entries report the empirical BG bias $b_{uv}$ ($n=4096$, standard errors $\le 0.3\cdot10^{-3}$) across $\rho\in\{0.5,0.9,0.99\}$ and $r\in\{32,128\}$, alongside the coordinatewise BG pair certificate (Appendix~\ref{app:bg}) at $\rho=0.99$.}
\label{tab:bias-grid}
\end{table}

As shown in Sections~\ref{sec:balanced} and~\ref{sec:setting}, fixing the row support size to $r$ reduces the variance of the captured energy $\sigma_u^2$ from $(1-q)k_u/r$ under Bernoulli supports to $f_q(k_u-1)/r$ under balanced supports. Because the second-order expansion \eqref{eq:bg-second-order} in Section~\ref{sec:bg} approximates the BG support bias as $-\frac\rho8\Var(\sigma_u^2)$ plus a smaller cross-covariance term, balancing the rows when $r\ll n$ should scale the BG bias by $(k_u-1)/k_u$: cutting it by one third on Gaussian-like inputs where $k_u\approx3$ (from $-\frac38\rho/r$ to $-\frac14\rho/r$), eliminating it completely on flat inputs where $k_u=1$ (Proposition~\ref{prop:bg-support}(iii)), and making only a minor difference on spiky inputs with large $k_u$.

Table~\ref{tab:bias-grid} verifies each of these predictions across Gaussian and one-hot pairs at $r\in\{32,128\}$ and $\rho\in\{0.5,0.9,0.99\}$. At $\rho=0.99$, where the cross-covariance term of \eqref{eq:bg-second-order} is negligible, every measured bias agrees with the leading term $-\frac\rho8\Var(\sigma_u^2)$ to within one standard error. At $\rho=0.5$, the cross-covariance term in \eqref{eq:bg-second-order} contributes an additional $6\%$ (Bernoulli) to $9\%$ (balanced) on the Gaussian pair and vanishes identically on the flat one-hot pair; including it brings \eqref{eq:bg-second-order} within $2.2$ standard errors of every cell in Table~\ref{tab:bias-grid}. Across all correlations on the Gaussian input, balanced supports reduce the magnitude of the BG bias by roughly one third. On the one-hot input, balanced supports eliminate the bias completely (matching zero within Monte Carlo precision), whereas Bernoulli supports still suffer a nonzero bias driven entirely by random fluctuations in row length. Conversely, on the spiky Walsh block of Figure~\ref{fig:bias-r}(c) ($k_u=16$), the two support laws produce nearly identical biases ($(k_u-1)/k_u=15/16$). Finally, the rightmost two columns of Table~\ref{tab:bias-grid} confirm that at $r=128$, the coordinatewise BG pair certificate sits within a factor of $3$ of the measured bias on the Gaussian input and vanishes as expected on the balanced one-hot input.

For SimHash (BB), by contrast, the empirical bias (Figures~\ref{fig:bias-r}(b) and~\ref{fig:bias-r}(d)) and overall RMSE (Figure~\ref{fig:rmse-panels}(c)) are virtually indistinguishable between Bernoulli and balanced supports. As noted in Section~\ref{sec:setting}, conditioned on its nonzero count $|\capset|$, a Bernoulli row is simply a rescaled balanced row, and multiplying a row projection by a positive scalar leaves its sign---and thus the disagreement angle---unchanged. Moreover, flat individual vectors do not eliminate the BB bias: on the one-hot pair where the balanced BG bias vanishes, Table~\ref{tab:e2e-rmse} (Appendix~\ref{app:e2e}) shows that the BB bias remains nonzero under both support laws because the off-diagonal metric entry $\sigma_{12}$ still fluctuates from row to row.

In summary, balanced supports remove a $1/k_u$ fraction of the BG support bias while leaving the BB bias essentially unchanged.

\subsection{Two further observations}\label{sec:exp-further}

\paragraph{Doubly balanced supports (balancing columns as well as rows).}
Whereas our balanced FJLT fixes the number of nonzeros per row, the sparse JL transforms of \citet{kane2014sparser} and \citet{nelson2013osnap} fix the number of nonzeros per column. Enforcing both constraints simultaneously yields \emph{doubly balanced} supports, in which every row has $r$ nonzeros and every column has $\ell r/n$ nonzeros. Because column regularity couples the rows, our independent-row concentration bounds do not directly apply, and establishing high-probability tail bounds for doubly balanced one-bit sketches remains an open question. Empirically, on the Gaussian pair ($\rho=0.9$, $r\in\{8,32\}$), doubly balanced supports leave the BG and BB biases unchanged within standard error---as expected, since linearity of expectation implies that the bias depends only on the marginal distribution of a single row. However, column regularity measurably reduces the variance of the linear estimator GG: because the energies captured across all $\ell$ rows are forced to sum deterministically to $\ell$, one of the three $V_u=\Var(\sigma_u^2)$ terms in the per-row squared-norm variance $\Var(X_i^2\mid D)=2+3V_u$ from \eqref{eq:gg-moments} cancels out across rows. This predicts a norm-RMSE reduction factor of $\sqrt{(2+2V_u)/(2+3V_u)}$ ($1.4\%$ to $4.7\%$ in our setup), matching the observed $3\%$ to $5\%$ reduction within roughly one standard error.

\begin{figure}[t]
\centering
\includegraphics[width=\textwidth]{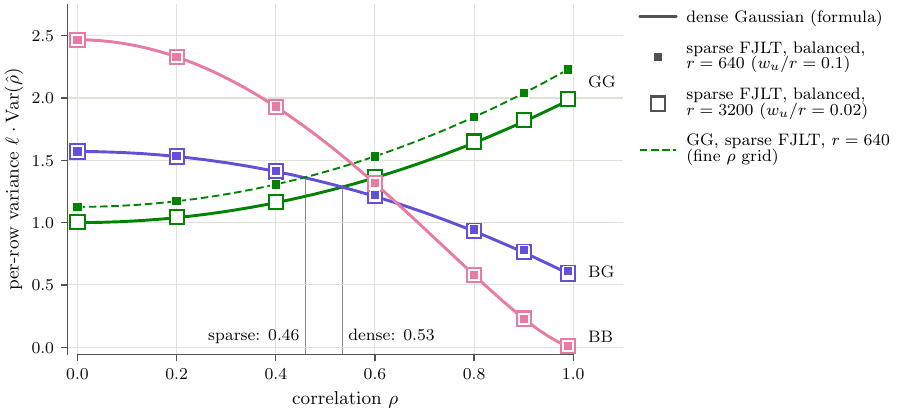}
\caption{Row sparsity inflates the per-row variance of the linear estimator GG while leaving the one-bit estimators BG and BB virtually unchanged, shifting the variance crossover to even lower correlations. Solid curves plot the dense Gaussian per-row variances across $\rho\in[0,1)$; markers show balanced sparse FJLT variances ($10^6$ supports per point) on a spiky stored vector ($k_u=w_u=64$, $n=4096$) at $r=640$ ($w_u/r=0.1$, filled) and $r=3200$ ($w_u/r=0.02$, open). The dashed curve tracks the sparse GG variance at $r=640$, moving the GG--BG crossover from $\rho=0.53$ (dense) down to $\rho=0.46$ (sparse).}
\label{fig:variance-crossover}
\end{figure}

\paragraph{Per-row variance crossover between linear and one-bit estimators.}
Recall from Section~\ref{sec:dense} that for dense Gaussian projections, a BG row score has smaller variance ($\pi/2-\rho^2$) than a linear GG row score ($1+\rho^2$) whenever $|\rho|\gtrsim 0.53$. Moving to sparse rows shifts this crossover even further in favor of the one-bit estimators: whereas the unconditional second moment of a BG row score remains locked at $\E[Z_i^2\mid D]=\pi/2$ (Section~\ref{sec:bg}) and a BB bit disagreement remains bounded in $[0,1]$, the variance of a GG row product grows by the $O(k_{uv}/r)$ fourth-moment term of \eqref{eq:gg-moments}. Figure~\ref{fig:variance-crossover} illustrates this effect over $10^6$ balanced supports for a spiky stored vector ($k_u=w_u=64$, $n=4096$). Even at moderate sparsity $r=640$ ($w_u/r=0.1$), the BG and BB per-row variances lie right on top of their dense curves (deviating by at most $0.022$ and $0.0023$), whereas the dashed sparse GG curve shifts upward by $0.13$ to $0.25$ across $\rho$ (returning toward the dense curve at $r=3200$). As marked by the vertical drop lines in Figure~\ref{fig:variance-crossover}, the correlation threshold above which one-bit scores have lower per-row variance than GG drops from $\rho=0.53$ to $0.46$ for BG, and similarly from $\rho=0.59$ to $0.55$ for BB.

\section{Discussion}\label{sec:discussion}

Since \citet{ailon2009fast} introduced the fast Johnson--Lindenstrauss transform nearly two decades ago, fast preconditioned sparse projections and one-bit quantization have largely evolved along separate tracks. Classical FJLT theory established sharp finite-sample guarantees for linear embeddings, while practical one-bit schemes---from SimHash \citep{charikar2002similarity} to modern asymmetric quantizers \citep{zandieh2024qjl,gao2024rabitq,zandieh2025turboquant}---either relied on dense $\Theta(n\ell)$ Gaussian projections, turned to structured circulant or Toeplitz matrices whose coupled rows complicate concentration, or applied sparse projections heuristically without non-asymptotic guarantees. This paper closes that gap by providing a unified finite-sample analysis across all three precision regimes: full precision on both sides (GG), one bit against full precision (BG), and one bit on both sides (BB).

Viewing each sparse row through the random metric $M_\capset$ it induces on the two-dimensional span of a vector pair clarifies why sign quantization fundamentally alters the behavior of sparse sketches. Linear inner-product estimates read a linear functional of $M_\capset$ and remain unbiased at every sparsity, whereas signed estimators read nonlinear functionals---a normalized projection for BG and a distorted angle for BB---whose curvature converts support-to-support energy fluctuations into a systematic $O(1/r)$ support bias that does not decay as the sketch length $\ell$ grows. Our Walsh-block lower bounds (Theorem~\ref{thm:lower}) prove that the resulting $\Theta(\mu/\eps)$ worst-case sparsity scaling is an intrinsic property of one-bit sparse sketching rather than an artifact of our proof, and simultaneously settle the necessity of the $\eps^{-1}$ density factor in the classical $\ell_1$ FJLT of \citet{ailon2009fast}. At the same time, conditioning on the realized preconditioner $D$ yields data-dependent certificates expressed in terms of empirical fourth moments rather than worst-case caps, reducing the certified sparsity on typical well-spread inputs from $15$--$17\%$ of the ambient dimension to roughly $1\%$ (Table~\ref{tab:budgets}).

Two algorithmic design choices play a central role throughout our results. First, replacing independent Bernoulli supports with \emph{balanced supports} of fixed cardinality $r$ aligns the projection with contiguous, branch-free gather kernels on modern SIMD and GPU hardware while strictly improving statistical concentration. Removing row-length fluctuations cuts the BG support bias by roughly one third on Gaussian-like inputs and eliminates it completely on flat vectors (Proposition~\ref{prop:bg-support}(iii), Table~\ref{tab:bias-grid}). Within the same unified analysis, our full-precision bounds extend the optimal linear FJLT sparsity of \citet{fandina2023fast} to balanced supports and arbitrary failure probability $\delta$, while supplying fully explicit constants that correct inaccuracies in the original constants of \citet{fandina2023fast}.

Second, replacing the sample average with \emph{Catoni's robust mean estimator} substantially strengthens both linear and asymmetric guarantees. For linear inner products, Catoni's decoder removes the $\eps$-dependence of the row sparsity altogether ($r=3\,\mu(m)$ for all $\eps\in(0,1]$). For asymmetric one-bit estimation (BG), it removes all flatness conditions on the query vector, accommodates queries chosen adaptively after observing $D$, and halves the certified sparsity on typical inputs (Table~\ref{tab:budgets}). Crucially, in its non-iterative one-step formulation (Remark~\ref{rem:catoni-fast}), the Catoni BG decoder reduces to softly clipping the uncompressed query projections once upon arrival and computing a standard sign dot product. Because this requires only a single streaming pass with negligible overhead beyond a standard inner product, it is directly suited to latency-critical deep learning and retrieval pipelines.

\subsection{Future work and open problems}\label{sec:limitations}

\paragraph{Applications to deep learning and vector databases.}
While our experiments here focus on controlled tests of the theoretical bounds, applying balanced signed FJLT sketches and single-pass Catoni decoding to large-scale machine learning and database systems is an immediate direction for future work. Natural targets include long-context key-value cache compression for transformer attention---where QJL \citep{zandieh2024qjl} and TurboQuant \citep{zandieh2025turboquant} currently pair asymmetric one-bit codes with dense projections or unanalyzed structured transforms, particularly when attention keys are grouped across heads to reach larger effective dimensions $n$---as well as billion-scale approximate nearest-neighbor indexes built on asymmetric sign quantization \citep{gao2024rabitq}. Co-designing fused fast Walsh--Hadamard, balanced sparse gather, and one-step Catoni kernels on GPUs and evaluating end-to-end recall and perplexity on production workloads remain important practical steps.

\paragraph{High-probability tail bounds for fourth moments.}
On the theoretical side, when passing from fixed-$D$ bounds to worst-case corollaries, we bound the fourth moments $k_x$ and $\kU$ by the corresponding caps $w_x$ and $2\wU$, which grow as $O(\log(nN/\delta))$ even though $\E_Dk_x\le3$ and $\E_D\kU\le8$ for any pre-specified input. Because $k_x$ and $\kU$ are degree-four polynomials in the Rademacher signs of $D$, standard hypercontractive inequalities yield dimension-free bounds of order $O(\log^2(N/\delta))$, removing the dependence on $n$ at the cost of squaring the logarithm, whereas Theorem~\ref{thm:lower}(d) shows only that $k_x$ (and hence $\kU\ge k_u$) can reach $\Omega(\min\{\Lambda,n\})$. Determining whether $k_x$ and $\kU$ concentrate at the linear logarithmic scale $O(\log(N/\delta))$ without paying a $\log n$ factor would directly sharpen the leading terms of our worst-case one-bit guarantees.

\paragraph{Additive-error lower bounds, bias calibration, and ranking.}
Although Theorem~\ref{thm:lower}(c) proves our $\Omega(\eps^{-1}\Lambda)$ sparsity lower bound for SimHash under relative angular error with balanced supports (and conditioning on support size suggests the same order under Bernoulli supports), two fundamental questions remain open: whether SimHash requires $r=\Omega(\eps^{-1}\Lambda)$ under \emph{additive} error, and whether alternative decoders can bypass the $\eps^{-1}$ sparsity barrier for BG. In particular, on the hard Walsh-block pairs of Theorem~\ref{thm:lower}(a), the conditional BG bias is purely multiplicative ($\E[\hat\rho_{\rm BG}\mid D]=\rho\,(1-\Delta)$ with $\Delta=1-\E[\sigma_u\mid D]$), so storing the single scalar factor $(1-\Delta)^{-1}$ alongside each database vector---analogous to the per-vector scaling in RaBitQ \citep{gao2024rabitq}---cancels the leading $-\frac\rho8\Var(\sigma_u^2)$ term of \eqref{eq:bg-second-order} for every stored vector. Whether such scalar calibration (or maximum-likelihood one-bit decoders; \citealt{li2019signfull}) can provably beat $\Theta(\mu(K)/\eps)$ sparsity in general, where the cross-covariance term of \eqref{eq:bg-second-order} also contributes, and what sparsity suffices for pure top-$k$ nearest-neighbor ranking rather than value estimation, are natural questions for future work.

\bibliographystyle{plainnat}
\bibliography{main}
\makeatletter\makeatother

\clearpage
\appendix
\section{Proofs}\label{app:proofs}

All of our proofs follow a common three-step blueprint. First, we condition on a fixed sign matrix $D$ and examine a single row of the sparse matrix $P$: conditioned on its support $\capset$, the row acts on any pair of vectors $(u,v)$ exactly as a dense Gaussian row acts on the subsampled pair $(\ut_\capset,\vt_\capset)$, inducing the random metric $M_\capset$ on their span (Lemma~\ref{lem:random-metric}), whose entries are rescaled coordinate sums over $\capset$ (Lemma~\ref{lem:sampling}). Because each estimator averages a specific functional of $M_\capset$ across the $\ell$ rows, we first bound the single-row support bias---the expected deviation of that functional over the random draw of $\capset$---and then bound the row-to-row sampling fluctuations across the $\ell$ conditionally independent rows. Finally, we take expectations over the random draw of $D$: invoking Lemma~\ref{lem:flat} bounds the empirical flatness parameters of all pre-specified inputs by the worst-case flatness level $\mu(K)$ of \eqref{eq:mu-def} with high probability, converting each fixed-$D$ theorem into its corresponding classical corollary.

The appendix is organized into six parts. Appendix~\ref{app:tools} collects concentration inequalities and robust estimation tools, while Appendix~\ref{app:devices} establishes Lemmas~\ref{lem:random-metric} and~\ref{lem:sampling} along with the algebraic relations among our flatness parameters. Appendices~\ref{app:gg}, \ref{app:bg}, and~\ref{app:bb} prove the guarantees for GG, BG, and BB, respectively, and Appendix~\ref{app:lower} establishes the lower bounds of Theorem~\ref{thm:lower}. Three additional results appear exclusively in this appendix: a scale-free Catoni estimator for pairwise squared distances (Theorem~\ref{thm:gg-dist}) and the coordinatewise pair certificates for the support bias of BG and BB (Propositions~\ref{prop:bg-bias} and~\ref{prop:bb-bias}). Table~\ref{tab:notation} summarizes our notation, and every numerical constant and elementary inequality in Appendices~\ref{app:tools}--\ref{app:lower} has been verified by an accompanying validation script (Appendix~\ref{app:experiments}).

\begin{table}[tp]\centering\footnotesize
\setlength{\tabcolsep}{3pt}
\renewcommand{\arraystretch}{0.93}
\newcommand{\nhead}[1]{\multicolumn{4}{@{}l}{\textit{#1}}\\}
\begin{tabular}{@{}l L{5.25cm} @{\hspace{10pt}} l L{5.52cm}@{}}
\toprule
\nhead{Conventions}
$\|x\|$, $\|x\|_p$ & Euclidean norm, $p$-norm & $x\circ y$ & entrywise product, $(x\circ y)_j=x_jy_j$\\
$[n]$ & the index set $\{1,\dots,n\}$ & $\|A\|_F$, $\|A\|_{\rm op}$ & Frobenius norm, operator norm\\
$I_2$ & the $2\times2$ identity matrix & $A\succeq B$ & $A-B$ is positive semidefinite\\
$e_1$, $e_j$ & \multicolumn{3}{l}{$e_1=(1,0)^\top\in\R^2$ holds the local coordinates of $u$ in its plane; $e_j$ is the $j$th basis vector of $\R^n$}\\
$\sign$, $\log$ & \multicolumn{3}{l}{$\signzero$, so that a sign is one bit; $\log$ is the natural logarithm}\\
\midrule
\nhead{Dimensions, counts and confidence}
$n$ & dimension, a power of $2$ & $\ell$ & sketch length (number of rows)\\
$r$ & row sparsity (its mean if Bernoulli) & $q$ & density $r/n$\\
$m$ & number of stored vectors (or points) & $N$ & number of pairs\\
$\Ufam$ & finite family of unit vectors & $\eps,\delta$ & accuracy, failure probability\\
$L$ & $\log(2N/\delta)$ & $\Lambda$ & $\log(\min\{N,n/2\}/\delta)$ (Theorem~\ref{thm:lower})\\
$\mu(K)$ & \multicolumn{3}{L{11.4cm}@{}}{flatness level $2\log(4nK/\delta)$ for an effective count $K$ of pre-specified vectors \eqref{eq:mu-def}, bounding their caps $w_x$ with high probability over $D$ (Lemma~\ref{lem:flat})}\\
\midrule
\nhead{The transform}
$\Phi=PHD$ & the FJLT, an $\ell\times n$ matrix & $D$ & diagonal of Rademacher signs\\
$H$ & normalized Walsh--Hadamard matrix & $P$ & sparse Gaussian, $P_{ij}=\mathbf 1\{j\in\capset_i\}\,g_{ij}$\\
$\capset$, $\capset_i$ & support of a row, of row $i$ & $g_{ij}$ & i.i.d.\ $\mathcal N(0,1/q)$ entries of $P$\\
$f_q$ & $\frac{n-r}{n-1}$ (balanced), $1-q$ (Bernoulli) & $\bal$ & $1$ (balanced), $0$ (Bernoulli)\\
$G$ & \multicolumn{3}{l}{dense matrix with i.i.d.\ $\mathcal N(0,1)$ entries (Section~\ref{sec:dense})}\\
\midrule
\nhead{Vectors and pairs, for a fixed $D$}
$u,v$ & unit vectors; BG stores $u$, queries $v$ & $\rho=\cos\theta$ & $\inner uv$, with $\theta\in[0,\pi]$\\
$\xt$ & preconditioned vector $HDx$ & $w_x$ & cap $n\|\xt\|_\infty^2$\\
$k_x$ & fourth moment $n\|\xt\|_4^4$ & $k_{uv}$ & cross fourth moment $n\|\ut\circ\vt\|^2$\\
$\wmax$, $\kmax$ & $\max\{w_u,w_v\}$, $\max\{k_u,k_v\}$ & $w_\Ufam$, $k_\Ufam$ & maxima of $w_x$, $k_x$ over $x\in\Ufam$\\
$U=[u_1\,u_2]$ & \multicolumn{3}{l}{orthonormal basis of $\mathrm{span}\{u,v\}$ with $u_1=u$ and $v=Ut_\theta$ (Section~\ref{sec:random-metric})}\\
$\tilde U$, $\cvec_j$ & $HDU=[\,\ut_1\ \,\ut_2\,]$ with rows $\cvec_j^\top$ & $\wU$, $\kU$ & plane cap, plane fourth moment \eqref{eq:frame}\\
$x_B$, $x_b$ & \multicolumn{3}{l}{normalized Walsh block $s^{-1/2}\mathbf 1_B$ of size $s$; $x_b$ for $B=B_b$ (Section~\ref{sec:lower}, Appendix~\ref{app:lower})}\\
\midrule
\nhead{One support $\capset$, for a fixed $D$}
$\xt_\capset$ & subsampled vector, $(\xt_\capset)_j=q^{-1/2}\xt_j\mathbf 1\{j\in\capset\}$ & $t_\theta$ & $(\cos\theta,\sin\theta)^\top$, local coordinates of $v$\\
$\Sigma_\capset$ & \multicolumn{3}{l}{conditional covariance $\begin{psmallmatrix}\sigma_u^2&\sigma_{uv}\\\sigma_{uv}&\sigma_v^2\end{psmallmatrix}$ of $(X,Y)$, with $\sigma_x^2=\|\xt_\capset\|^2$ and $\sigma_{uv}=\inner{\ut_\capset}{\vt_\capset}$}\\
$M_\capset$ & \multicolumn{3}{l}{random metric $\tilde U_\capset^\top\tilde U_\capset=\begin{psmallmatrix}\sigma_1^2&\sigma_{12}\\\sigma_{12}&\sigma_2^2\end{psmallmatrix}$ \eqref{eq:random-metric}, with $\sigma_1^2=\sigma_u^2$, $\sigma_2^2=\|\ut_{2,\capset}\|^2$, $\sigma_{12}=\inner{\ut_{1,\capset}}{\ut_{2,\capset}}$}\\
$\zeta_\capset$, $\bar\zeta$ & $\sigma_{uv}/\sigma_u=\E[Z\mid\capset]$; $\bar\zeta=\E[\zeta_\capset]$ & $\rho_\capset$, $\theta_\capset$ & $\sigma_{uv}/(\sigma_u\sigma_v)=\cos\theta_\capset$; angle $\angle(\ut_\capset,\vt_\capset)$\\
$\omega$ & $\sigma_{uv}-\rho\sigma_u^2=\sqrt{1-\rho^2}\,\sigma_{12}$ (BG) & $E_\capset$, $\mathcal G$ & $M_\capset-I_2$; $\{\frac12I_2\preceq M_\capset\preceq2I_2\}$ (BB)\\
$\vartheta_\theta(M)$ & \multicolumn{3}{l}{angle between $e_1$ and $t_\theta$ under the inner product $s^\top Ms'$ (Section~\ref{sec:random-metric})}\\
\midrule
\nhead{Estimators and their errors}
$X_i,Y_i$ & row projections $(\Phi u)_i$, $(\Phi v)_i$ & $Z_i$ & BG row score $\sqrt{\pi/2}\,\sign(X_i)\,Y_i$\\
$\hat\rho_{\rm GG}$, $\hat\rho_{\rm BG}$ & GG and BG estimates \eqref{eq:estimators} & $\hat\theta$, $\hat\rho_{\rm BB}$ & BB estimates \eqref{eq:estimators}\\
$\psi$ & Catoni's influence function \eqref{eq:catoni-def} & $\Cat_\alpha$ & Catoni's estimate with scale $\alpha$ \eqref{eq:catoni-def}\\
$\sigma_*^2$ & variance bound (Lemma~\ref{lem:catoni}) & $b_{uv}$ & BG support bias $\E[\hat\rho_{\rm BG}\mid D]-\rho$\\
$V_u$, $V_{uv}$ & $\Var(\sigma_u^2)$, $\Var(\sigma_{uv}-\rho\sigma_u^2)$ (BG cert.) & $V_M$ & $\E\|M_\capset-I_2\|_F^2$ (BB certificate)\\
$\Delta$ & \multicolumn{3}{l}{relative self-pair BG bias $\Delta(u)=1-\E\sigma_u$; $\Delta_s$ on Walsh blocks of size $s$ (Theorem~\ref{thm:lower})}\\
\midrule
\nhead{Functions}
$h(t)$ & Bennett's function $(1+t)\log(1+t)-t$ & $\Phi_0$ & standard normal distribution function\\
\bottomrule
\end{tabular}
\caption{Summary of notation used across multiple sections. Local variables used only within a single statement or proof (such as $t$, $\lambda$, $\tau$, $R$, and $V$) are defined where they appear.}
\label{tab:notation}
\end{table}

\subsection{Tools}\label{app:tools}

In this section, we collect the probabilistic and statistical tools used throughout the proofs. After stating standard concentration inequalities for sums of independent scalars and symmetric matrices, we establish the coupling principle that transfers these inequalities to sampling without replacement under balanced supports (Lemma~\ref{lem:transfer}), followed by the high-probability guarantees for Catoni's robust mean estimator and its fast one-step variants (Lemmas~\ref{lem:catoni} and~\ref{lem:catoni-uncentered}).

\paragraph{Concentration of sums of independent scalar random variables.}
\emph{Bennett and Bernstein inequalities.} Let $Z:=\sum_jZ_j$ be a finite sum of independent, centered real random variables satisfying $Z_j\le R$ almost surely for some $R>0$ and $\sum_j\E Z_j^2\le V$. Define Bennett's function $h(t):=(1+t)\log(1+t)-t$ for $t\ge-1$, with $h(-1):=1$. For all $\lambda\ge0$ and $t\ge0$, Bennett's inequality (first line) and Bernstein's inequality (second line) bound the cumulant generating function and upper tail of $Z$:
\begin{align}
\log\E e^{\lambda Z}&\le\frac V{R^2}\big(e^{\lambda R}-1-\lambda R\big),&
\Pr[Z\ge t]&\le\exp\Big\{-\frac V{R^2}\,h\Big(\frac{Rt}V\Big)\Big\};
\label{eq:app-bennett}\\
\log\E e^{\lambda Z}&\le\frac{\lambda^2V}{2(1-\lambda R/3)},&
\Pr[Z\ge t]&\le\exp\Big\{-\frac{t^2}{2(V+Rt/3)}\Big\}
&&(\lambda<3/R).
\label{eq:app-bernstein}
\end{align}
These are Theorem~2.9, Theorem~2.10, and Corollary~2.11 of \citet{boucheron2013concentration} (noting that all four bounds are monotone increasing in $V$). When $|Z_j|\le R$, the same bounds apply to $-Z$. We use Bennett's inequality in Lemma~\ref{lem:sampling}, and Bernstein's inequality in Appendix~\ref{app:bg} and Theorems~\ref{thm:dense-bb}(b) and~\ref{thm:bb-param}(b).

\emph{Maurer's lower-tail inequality.} When summands are nonnegative, their lower tail concentrates without requiring an upper bound on their range. Let $W_j$ be finitely many independent \emph{nonnegative} random variables with $\E W_j^2<\infty$, and set $Z:=\sum_j(W_j-\E W_j)$. For all $\lambda\ge0$ and $t\ge0$,
\begin{equation}
\log\E e^{-\lambda Z}\le\frac{\lambda^2}2\sum_j\E W_j^2,\qquad
\Pr[Z\le-t]\le\exp\Big\{-\frac{t^2}{2\sum_j\E W_j^2}\Big\}.
\label{eq:app-maurer}
\end{equation}
The tail bound is Theorem~2.1 of \citet{maurer2003bound}, and the cumulant bound is the first step of its proof. We use Maurer's inequality in Lemma~\ref{lem:sampling} and Appendix~\ref{app:gg}.

\emph{Hoeffding's inequality.} Let $Z_j$ be finitely many independent real random variables, where each $Z_j$ takes values in an interval $I_j$ of length $|I_j|$. For any $t>0$,
\begin{equation}
\Pr\Big[\sum_j(Z_j-\E Z_j)\ge t\Big]\le\exp\Big\{-\frac{2t^2}{\sum_j|I_j|^2}\Big\},
\label{eq:app-hoeffding}
\end{equation}
and the same bound holds for the lower tail \citep[Thm.~2]{hoeffding1963probability}. We apply Hoeffding's inequality in Lemma~\ref{lem:flat} and Theorems~\ref{thm:dense-bb}(a) and~\ref{thm:bb-param}(a).

\paragraph{Concentration of sums of independent $2\times2$ matrices.}
Let $\lambda_{\min}(A)$, $\lambda_{\max}(A)$, and $\|A\|_{\rm op}$ denote the minimum eigenvalue, maximum eigenvalue, and operator norm of a symmetric matrix $A$, respectively, and let $\mathbf Z:=\sum_j\mathbf Z_j$ be a finite sum of independent random symmetric $2\times2$ matrices.
\begin{enumerate}
\item[(a)] \emph{Matrix Chernoff.} If $\mathbf Z_j\succeq0$, $\lambda_{\max}(\mathbf Z_j)\le R$ almost surely, and $\E\mathbf Z=I_2$, then with Bennett's function $h$,
\begin{equation}
\begin{aligned}
\Pr[\lambda_{\min}(\mathbf Z)\le1-\tau]&\le2e^{-h(-\tau)/R}&&(0\le\tau\le1),\\
\Pr[\lambda_{\max}(\mathbf Z)\ge1+\tau]&\le2e^{-h(\tau)/R}&&(\tau\ge0).
\end{aligned}
\label{eq:app-matrix}
\end{equation}
\item[(b)] \emph{Matrix Bernstein.} If $\E\mathbf Z_j=0$, $\|\mathbf Z_j\|_{\rm op}\le R$ almost surely, and $\|\sum_j\E\mathbf Z_j^2\|_{\rm op}\le V$, then for all $t\ge0$,
\begin{equation}
\Pr\big[\|\mathbf Z\|_{\rm op}\ge t\big]\le4\exp\Big\{-\frac{t^2}{2(V+Rt/3)}\Big\}.
\label{eq:app-matrix-bernstein}
\end{equation}
\end{enumerate}
Part~(a) is \citet[Thm.~1.1]{tropp2012user} specialized to dimension $2$ and rewritten using the identity $e^{-h(\pm\tau)}=e^{\pm\tau}(1\pm\tau)^{-(1\pm\tau)}$, while part~(b) is \citet[Thm.~1.4]{tropp2012user} applied to $\mathbf Z_j$ and $-\mathbf Z_j$. Both are used in Appendix~\ref{app:bb}.

\paragraph{Balanced supports and the transfer principle.}
Although the concentration inequalities above assume independent summands, a balanced support $\capset$ of size $r$ is drawn without replacement. By a classical theorem of \citet[Thm.~4]{hoeffding1963probability}, passing from sampling without replacement to sampling with replacement (independent draws) can only increase the expectation of any convex function of the sum. The lemma below, which follows the exchangeability coupling of \citet[Lemma~2]{gross2010note}, establishes this dominance on any real vector space, covering both scalars and symmetric matrices.

\begin{lemma}[Sampling without replacement]\label{lem:transfer}
Let $y_1,\dots,y_n$ be elements of a real vector space and let $1\le r\le n$. Set $T:=\sum_{j\in\capset}y_j$, where $\capset$ is drawn uniformly from the $\binom nr$ subsets of $[n]=\{1,\dots,n\}$ of size $r$, and set $T^*:=\sum_{m=1}^ry_{J_m}$, where $J_1,\dots,J_r$ are drawn independently and uniformly from $[n]$. Then $\capset$ and $(J_1,\dots,J_r)$ can be coupled on a common probability space such that $\E[T^*\mid\capset]=T$. Consequently, $\E f(T)\le\E f(T^*)$ for every convex function $f$.
\end{lemma}
\begin{proof}[Proof of Lemma~\ref{lem:transfer}]
First draw $J_1,\dots,J_r$ independently and uniformly from $[n]$, and then draw $\capset$ uniformly from all subsets of $[n]$ of size $r$ that contain the set of distinct indices $\{J_1,\dots,J_r\}$. Because this joint construction is invariant under arbitrary permutations of $[n]$, the marginal distribution of $\capset$ is uniform over all $\binom nr$ subsets of size $r$, and conditioned on $\capset$, the multiplicity counts $N_j:=\#\{m:J_m=j\}$ for $j\in\capset$ are exchangeable. Since $\sum_{j\in\capset}N_j=r$, exchangeability implies $\E[N_j\mid\capset]=1$ for every $j\in\capset$. Consequently, $\E[T^*\mid\capset]=\sum_{j\in\capset}\E[N_j\mid\capset]\,y_j=\sum_{j\in\capset}y_j=T$. Because $T^*$ takes only finitely many values given $\capset$, conditional Jensen's inequality yields $f(T)\le\E[f(T^*)\mid\capset]$, and taking expectations over $\capset$ completes the proof.
\end{proof}

Every tail bound in \eqref{eq:app-bennett}--\eqref{eq:app-matrix-bernstein} is derived by bounding the expectation of a convex exponential functional of the sum---namely $e^{\lambda T}$ in the scalar case, and $\operatorname{tr}e^{\lambda(T-B)}$ for a fixed symmetric matrix $B$ in the matrix case (via the Laplace transform method $\Pr[\lambda_{\max}(\pm\mathbf Z)\ge t]\le e^{-\lambda t}\,\E\operatorname{tr}e^{\pm\lambda\mathbf Z}$ for $\lambda>0$; see \citealp[Prop.~3.1]{tropp2012user})---and applying Markov's inequality. Because both functionals are convex in $T$ for all $\lambda\in\R$, Lemma~\ref{lem:transfer} immediately yields the \emph{transfer principle} used throughout our analysis of balanced supports:
\begin{quote}
\emph{Every concentration inequality in \eqref{eq:app-bennett}--\eqref{eq:app-matrix-bernstein} holds for the without-replacement sum $T=\sum_{j\in\capset}y_j$ over a balanced support whenever its hypotheses hold for the $r$ independent with-replacement draws $y_{J_1},\dots,y_{J_r}$ (centering both $T$ and the individual draws at their expectations when centered summands are required).}
\end{quote}
Under this transfer principle, the range parameter $R$ and variance parameter $V$ are simply those of $r$ independent draws with replacement (ignoring the finite-population variance reduction factor $f_q$). We also invoke Lemma~\ref{lem:transfer} for polynomial moments such as $f(T)=(T-c)^4$ in Appendix~\ref{app:bg}.

\subsection{Catoni Decoder}\label{app:catoni}

\paragraph{Catoni's robust mean estimator.}
Recall from Section~\ref{sec:dense} that Catoni's estimator $\Cat_\alpha(W_1,\dots,W_\ell)$ with scale parameter $\alpha>0$ is the root $c$ of $\sum_{i=1}^\ell\psi(\alpha(W_i-c))=0$, where $\psi(x)=\sign(x)\log(1+|x|+x^2/2)$ is the influence function defined in \eqref{eq:catoni-def}. The guarantees in part~(b) of Theorems~\ref{thm:gg-param} and~\ref{thm:bg-param} rely only on the fact that $\psi$ is continuous, nondecreasing, and sandwiched between two quadratic logarithms, which we call its \emph{envelope}:
\begin{equation}
-\log\big(1-x+\tfrac{x^2}2\big)\le\psi(x)\le\log\big(1+x+\tfrac{x^2}2\big)\qquad(x\in\R).
\label{eq:catoni-envelope}
\end{equation}
Notice that Catoni's $\psi(x)$ coincides with the upper envelope for $x\ge0$ and the lower envelope for $x\le0$; since $(1+x+\frac{x^2}2)(1-x+\frac{x^2}2)=1+\frac{x^4}4\ge1$, the lower bound never exceeds the upper bound, making $\psi$ the extremal influence function permitted by the envelope \citep[eq.~(2.2)]{catoni2012challenging}.

For any function $\psi$ satisfying \eqref{eq:catoni-envelope} and any random variable $W$ with mean $c_0$ and variance at most $\sigma^2$, every $c\in\R$ and $\alpha>0$ satisfy
\begin{equation}
\begin{aligned}
\E\exp\big\{\pm\psi\big(\alpha(W-c)\big)\big\}
&\le1\pm\alpha(c_0-c)+\frac{\alpha^2}2\big(\sigma^2+(c_0-c)^2\big)\\
&\le\exp\Big\{\pm\alpha(c_0-c)+\frac{\alpha^2}2\big(\sigma^2+(c_0-c)^2\big)\Big\},
\end{aligned}
\label{eq:catoni-mgf}
\end{equation}
where the $\pm$ signs correspond throughout \citep[Prop.~2.1]{catoni2012challenging}. Indeed, exponentiating \eqref{eq:catoni-envelope} gives $e^{\pm\psi(x)}\le1\pm x+\frac{x^2}2$; substituting $x=\alpha(W-c)$, taking expectations using $\E(W-c)=c_0-c$ and $\E(W-c)^2\le\sigma^2+(c_0-c)^2$, and applying $1+y\le e^y$ yields \eqref{eq:catoni-mgf}. The next two lemmas formulate Catoni's guarantee simultaneously across $N$ estimation \emph{tasks} (where each task corresponds to a vector pair and its $\ell$ samples are the pair's $\ell$ row scores).

\begin{lemma}[Catoni's estimate \citep{catoni2012challenging}]\label{lem:catoni}
Let $\delta\in(0,1)$ and $\eps>0$, and consider $N$ estimation tasks $t=1,\dots,N$. For each task $t$, let $W_{t,1},\dots,W_{t,\ell}$ be independent random variables with common mean $c_t$ and variances at most $\sigma_*^2$ (samples across different tasks may be arbitrarily dependent). Set $L:=\log(2N/\delta)$. If
\[
\ell\ \ge\ 2(\sigma_*^2+\eps^2)\,\eps^{-2}L\qquad\text{and}\qquad\alpha=\frac{\eps}{\sigma_*^2+\eps^2},
\]
then with probability at least $1-\delta$, all $N$ tasks simultaneously satisfy $|\Cat_\alpha(W_{t,1},\dots,W_{t,\ell})-c_t|\le\eps$.

\emph{Deviation form.} If $\sigma_*^2>0$ and $\ell>2L$, the same guarantee holds with $\eps=s:=\sqrt{2\sigma_*^2L/(\ell-2L)}$ and $\alpha=s/(\sigma_*^2+s^2)$.

Both statements remain valid (for any root $c$) when $\psi$ in \eqref{eq:catoni-def} is replaced by any continuous nondecreasing function satisfying the envelope \eqref{eq:catoni-envelope}.
\end{lemma}
For a single task and Catoni's $\psi$, the deviation form is \citet[Prop.~2.4]{catoni2012challenging}. We include the short proof below to cover multiple tasks, fixed target accuracy $\eps$, and general envelope functions $\psi$, as well as to prepare for the scale-free distance estimator of Theorem~\ref{thm:gg-dist}.

\begin{proof}[Proof of Lemma~\ref{lem:catoni}]
Fix a task $t$, abbreviate $W_i:=W_{t,i}$, and set $\Psi(c):=\sum_{i=1}^\ell\psi(\alpha(W_i-c))$ for any continuous nondecreasing function $\psi$ satisfying \eqref{eq:catoni-envelope}. Then $\Psi$ is continuous and nonincreasing in $c$. Because the lower bound in \eqref{eq:catoni-envelope} is strictly positive on $(0,2)$, monotonicity implies $\psi(x)>0$ for all $x>0$ and $\psi(x)<0$ for all $x<0$. Thus $\Psi(c)>0$ for $c<\min_i W_i$ and $\Psi(c)<0$ for $c>\max_i W_i$, guaranteeing the existence of at least one root $\hat c$ (which is unique when $\psi$ is strictly increasing, as in \eqref{eq:catoni-def}).

If $\hat c>c_t+\eps$, monotonicity gives $\Psi(c_t+\eps)\ge\Psi(\hat c)=0$. By Markov's inequality, independence across $i=1,\dots,\ell$, and \eqref{eq:catoni-mgf} (applied with the $+$ sign at $c=c_t+\eps$), any $\alpha>0$ satisfies
\begin{equation}
\Pr[\hat c>c_t+\eps]\le\E e^{\Psi(c_t+\eps)}=\prod_{i=1}^\ell\E e^{\psi(\alpha(W_i-c_t-\eps))}
\le\Big(1-\alpha\eps+\frac{\alpha^2}2(\sigma_*^2+\eps^2)\Big)^\ell.
\label{eq:catoni-tail}
\end{equation}
Symmetrically, if $\hat c<c_t-\eps$, then $\Psi(c_t-\eps)\le0$, and applying \eqref{eq:catoni-mgf} with the $-$ sign at $c=c_t-\eps$ bounds $\Pr[\hat c<c_t-\eps]\le\E e^{-\Psi(c_t-\eps)}$ by the exact same quantity. Choosing $\alpha=\eps/(\sigma_*^2+\eps^2)$ minimizes the base in \eqref{eq:catoni-tail}, reducing it to $1-\eps^2/(2(\sigma_*^2+\eps^2))\le\exp\{-\eps^2/(2(\sigma_*^2+\eps^2))\}$. The condition on $\ell$ ensures that each one-sided tail probability is at most $e^{-L}=\delta/(2N)$, and a union bound over all $2N$ tails establishes the first claim. The deviation form follows immediately by setting $\eps=s$, which solves $\ell=2(\sigma_*^2+s^2)s^{-2}L$.
\end{proof}

\paragraph{Faster evaluation.}
While finding the exact root of \eqref{eq:catoni-def} requires a few passes over the $\ell$ samples of each pair and evaluates a logarithm per sample, three simple techniques eliminate this overhead in practice:

\emph{A logarithm-free influence function.} The clamped cubic polynomial $\psi_3(x):=x-x^3/6$ for $|x|\le\sqrt2$ (and constant $\pm2\sqrt2/3$ outside $[-\sqrt2,\sqrt2]$) is continuous, nondecreasing, and satisfies the envelope \eqref{eq:catoni-envelope} \citep[eq.~(1) and Lemma~3.1]{catoni2017dimension}. By the final clause of Lemma~\ref{lem:catoni}, $\psi_3$ can be substituted directly for $\psi$, replacing transcendental logarithms with two multiplications per score.

\emph{Closed-form evaluation without root-finding.} Evaluating the influence function just once at $c=0$ yields the non-iterative estimator $(\alpha\ell)^{-1}\sum_i\psi(\alpha W_i)$, which trades the variance $\sigma^2$ for the raw second moment $\E W_i^2$. The following guarantee is the scalar specialization of \citet[Thm.~3.1]{minsker2018subgaussian}.

\begin{lemma}[Catoni's estimate without a root, after \citet{minsker2018subgaussian}]\label{lem:catoni-uncentered}
In the setting of Lemma~\ref{lem:catoni}, suppose the samples of task $t$ have mean $c_t$ and raw second moment $\E W_{t,i}^2\le\nu_*$ for some $\nu_*>0$, and let $\psi$ be nondecreasing and satisfy \eqref{eq:catoni-envelope}. Setting $\alpha=\sqrt{2L/(\ell\nu_*)}$, with probability at least $1-\delta$ simultaneously across all $N$ tasks,
\[
\Big|\frac1{\alpha\ell}\sum_{i=1}^\ell\psi(\alpha W_{t,i})-c_t\Big|\ \le\ \sqrt{\frac{2\nu_*L}\ell}.
\]
\end{lemma}
\begin{proof}[Proof of Lemma~\ref{lem:catoni-uncentered}]
Fix a task $t$ and write $W_i:=W_{t,i}$. Because $\sigma^2+c_t^2=\E W_i^2\le\nu_*$, evaluating \eqref{eq:catoni-mgf} at $c=0$ yields $\E e^{\pm\psi(\alpha W_i)}\le\exp\{\pm\alpha c_t+\frac{\alpha^2}2\nu_*\}$. By independence and Markov's inequality, each of the two events $\pm\big(\sum_{i=1}^\ell\psi(\alpha W_i)-\alpha\ell c_t\big)\ge\frac{\alpha^2}2\ell\nu_*+L$ has probability at most $e^{-L}=\delta/(2N)$. Outside the union of these $2N$ events, every task satisfies $|(\alpha\ell)^{-1}\sum_i\psi(\alpha W_{t,i})-c_t|\le\frac\alpha2\nu_*+\frac L{\alpha\ell}$, and substituting $\alpha=\sqrt{2L/(\ell\nu_*)}$ minimizes the right-hand side to give $\sqrt{2\nu_*L/\ell}$.
\end{proof}

For BG, this non-iterative estimator sacrifices no statistical accuracy at all: the variance bound $\sigma_*^2=\pi/2$ used in Theorem~\ref{thm:bg-param}(b) is already the raw second moment $\E[Z_i^2\mid D]=\pi/2$. Setting $\nu_*=\pi/2$ and $\alpha=\sqrt{4L/(\pi\ell)}$ tightens the error radius from $\sqrt{\pi L/(\ell-2L)}$ to $\sqrt{\pi L/\ell}$, removing the additive $+2$ term in Corollary~\ref{cor:bg-classical}(ii). Moreover, because $\psi$ is odd, $\psi(\alpha Z_i)=\sign(X_i)\,\psi(\alpha\sqrt{\pi/2}\,Y_i)$, so the estimator reduces to $\ell^{-1}\sum_i\sign(X_i)\,\phi(Y_i)$ with $\phi(y):=\alpha^{-1}\psi(\alpha\sqrt{\pi/2}\,y)$---a standard inner product between the stored sign code and query projections that have been softly clipped once upon arrival. For GG, the raw second moment $\E(X_iY_i)^2$ exceeds the variance by $\rho^2\le1$; replacing $\sigma_*^2=3$ with $\nu_*=4$ in Corollary~\ref{cor:gg-classical}(ii) requires $8\eps^{-2}L$ rows, which matches the sketch length already specified in the corollary.

\emph{Single Newton step.} When the exact root of \eqref{eq:catoni-def} is desired, initializing at the closed-form estimate above and taking a single Newton step suffices in practice, in the spirit of one-step M-estimators \citep{bickel1975onestep}. Across our sparse GG and BG experiments on the Gaussian pair ($r\in\{8,32\}$, $\ell\in\{2^9,2^{15}\}$, both support laws), one Newton step lands within $2.9\cdot10^{-6}\,s$ of the exact root (where $s$ is the deviation bound of Lemma~\ref{lem:catoni}), and two Newton steps land within $10^{-14}$ in absolute value.

\paragraph{Dense reference bounds for Catoni's decoder.}
Finally, specializing Lemma~\ref{lem:catoni} to a dense Gaussian matrix $\Phi=G$ yields the dense Catoni sketch lengths reported in Table~\ref{tab:budgets}.

\needspace{6\baselineskip}%
\begin{corollary}[Dense inner products, Catoni's estimate]\label{cor:dense-inner}
With $\Phi=G$ in \eqref{eq:estimators}, let $\delta\in(0,1)$, fix $N$ pairs of unit vectors in advance, and let $L:=\log(2N/\delta)$ and $\ell>2L$. For GG, set $s:=\sqrt{4L/(\ell-2L)}$ and $\alpha=s/(2+s^2)$; then with probability at least $1-\delta$, all pairs satisfy $|\Cat_\alpha(X_1Y_1,\dots,X_\ell Y_\ell)-\rho|\le s$. For BG, set $s:=\sqrt{\pi L/(\ell-2L)}$ and $\alpha=s/(\pi/2+s^2)$; then with probability at least $1-\delta$, all pairs satisfy $|\Cat_\alpha(Z_1,\dots,Z_\ell)-\rho|\le s$.
\end{corollary}
\begin{proof}[Proof of Corollary~\ref{cor:dense-inner}]
We apply the deviation form of Lemma~\ref{lem:catoni} with one task per pair. For a dense Gaussian row, $(X_i,Y_i)$ is standard bivariate normal with correlation $\rho$, so the GG products $X_iY_i$ are i.i.d.\ with mean $\rho$ and variance $\E X_i^2Y_i^2-\rho^2=(1+2\rho^2)-\rho^2=1+\rho^2\le2$; taking $\sigma_*^2=2$ gives $s=\sqrt{4L/(\ell-2L)}$. For BG, the scores $Z_i$ are i.i.d.\ with mean $\rho$ and variance $\pi/2-\rho^2\le\pi/2$ (Theorem~\ref{thm:dense-bg}); taking $\sigma_*^2=\pi/2$ gives $s=\sqrt{\pi L/(\ell-2L)}$.
\end{proof}

\subsection{Sampling the coordinates}\label{app:devices}

This section establishes the algebraic relationships among our flatness parameters (Lemma~\ref{lem:params}), proves the two foundational lemmas that underpin all subsequent sections (Lemmas~\ref{lem:random-metric} and~\ref{lem:sampling}), and records the exactness of balanced supports on flat stored vectors (Corollary~\ref{cor:flat-exact}).

\paragraph{Properties of the flatness parameters.}
The following lemma proves the parameter inequalities stated in \eqref{eq:params} and after \eqref{eq:frame}, together with the expectations over $D$ cited in Section~\ref{sec:results}. Let $h_j^\top$ denote the $j$th row of the orthogonal matrix $HD$, so that $h_1,\dots,h_n\in\R^n$ form an orthonormal basis and $\|\xt\|=\|x\|$. In this notation, $\xt_j=\inner{h_j}x$, and $\cvec_j=(\inner{h_j}{u_1},\inner{h_j}{u_2})^\top=U^\top h_j$ is the $j$th row of $\tilde U=[\,\ut_1\ \ \ut_2\,]$ transposed.

\begin{lemma}[Relations between the parameters]\label{lem:params}
Fix $D$. Let $x$, $u$, and $v$ be unit vectors in $\R^n$, and let $U=[\,u_1\ \ u_2\,]\in\R^{n\times2}$ be an orthonormal basis for the plane $\mathrm{span}\{u,v\}$ (with orthogonal projector $UU^\top$).
\begin{enumerate}
\item[(a)] $1\le k_x\le w_x\le n$, with $k_x=1$ if and only if $\xt$ is flat ($\xt_j^2=1/n$ for all $j$).
\item[(b)] $k_{uv}\le\min\{w_u,\sqrt{k_uk_v}\}$.
\item[(c)] $\|\cvec_j\|=\|UU^\top h_j\|$ for every $j$, so $\wU$ and $\kU$ depend only on the subspace $\mathrm{span}\{u_1,u_2\}$ and not on the choice of orthonormal basis $U$. Moreover, $\wU$ equals the maximum of $w_x$ over all unit vectors $x\in\mathrm{span}\{u_1,u_2\}$, and $\frac38\kU$ equals the average of $k_x$ over the unit circle in $\mathrm{span}\{u_1,u_2\}$. Finally, $2\le \wU\le w_{u_1}+w_{u_2}$, $4\le \kU\le2\wU\le2n$, and $\kU\le2(k_{u_1}+k_{u_2})$.
\item[(d)] When $D$ is drawn at random and $x$ and $\mathrm{span}\{u_1,u_2\}$ are fixed independently of $D$, we have $\E_Dk_x=3-2\|x\|_4^4\le3$ and $\E_D\kU=8-2\sum_{i=1}^n(UU^\top)_{ii}^2\le8$.
\end{enumerate}
\end{lemma}

\begin{proof}[Proof of Lemma~\ref{lem:params}]
(a) Applying the Cauchy--Schwarz inequality to the vector $(\xt_j^2)_{j=1}^n$ and the all-ones vector gives $1=\|\xt\|^4\le n\,\|\xt\|_4^4=k_x$, with equality if and only if $\xt_j^2=1/n$ for all $j$. Moreover, $k_x=n\|\xt\|_4^4\le n\|\xt\|_\infty^2\|\xt\|^2=w_x$, and $w_x\le n$ because $\|\xt\|_\infty\le\|\xt\|=1$.

(b) Similarly, $k_{uv}=n\|\ut\circ\vt\|^2\le n\|\ut\|_\infty^2\|\vt\|^2=w_u$, while Cauchy--Schwarz yields $\|\ut\circ\vt\|^2\le\|\ut\|_4^2\|\vt\|_4^2$, so that $k_{uv}\le\sqrt{k_uk_v}$.

(c) Because $\cvec_j=U^\top h_j$ and $U$ has orthonormal columns, $\|\cvec_j\|=\|UU^\top h_j\|$ for every $j$. Any unit vector in the plane takes the form $x=\cos\phi\,u_1+\sin\phi\,u_2=Ut_\phi$ for some angle $\phi\in[0,2\pi)$, giving $\xt_j=t_\phi^\top\cvec_j$ with $t_\phi=(\cos\phi,\sin\phi)^\top$. Writing $\cvec_j=\|\cvec_j\|\,t_{\phi_j}$ in polar coordinates yields $\xt_j=\|\cvec_j\|\cos(\phi-\phi_j)$. Consequently, $\max_\phi\xt_j^2=\|\cvec_j\|^2$, so $\max_{\|x\|=1,x\in\mathrm{span}\{u_1,u_2\}}w_x=n\max_j\|\cvec_j\|^2=\wU$. Likewise, because $\frac1{2\pi}\int_0^{2\pi}\cos^4(\phi-\phi_j)\,d\phi=\frac38$, averaging $k_x=n\sum_j\|\cvec_j\|^4\cos^4(\phi-\phi_j)$ over $\phi\in[0,2\pi)$ gives $\frac38n\sum_j\|\cvec_j\|^4=\frac38\kU$.

The remaining inequalities follow from $\sum_j\|\cvec_j\|^2=\|\ut_1\|^2+\|\ut_2\|^2=2$. By Cauchy--Schwarz, $4=\big(\sum_j\|\cvec_j\|^2\big)^2\le n\sum_j\|\cvec_j\|^4=\kU$. Factoring out the maximum gives $\kU\le n\max_j\|\cvec_j\|^2\sum_j\|\cvec_j\|^2=2\wU$, and $\wU\le n$ since $\|UU^\top h_j\|\le\|h_j\|=1$. Because the maximum of $n$ numbers is at least their average, $\wU=n\max_j\|\cvec_j\|^2\ge\sum_j\|\cvec_j\|^2=2$. Triangle inequalities on the coordinates yield $\wU\le n\max_j\ut_{1,j}^2+n\max_j\ut_{2,j}^2=w_{u_1}+w_{u_2}$, and summing $\big(\ut_{1,j}^2+\ut_{2,j}^2\big)^2\le2\ut_{1,j}^4+2\ut_{2,j}^4$ over $j$ gives $\kU\le2(k_{u_1}+k_{u_2})$.

(d) For any vector $\varsigma\in\{\pm1\}^n$ of independent Rademacher signs and any symmetric matrix $A\in\R^{n\times n}$, expanding $\varsigma^\top A\varsigma=\operatorname{tr}A+2\sum_{i<i'}A_{ii'}\varsigma_i\varsigma_{i'}$ and noting that the distinct-pair products $\{\varsigma_i\varsigma_{i'}\}_{i<i'}$ are uncorrelated with mean zero and unit variance gives $\E\,(\varsigma^\top A\varsigma)^2=(\operatorname{tr}A)^2+4\sum_{i<i'}A_{ii'}^2=(\operatorname{tr}A)^2+2\|A\|_F^2-2\sum_iA_{ii}^2$. For each fixed row $j$, the entries of $h_j$ are $H_{ji}D_{ii}=\varsigma_i/\sqrt n$, where $\varsigma_i:=\sqrt n\,H_{ji}D_{ii}$ are independent Rademacher signs. Thus $\xt_j^2=\inner{h_j}x^2=\varsigma^\top(xx^\top)\varsigma/n$ and $\|\cvec_j\|^2=\|UU^\top h_j\|^2=\varsigma^\top(UU^\top)\varsigma/n$. Setting $A=xx^\top$ (for which $\operatorname{tr}A=\|A\|_F=1$ and $A_{ii}=x_i^2$) gives $\E_D\xt_j^4=(3-2\|x\|_4^4)/n^2$, while setting $A=UU^\top$ (for which $\operatorname{tr}A=\|A\|_F^2=2$) gives $\E_D\|\cvec_j\|^4=(8-2\sum_i(UU^\top)_{ii}^2)/n^2$. Summing over $j=1,\dots,n$ and multiplying by $n$ establishes both identities.
\end{proof}

\paragraph{Proof of the random-metric reduction.}
We now prove Lemma~\ref{lem:random-metric} from Section~\ref{sec:random-metric}.

\begin{proof}[Proof of Lemma~\ref{lem:random-metric}]
\emph{(i) Conditional distribution of $(X,Y)$.} Any vector $x=s_1u_1+s_2u_2=Us$ in the plane has preconditioned coordinates $\xt_j=s^\top\cvec_j$; in particular, $e_1^\top\cvec_j=\ut_j$ and $t_\theta^\top\cvec_j=\vt_j$ since $v=\cos\theta\,u_1+\sin\theta\,u_2$ (including when $|\rho|=1$, where $\sin\theta=0$). Let $(e_1,t_\theta)$ denote the $2\times2$ matrix with columns $e_1$ and $t_\theta$. Conditioned on $\capset$, the nonzero weights $g_j$ ($j\in\capset$) are i.i.d.\ $\mathcal N(0,1/q)$, so writing $\xi_j:=\sqrt q\,g_j\sim\mathcal N(0,1)$ for $j\in\capset$ and drawing independent $\xi_j\sim\mathcal N(0,1)$ for $j\notin\capset$ gives $g=(\xi_1,\dots,\xi_n)^\top\sim\mathcal N(0,I_n)$ with $X=\sum_{j\in\capset}\xi_j\,q^{-1/2}\ut_j=\inner g{\ut_\capset}$ and $Y=\inner g{\vt_\capset}$. Thus $(X,Y)$ has the exact conditional law of $(\inner g{\ut_\capset},\inner g{\vt_\capset})$, which is centered bivariate normal with covariance
\begin{equation}
\Sigma_\capset:=\Cov\big((X,Y)^\top\mid\capset\big)=\begin{pmatrix}\|\ut_\capset\|^2&\inner{\ut_\capset}{\vt_\capset}\\\inner{\ut_\capset}{\vt_\capset}&\|\vt_\capset\|^2\end{pmatrix}=(e_1,t_\theta)^\top M_\capset\,(e_1,t_\theta)=\begin{pmatrix}\sigma_u^2&\sigma_{uv}\\\sigma_{uv}&\sigma_v^2\end{pmatrix}.
\label{eq:app-cov-factor}
\end{equation}

\emph{(ii) Expected metric.} Under both Bernoulli and balanced supports, each coordinate $j\in[n]$ belongs to $\capset$ with marginal probability $q$, so $\E M_\capset=\sum_j\cvec_j\cvec_j^\top=\tilde U^\top\tilde U=I_2$. Reading off entries gives $\E\sigma_1^2=\E\sigma_2^2=1$ and $\E\sigma_{12}=0$, and linearity gives $\E\sigma_u^2=1$, $\E\sigma_{uv}=e_1^\top t_\theta=\rho$, and $\E\sigma_v^2=\|t_\theta\|^2=1$.

\emph{(iii) Conditional score expectations.} From \eqref{eq:app-cov-factor}, $\E[XY\mid\capset]=\sigma_{uv}$. For BG and BB, part~(i) allows us to replace $(X,Y)$ with $(\inner g{\ut_\capset},\inner g{\vt_\capset})$ for $g\sim\mathcal N(0,I_n)$. For the BG score $Z=\sqrt{\pi/2}\,\sign(X)\,Y$, suppose first that $\sigma_u>0$, so $\ut_\capset\ne0$. If $\vt_\capset\ne0$ as well, the second Gaussian identity in \eqref{eq:dense-identities} yields $\E[\sign(X)\,Y\mid\capset]=\sqrt{2/\pi}\,\inner{\ut_\capset}{\vt_\capset}/\|\ut_\capset\|$, so $\E[Z\mid\capset]=\sigma_{uv}/\sigma_u=\zeta_\capset$; if $\vt_\capset=0$, then $Y=0$ and $\sigma_{uv}=0$, so $\E[Z\mid\capset]=0=\zeta_\capset$. Finally, if $\sigma_u=0$, then $X=0$ and $\sign X=+1$ by the convention $\signzero$, giving $\E[Z\mid\capset]=\sqrt{\pi/2}\,\E[Y\mid\capset]=0=\zeta_\capset$.

For BB, we check the three cases defining $\theta_\capset$. When $\sigma_u\sigma_v>0$, both $\ut_\capset$ and $\vt_\capset$ are nonzero, and the third identity in \eqref{eq:dense-identities} gives $\Pr[\sign X\ne\sign Y\mid\capset]=\angle(\ut_\capset,\vt_\capset)/\pi$ with $\cos\angle(\ut_\capset,\vt_\capset)=\sigma_{uv}/(\sigma_u\sigma_v)=\rho_\capset=\cos\theta_\capset$. When exactly one of $\sigma_u$ or $\sigma_v$ vanishes, that projection is identically zero (with sign $+1$) while the other is a nondegenerate centered normal variable (negative with probability $\frac12=\theta_\capset/\pi$). When $\sigma_u=\sigma_v=0$, both projections vanish and have sign $+1$, so they disagree with probability $0=\theta_\capset/\pi$. Whenever $M_\capset\succ0$, we have $\sigma_u,\sigma_v>0$ and $\cos\vartheta_\theta(M_\capset)=\sigma_{uv}/(\sigma_u\sigma_v)=\cos\theta_\capset$, so $\theta_\capset=\vartheta_\theta(M_\capset)$ since both lie in $[0,\pi]$.

\emph{(iv) Gram identity.} Because $\det(e_1,t_\theta)=\sin\theta$, taking determinants in \eqref{eq:app-cov-factor} gives $\sigma_u^2\sigma_v^2-\sigma_{uv}^2=\det M_\capset\,\sin^2\theta=(1-\rho^2)\det M_\capset\ge0$, which implies $|\sigma_{uv}|\le\sigma_u\sigma_v$. For any $M\succ0$, dividing this identity by $(e_1^\top Me_1)(t_\theta^\top Mt_\theta)>0$ gives
\[
\sin^2\vartheta_\theta(M)=1-\frac{(e_1^\top Mt_\theta)^2}{(e_1^\top Me_1)(t_\theta^\top Mt_\theta)}=\frac{\sin^2\theta\,\det M}{(e_1^\top Me_1)(t_\theta^\top Mt_\theta)}\le\sin^2\theta\,\frac{\lambda_{\max}(M)}{\lambda_{\min}(M)} ,
\]
since $\det M=\lambda_{\max}(M)\lambda_{\min}(M)$ and both Rayleigh quotients in the denominator are at least $\lambda_{\min}(M)$. Taking square roots (as $\theta,\vartheta_\theta(M)\in[0,\pi]$) completes the proof.
\end{proof}

\paragraph{Moments and tails of coordinate sums over a random support.}
Every entry of $M_\capset$---as well as linear combinations such as $\sigma_{uv}$ and $\sigma_v^2$---has the form $T=q^{-1}\sum_{j\in\capset}y_j$ for a deterministic vector $y\in\R^n$. The next lemma establishes the covariance of any two such sums and bounds both tails of the captured energy $\sigma_x^2$.

\begin{lemma}[Sums over a random support, both laws]\label{lem:sampling}
Fix $D$, and let $\capset$ be the support of a single row under either support law. For any $y,y'\in\R^n$, set $T:=q^{-1}\sum_{j\in\capset}y_j$ and $T':=q^{-1}\sum_{j\in\capset}y'_j$. Then $\E T=\sum_jy_j$ and
\[
\Cov(T,T')=\frac{f_q}{q}\Big(\inner y{y'}-\frac{\bal}{n}\sum_jy_j\sum_jy'_j\Big).
\]
In particular, for any unit vector $x$ with captured energy $\sigma_x^2=\|\xt_\capset\|^2$:
\begin{enumerate}
\item[(i)] $\E\sigma_x^2=1$, $\Var(\sigma_x^2)=f_q(k_x-\bal)/r\le w_x/r$, and $\sigma_x^2\le n/r$ almost surely;
\item[(ii)] $\Pr[\sigma_x^2<t]\le\exp\{-(1-t)^2r/(2k_x)\}$ for all $t\in(0,1)$;
\item[(iii)] $\Pr[\sigma_x^2\ge1+t]\le\exp\{-(r/w_x)\,h(t)\}$ for all $t>0$, where $h(t)=(1+t)\log(1+t)-t$;
\item[(iv)] as shown in Lemma~\ref{lem:params}, $k_x\ge1$ (with equality if and only if $\xt$ is flat), and $\E_Dk_x=3-2\|x\|_4^4\le3$ when $x$ is fixed independently of $D$;
\item[(v)] \emph{Domination by independent summands.} Define the auxiliary sum $S_*$ by
\begin{equation}
\begin{aligned}
S_*&:=\sum_{j=1}^nW_j,&W_j&:=\mathbf 1\{j\in\capset\}\,\xt_j^2/q&&\text{(Bernoulli supports)},\\
S_*&:=\sum_{m=1}^rW_m,&W_m&:=\xt_{J_m}^2/q&&\text{(balanced supports)},
\end{aligned}
\label{eq:app-indep}
\end{equation}
where $J_1,\dots,J_r$ are drawn independently and uniformly from $[n]$. Under both laws, the summands of $S_*$ are mutually independent, take values in $[0,w_x/r]$, have means summing to $1$, and have second moments summing to $k_x/r$. Moreover, $\E f(\sigma_x^2)\le\E f(S_*)$ for every convex function $f:\R\to\R$.
\end{enumerate}
\end{lemma}

\begin{proof}[Proof of Lemma~\ref{lem:sampling}]
\emph{Covariance identity.} Writing $\xi_j:=\mathbf 1\{j\in\capset\}$ gives $T=q^{-1}\sum_j\xi_jy_j$. Under either support law, $\E\xi_j=q$ (so $\E T=\sum_jy_j$) and $\Var\xi_j=q(1-q)$. Under Bernoulli supports, the indicators $\xi_j$ are independent, so $\Cov(\xi_j,\xi_{j'})=0$ for $j\ne j'$. Under balanced supports, $\sum_{j=1}^n\xi_j=r$ is constant and the indicators are exchangeable, so $0=\Cov(\xi_j,\sum_{j'}\xi_{j'})=q(1-q)+(n-1)\Cov(\xi_j,\xi_{j'})$ for $j\ne j'$. Using $f_q$ and $\bal$ from Definition~\ref{def:balanced}, both cases combine into $\Cov(\xi_j,\xi_{j'})=q\,f_q\big(\mathbf 1\{j=j'\}-\frac{\bal}n\big)$, and substituting this into $\Cov(T,T')=q^{-2}\sum_{j,j'}y_jy'_{j'}\Cov(\xi_j,\xi_{j'})$ yields the covariance formula.

\emph{(i) Mean, variance, and range of $\sigma_x^2$.} Specializing to $y_j=y'_j=\xt_j^2$ gives $T=T'=\sigma_x^2$, $\sum_jy_j=\|\xt\|^2=1$, and $\inner yy=\|\xt\|_4^4=k_x/n$. Since $qn=r$, the covariance formula yields $\Var(\sigma_x^2)=f_q(k_x-\bal)/r\le w_x/r$ (as $f_q\le1$ and $k_x\le w_x$), while $\sum_{j\in\capset}\xt_j^2\le\|\xt\|^2=1$ implies $\sigma_x^2\le 1/q=n/r$. Part~(iv) simply restates Lemma~\ref{lem:params}(a) and~(d).

\emph{(v) Domination by independent summands.} Under Bernoulli supports, $S_*=\sigma_x^2$ identically, and the independent summands $W_j=\xi_j\xt_j^2/q$ have means $\xt_j^2$ and second moments $\xt_j^4/q$. Under balanced supports, $\sigma_x^2$ is the sum of $r$ draws without replacement from $\{\xt_j^2/q\}_{j=1}^n$, whereas $S_*$ is the sum of $r$ independent draws with replacement, each having mean $(nq)^{-1}=1/r$ and second moment $n^{-1}q^{-2}\|\xt\|_4^4=k_x/r^2$. In both cases, every summand lies in $[0,\|\xt\|_\infty^2/q]=[0,w_x/r]$, the means sum to $\|\xt\|^2=1$, and the second moments sum to $q^{-1}\|\xt\|_4^4=k_x/r$. The convex ordering $\E f(\sigma_x^2)\le\E f(S_*)$ holds with equality for Bernoulli supports and follows from Lemma~\ref{lem:transfer} for balanced supports.

\emph{(ii) and (iii) Tail bounds.} By the transfer principle of Appendix~\ref{app:tools} (and part~(v)), it suffices to prove both tail bounds for the independent sum $S_*$. For the lower tail~(ii), applying Maurer's inequality \eqref{eq:app-maurer} to the nonnegative summands of $S_*$ at deviation $1-t$ gives $\Pr[S_*<t]\le\Pr[1-S_*\ge1-t]\le e^{-(1-t)^2r/(2k_x)}$. For the upper tail~(iii), the centered summands $W-\E W$ are bounded above by $w_x/r$ and have total variance at most $\sum\E W^2=k_x/r\le w_x/r$; applying Bennett's inequality \eqref{eq:app-bennett} with $R=V=w_x/r$ yields $\Pr[S_*\ge1+t]\le\exp\{-(V/R^2)\,h(Rt/V)\}=e^{-(r/w_x)h(t)}$.
\end{proof}

Notice how the covariance formula in Lemma~\ref{lem:sampling} reflects the support law: for $y'=y$ under balanced supports, the parenthetical factor is $n$ times the population variance of $\{y_j\}_{j=1}^n$ multiplied by the finite-population correction $f_q=(n-r)/(n-1)$; under Bernoulli supports, the centering term is absent ($\bal=0$) because the random support size $|\capset|$ adds variance even when $y$ is constant. In particular, when $k_u=1$, balanced supports have zero energy variance.

\begin{corollary}[Flat stored vectors]\label{cor:flat-exact}
Fix $D$. Under balanced supports, if $k_u=1$, then $\sigma_u^2=1$ deterministically on every support. Consequently, conditioned on $D$, we have $\ell^{-1}\|\Phi u\|^2\sim\chi^2_\ell/\ell$ exactly, and $\E[\hat\rho_{\rm BG}\mid D]=\rho$ for every query $v$.
\end{corollary}
\begin{proof}[Proof of Corollary~\ref{cor:flat-exact}]
By Lemma~\ref{lem:params}(a), $k_u=1$ implies $\ut_j^2=1/n$ for all $j$, so every subset $\capset$ of size $r$ captures energy $\sigma_u^2=q^{-1}(r/n)=1$. Conditioned on $D$, the rows of $\Phi$ are independent, and by Lemma~\ref{lem:random-metric}(i) each entry $(\Phi u)_i$ is distributed as $\mathcal N(0,\sigma_{u,i}^2)=\mathcal N(0,1)$ given $\capset_i$ (and hence given $D$), so $\|\Phi u\|^2\sim\chi^2_\ell$. For any query $v$, Lemma~\ref{lem:random-metric}(iii) and~(ii) give $\E[Z_i\mid\capset_i]=\sigma_{uv,i}/\sigma_{u,i}=\sigma_{uv,i}$ and $\E\sigma_{uv,i}=\rho$, whence $\E[\hat\rho_{\rm BG}\mid D]=\rho$.
\end{proof}

By contrast, under Bernoulli supports, a flat vector $u$ has captured energy $\sigma_u^2=|\capset|/r$, which fluctuates with the Binomial$(n,q)$ support size.

\subsection{Full precision (GG)}\label{app:gg}

For the full-precision linear estimator GG, sparse sampling inflates the spread of each row score without introducing any bias (Section~\ref{sec:gg}). Fix $D$ and a unit vector $x$. Conditioned on $D$, the rows of $\Phi$ are i.i.d., and conditioned on the support $\capset_i$, the projection $(\Phi x)_i$ is distributed as $\mathcal N(0,\sigma_{x,i}^2)$, where $\sigma_{x,i}^2=\|\xt_{\capset_i}\|^2$ is the energy captured from $x$ by row $i$ (Lemma~\ref{lem:random-metric}(i) with $u=v=x$). Consequently, the squared projections $(\Phi x)_i^2$ whose average forms $\ell^{-1}\|\Phi x\|^2$ are i.i.d.\ copies of
\begin{equation}\label{eq:gg-row}
T=\sigma_x^2\,\xi^2,\qquad \xi\sim\mathcal N(0,1)\text{ independent of }\sigma_x^2,\qquad
\E\sigma_x^2=1,\quad \Var(\sigma_x^2)=\frac{f_q\,(k_x-\bal)}{r},
\end{equation}
where the moments of $\sigma_x^2$ are given by Lemma~\ref{lem:sampling}(i). Each squared entry is thus a $\chi^2_1$ random variable modulated by a random scale factor $\sigma_x^2$ of unit mean. Catoni's robust decoder requires only the mean and variance of the per-row product $X_iY_i$ (Proposition~\ref{prop:gg-moments}). The standard sample average requires more delicate control: while Maurer's inequality bounds its lower tail using only $\E T^2$, its upper tail requires bounding the moment generating function of $T$ on the high-probability event that no row captures excessive energy (Lemma~\ref{lem:gg-row-mgf}). At the end of this section, we also introduce and analyze a scale-free Catoni estimator for pairwise squared distances (Theorem~\ref{thm:gg-dist}).

Throughout this section, unless $D$ is explicitly drawn at random, we condition on a fixed $D$ and treat admissible vectors as deterministic.

\paragraph{Moments of a row product.}
We first compute the exact mean and variance of a single row product $X_iY_i$ for a pair of unit vectors $u,v$, where $X_i=(\Phi u)_i$ and $Y_i=(\Phi v)_i$.

\begin{proposition}[GG row moments]\label{prop:gg-moments}
Fix $D$ and unit vectors $u,v$. Under either support law, $\E[X_iY_i]=\rho$ and
\[
\Var(X_iY_i)=1+\rho^2+\frac{f_q}{r}\big(3k_{uv}-\bal(1+2\rho^2)\big)\le1+\rho^2+\frac{3f_q\,k_{uv}}r .
\]
In particular, when $v=u$, we have $\Var(X_i^2)=2+3\Var(\sigma_u^2)$.
\end{proposition}

\begin{proof}[Proof of Proposition~\ref{prop:gg-moments}]
Fix a row and omit its subscript $i$. Conditioned on the support $\capset$, the pair $(X,Y)$ is centered bivariate normal with variances $\sigma_u^2$ and $\sigma_v^2$ and covariance $\sigma_{uv}$ (Lemma~\ref{lem:random-metric}(i)). By Isserlis' theorem (Wick's formula), $\E[XY\mid\capset]=\sigma_{uv}$ and $\E[X^2Y^2\mid\capset]=\sigma_u^2\sigma_v^2+2\sigma_{uv}^2$.

Notice that $\sigma_u^2$, $\sigma_v^2$, and $\sigma_{uv}$ are coordinate sums $q^{-1}\sum_{j\in\capset}y_j$ over $\capset$ with $y_j=\ut_j^2$, $\vt_j^2$, and $\ut_j\vt_j$, respectively. Because $HD$ is orthogonal, the coordinate totals of these three vectors are $1$, $1$, and $\rho$. Lemma~\ref{lem:sampling} therefore gives $\E\sigma_u^2=\E\sigma_v^2=1$ and $\E[XY]=\E\sigma_{uv}=\rho$, so that
\[
\Var(XY)=\E[\sigma_u^2\sigma_v^2]+2\,\E\sigma_{uv}^2-\rho^2=1+\rho^2+\Cov(\sigma_u^2,\sigma_v^2)+2\Var(\sigma_{uv}) .
\]
In the covariance formula of Lemma~\ref{lem:sampling}, the inner product $\inner y{y'}$ equals $\|\ut\circ\vt\|^2=k_{uv}/n$ both for the pair $(\sigma_u^2,\sigma_v^2)$ and for $(\sigma_{uv},\sigma_{uv})$, while the product of coordinate totals is $1$ for $(\sigma_u^2,\sigma_v^2)$ and $\rho^2$ for $(\sigma_{uv},\sigma_{uv})$. Using $qn=r$, we obtain
\[
\Cov(\sigma_u^2,\sigma_v^2)=\frac{f_q}r\,(k_{uv}-\bal),\qquad
\Var(\sigma_{uv})=\frac{f_q}r\,(k_{uv}-\bal\,\rho^2).
\]
Substituting these expressions yields the exact variance identity, and dropping the nonnegative term $f_q\bal(1+2\rho^2)/r$ gives the upper bound. Setting $v=u$ ($\rho=1$ and $k_{uu}=k_u$) gives $\Var(X^2)=2+3f_q(k_u-\bal)/r=2+3\Var(\sigma_u^2)$.
\end{proof}

Notice that the variance bound $1+\rho^2+3f_qk_{uv}/r$ used in Theorem~\ref{thm:gg-param}(b) drops only the finite-population correction term $f_q\bal(1+2\rho^2)/r$. Moreover, replacing $k_{uv}$ by the stored cap $w_u$ (as in the final sentence of Theorem~\ref{thm:gg-param}(b)) is tight on worst-case inputs: for the normalized Walsh block $x$ of size $s$ in Theorem~\ref{thm:lower}(a) on the event $\mathcal E$ where $w_x=k_x=s$, the self-pair $u=v=x$ has $\Var(X_i^2)=2+3f_q(s-\bal)/r$, and both $k_{uv}$ and $w_u$ evaluate to $s$.

\paragraph{Analysis of the sample average.}
For $\lambda\ge0$, conditioning on $\sigma_x^2$ in \eqref{eq:gg-row} gives $\E[e^{\lambda T}\mid \sigma_x^2]=\mathcal L(\sigma_x^2)$, where $\mathcal L(t):=(1-2\lambda t)^{-1/2}$ is the $\chi^2_1$ moment generating function for $2\lambda t<1$. Because $\mathcal L(t)$ diverges as $t\uparrow 1/(2\lambda)$, we truncate $\sigma_x^2$ at a threshold $\tau\ge 1$ satisfying $4\lambda\tau\le1$.

\begin{lemma}[Truncated moment generating function of a squared entry]\label{lem:gg-row-mgf}
Let $\sigma_x^2\ge0$ satisfy $\E\sigma_x^2=1$ and $\Var(\sigma_x^2)\le V$, let $\xi\sim\mathcal N(0,1)$ be independent of $\sigma_x^2$, and set $T=\sigma_x^2\,\xi^2$. For any $\lambda\ge0$ and $\tau\ge1$ satisfying $4\lambda\tau\le1$,
\begin{equation}\label{eq:gg-mgf-upper}
\log\E\big[e^{\lambda(T-1)}\mathbf 1\{\sigma_x^2\le \tau\}\big]
\le\frac{\lambda^2}{1-2\lambda}+8.5\,\lambda^2V .
\end{equation}
\end{lemma}

\begin{proof}[Proof of Lemma~\ref{lem:gg-row-mgf}]
Differentiating $\mathcal L(t)=(1-2\lambda t)^{-1/2}$ gives $\mathcal L'(t)=\lambda(1-2\lambda t)^{-3/2}\ge0$ and $\mathcal L''(t)=3\lambda^2(1-2\lambda t)^{-5/2}$. Whenever $\sigma_x^2\le \tau$, the interval between $1$ and $\sigma_x^2$ lies inside $[0,\tau]$, where $1-2\lambda t\ge\frac12$ because $4\lambda\tau\le1$. Thus $\mathcal L''(t)\le3\cdot2^{5/2}\lambda^2\le17\lambda^2$ throughout the interval, and Taylor's theorem with Lagrange remainder around $t=1$ gives
\[
\mathcal L(\sigma_x^2)\le\mathcal L(1)+\mathcal L'(1)\,(\sigma_x^2-1)+8.5\,\lambda^2(\sigma_x^2-1)^2\qquad\text{whenever }\sigma_x^2\le \tau .
\]
Multiplying by $\mathbf 1\{\sigma_x^2\le \tau\}$ and taking expectations, the constant term is bounded by $\mathcal L(1)$, the linear term is nonpositive because $\mathcal L'(1)\ge0$ and $\E[(\sigma_x^2-1)\mathbf 1\{\sigma_x^2\le \tau\}]=-\E[(\sigma_x^2-1)\mathbf 1\{\sigma_x^2>\tau\}]\le0$ (since $\E\sigma_x^2=1$ and $\tau\ge1$), and the quadratic term is bounded by $8.5\,\lambda^2\Var(\sigma_x^2)\le 8.5\,\lambda^2V$. Using $\mathcal L(1)\ge1$ and $1+y\le e^y$, we obtain
\[
\E[\mathcal L(\sigma_x^2)\mathbf 1\{\sigma_x^2\le \tau\}]\le\mathcal L(1)+8.5\,\lambda^2V\le\mathcal L(1)\,e^{8.5\lambda^2V}.
\]
Multiplying by $e^{-\lambda}$ and taking logarithms bounds the left-hand side of \eqref{eq:gg-mgf-upper} by $-\lambda-\frac12\log(1-2\lambda)+8.5\,\lambda^2V$. Finally, since $2\lambda\le2\lambda\tau\le\frac12$, expanding the logarithm in a power series yields $-\lambda-\frac12\log(1-2\lambda)=\sum_{p\ge2}(2\lambda)^p/(2p)\le\frac14\sum_{p\ge2}(2\lambda)^p=\lambda^2/(1-2\lambda)$.
\end{proof}

\begin{proof}[Proof of Theorem~\ref{thm:gg-param}]
(a) As described in Section~\ref{sec:gg}, the sparsity requirement on $r$ is the maximum of a variance condition and a capture condition. For each $x\in\Ufam$, let $\sigma_{x,i}^2=\|\xt_{\capset_i}\|^2$ denote the energy captured from $x$ by row $i$, let $T_i:=(\Phi x)_i^2$, and let $\bar T:=\ell^{-1}\|\Phi x\|^2$. By \eqref{eq:gg-row}, the pairs $(\sigma_{x,i}^2,T_i)$ for $i=1,\dots,\ell$ are i.i.d.\ with the distribution of $(\sigma_x^2,\sigma_x^2\,\xi^2)$.

\emph{Variance control.} For every $x\in\Ufam$, the variance condition $r\ge8f_qk_\Ufam$ and \eqref{eq:gg-row} ensure $\Var(\sigma_x^2)=f_q(k_x-\bal)/r\le f_qk_\Ufam/r\le\frac18$.

\emph{Capture event.} Set $\tau:=1+1/\eps$, and let $\mathcal C$ denote the event that $\sigma_{x,i}^2\le \tau$ holds simultaneously for all $i\in[\ell]$ and all $x\in\Ufam$. If the first branch of the capture condition holds ($r\ge\eps n/(1+\eps)$), then $\sigma_{x,i}^2\le n/r\le \tau$ deterministically (Lemma~\ref{lem:sampling}(i)), so $\Pr[\mathcal C^c]=0$. Otherwise, the second branch $r\ge w_\Ufam\log(3\ell|\Ufam|/\delta)/h(1/\eps)$ holds; applying Bennett's upper-tail bound (Lemma~\ref{lem:sampling}(iii) with $t=1/\eps$) and $w_x\le w_\Ufam$ gives
\[
\Pr[\sigma_{x,i}^2>\tau]\le\exp\{-(r/w_\Ufam)\,h(1/\eps)\}\le\frac{\delta}{3\ell|\Ufam|}
\]
for each of the $\ell|\Ufam|$ pairs $(i,x)$, so a union bound yields $\Pr[\mathcal C^c]\le\delta/3$. (Note that truncating at $\tau=1+1/\eps$, where $h(1/\eps)=\Theta(\eps^{-1}\log(1+1/\eps))$, rather than at a constant threshold $\tau=2$ is precisely where Theorem~\ref{thm:fhl23} saves a logarithmic factor over Theorem~\ref{thm:ac09}.)

\emph{Upper tail.} Fix $x\in\Ufam$ and set $\lambda:=\eps/8$. Since $4\lambda\tau=(1+\eps)/2\le1$, Lemma~\ref{lem:gg-row-mgf} applies with $V=\frac18$ (giving $8.5\,\lambda^2V=\frac{17}{16}\lambda^2$). Because $\mathbf 1\{\sigma_{x,i}^2\le \tau\}=1$ for all $i$ on $\mathcal C$, applying Markov's inequality to $\prod_{i=1}^\ell e^{\lambda(T_i-1)}\mathbf 1\{\sigma_{x,i}^2\le \tau\}$ and using row independence and \eqref{eq:gg-mgf-upper} yields
\[
\Pr\big[\{\bar T\ge1+\eps\}\cap\mathcal C\big]
\le e^{-\ell\lambda\eps}\big(\E[e^{\lambda(T-1)}\mathbf 1\{\sigma_x^2\le \tau\}]\big)^\ell
\le\exp\Big\{-\ell\Big(\lambda\eps-\frac{\lambda^2}{1-2\lambda}-\frac{17}{16}\lambda^2\Big)\Big\}.
\]
Since $\lambda\le\frac18$, we have $(1-2\lambda)^{-1}\le\frac43$ and $\frac43+\frac{17}{16}<2.4$, so the quantity in parentheses is at least $\lambda\eps-2.4\,\lambda^2=(\frac18-\frac{2.4}{64})\,\eps^2=0.0875\,\eps^2\ge\eps^2/12$.

\emph{Lower tail.} Because the variables $T_i\ge 0$ are independent with $\E T_i=1$ and $\E T_i^2=\E\sigma_x^4\,\E\xi^4=3(1+\Var(\sigma_x^2))\le\frac{27}8$, Maurer's inequality \eqref{eq:app-maurer} applied to $\ell\bar T$ at deviation $\ell\eps$ gives $\Pr[\bar T\le1-\eps]\le\exp\{-(\ell\eps)^2/(2\ell\cdot\frac{27}8)\}=e^{-4\ell\eps^2/27}\le e^{-\ell\eps^2/12}$.

\emph{Union bound.} The sketch-length condition $\ell\ge12\,\eps^{-2}\log(3|\Ufam|/\delta)$ ensures $e^{-\ell\eps^2/12}\le\delta/(3|\Ufam|)$. Summing over the $2|\Ufam|$ upper- and lower-tail events (contributing at most $2\delta/3$) and adding $\Pr[\mathcal C^c]\le\delta/3$ bounds the total failure probability by $\delta$.

(b) By Proposition~\ref{prop:gg-moments}, the row products $X_iY_i$ of each pair are i.i.d.\ with mean $\rho$ and variance at most $\sigma_*^2$, so the claim follows immediately from the deviation form of Lemma~\ref{lem:catoni}. The final sentence follows from $k_{uv}\le w_u$ \eqref{eq:params}, $\rho^2\le1$, and $f_q\le1$.
\end{proof}

\begin{proof}[Proof of Corollary~\ref{cor:gg-classical}]
(i) Abbreviate $\mu_1:=\mu(2|\Ufam|)=2\log(8n|\Ufam|/\delta)$.

\emph{Flattening.} Applying Lemma~\ref{lem:flat} with $\delta_0=\delta/4$ guarantees $k_\Ufam\le w_\Ufam\le2\log(8n|\Ufam|/\delta)=\mu_1$ with probability at least $1-\delta/4$ over $D$. Fix any $D$ in this event.

\emph{Conditional bound.} We invoke Theorem~\ref{thm:gg-param}(a) with failure budget $3\delta/4$ (so that the overall failure probability is at most $\delta/4+3\delta/4=\delta$). Its logarithmic factors become $\log(4|\Ufam|/\delta)$ and $\log(4\ell|\Ufam|/\delta)$, matching our condition on $\ell$. Its variance condition $r\ge8f_qk_\Ufam$ holds because $r\ge\min\{n,8\mu_1\}$ implies either $r=n$ (where $f_q=0$) or $r\ge8\mu_1\ge8f_qk_\Ufam$. Its capture condition holds via its first branch if $r\ge\eps n/(1+\eps)$, and otherwise follows from $w_\Ufam\le\mu_1$ and the elementary bound $h(t)\ge\frac t2\log(1+t)$ at $t=1/\eps$:
\[
r\ge\frac{2\eps\mu_1\log(4\ell|\Ufam|/\delta)}{\log(1+1/\eps)}
\ge\frac{w_\Ufam\log(4\ell|\Ufam|/\delta)}{h(1/\eps)}.
\]
(To verify $h(t)\ge\frac t2\log(1+t)$ for $t\ge0$, note that both sides vanish at $t=0$ and their difference has derivative $\frac12\log(1+t)-\frac t{2(1+t)}\ge0$.)

\emph{Distances and inner products.} To preserve pairwise distances among $m$ points, take $\Ufam$ to be the set of normalized difference vectors $(x-y)/\|x-y\|$ for all distinct unordered pairs ($|\Ufam|<m^2/2$). To preserve inner products of $N$ unit-vector pairs, take $\Ufam$ to be the normalized nonzero vectors among $u\pm v$ ($|\Ufam|\le2N$); polarization $4(\hat\rho_{\rm GG}-\rho)=(\ell^{-1}\|\Phi(u+v)\|^2-\|u+v\|^2)-(\ell^{-1}\|\Phi(u-v)\|^2-\|u-v\|^2)$ together with $\|u+v\|^2+\|u-v\|^2=4$ then gives $|\hat\rho_{\rm GG}-\rho|\le\eps$.

(ii) \emph{Flattening and Catoni's decoder.} Applying Lemma~\ref{lem:flat} with $\delta_0=\delta/2$ to the $m$ fixed database vectors ensures $w_u\le2\log(4nm/\delta)=\mu(m)$ simultaneously for all stored vectors with probability at least $1-\delta/2$ over $D$ (since the bound uses only $w_u$, the queries $v$ may depend on $D$). On this event, Proposition~\ref{prop:gg-moments} and $k_{uv}\le w_u$ bound the row-product variance of every pair by $2+3f_qw_u/r\le3$ (as either $r=n$ and $f_q=0$, or $r\ge3\,\mu(m)\ge3w_u$). Applying Lemma~\ref{lem:catoni} with $\sigma_*^2=3$, accuracy $\eps$, scale $\alpha=\eps/(3+\eps^2)$, and failure probability $\delta/2$ requires $\ell\ge2(3+\eps^2)\,\eps^{-2}\log(4N/\delta)$, which is satisfied since $2(3+\eps^2)\le8$.

\emph{Encoding complexity.} Set $L_\Ufam:=\log(4|\Ufam|/\delta)$. Because rounding $\ell$ up to an integer adds at most $1<\eps^{-2}\log4$, the minimum allowed sketch length satisfies $\ell\le13\,\eps^{-2}L_\Ufam$ in~(i) and $\ell\le9\,\eps^{-2}\log(4N/\delta)$ in~(ii), while in~(ii) the minimum sparsity satisfies $r\le 1+3\,\mu(m)$. In~(i), evaluating the second capture branch at $\ell\le13\,\eps^{-2}L_\Ufam$ gives
\begin{align*}
\log\frac{4\ell|\Ufam|}\delta=L_\Ufam+\log\ell
&\le L_\Ufam+\log13+2\log\frac1\eps+\log L_\Ufam
\le2\mu_1+2\log\frac1\eps ,\\
\frac{2\eps\mu_1\log(4\ell|\Ufam|/\delta)}{\log(1+1/\eps)}
&\le\frac{4\eps\mu_1\log(1/\eps)}{\log(1+1/\eps)}+\frac{4\eps\mu_1^2}{\log(1+1/\eps)}
\le4\eps\mu_1+\frac{8\eps\mu_1^2}{\log(2/\eps)} ,
\end{align*}
where the first line uses $\log L_\Ufam\le L_\Ufam$, $2L_\Ufam\le\mu_1$, and $\log13<\mu_1$, and the second line uses $\log(1/\eps)\le\log(1+1/\eps)$ and $\log(1+1/\eps)\ge\frac12\log(2/\eps)$ (since $(1+1/\eps)^2\ge2/\eps$ for $\eps\le1$). Thus the smallest integer $r$ in~(i) is at most $1+\max\{8\mu_1,\,4\eps\mu_1+8\eps\mu_1^2/\log(2/\eps)\}\le 1+12\mu_1\max\{1,\eps\mu_1/\log(2/\eps)\}$, and multiplying by $\ell$ yields the stated operation counts.
\end{proof}

\paragraph{Scale-free robust estimation of squared distances.}
While Corollary~\ref{cor:gg-classical}(ii) estimates inner products of unit vectors at $\eps$-independent sparsity $r=3\,\mu(m)=O(\log(nm/\delta))$, extending Catoni's estimator to unnormalized pairwise squared distances $\|d\|^2=\|x-y\|^2$ faces a scaling obstacle: the squared projections $(\Phi d)_i^2$ have mean $\|d\|^2$ and standard deviation $\Theta(\|d\|^2)$, so the optimal scale parameter $\alpha$ in \eqref{eq:catoni-def} would need to scale as $\eps/\|d\|^2$, which depends on the very distance we wish to estimate. We overcome this obstacle by normalizing each squared projection by the candidate value $c>0$ inside the influence function. Specifically, for $\alpha>0$ and any $d\in\R^n$ with $\Phi d\ne0$, consider the root-finding equation
\begin{equation}\label{eq:gg-dist-root}
\sum_{i=1}^\ell\psi\Big(\alpha\Big\{\frac{(\Phi d)_i^2}{c}-1\Big\}\Big)=0
\end{equation}
in $c>0$, where $\psi$ is Catoni's influence function \eqref{eq:catoni-def}. Because the left-hand side is continuous and strictly decreasing in $c$---ranging from $+\infty$ as $c\downarrow 0$ to $\ell\psi(-\alpha)<0$ as $c\to\infty$---equation \eqref{eq:gg-dist-root} admits a unique positive root $\widehat{\|d\|^2}$ (readily found by bisection on $\log c$; we set $\widehat{\|d\|^2}:=0$ if $\Phi d=0$). We call $\widehat{\|d\|^2}$ the \emph{self-normalized Catoni estimate} of $\|d\|^2$. Notice that for $d=x-y$, the estimator depends only on the linear sketches $\Phi x$ and $\Phi y$, and replacing $d$ by $td$ scales $\widehat{\|d\|^2}$ by $t^2$, allowing a single universal parameter $\alpha$ to work simultaneously across all distance scales.

\begin{theorem}[GG squared distances]\label{thm:gg-dist}
Use either support law, and let $\delta\in(0,1)$ and $\eps\in(0,1)$.
\begin{enumerate}
\item[(a)] \emph{Realized preconditioner.} Fix any $D$. Let $\mathcal B$ be an admissible finite set of nonzero vectors, set $k_{\mathcal B}:=\max_{d\in\mathcal B}k_{d/\|d\|}$, and let $\sigma_*^2\ge2+3f_q\,k_{\mathcal B}/r$. If
\[
\alpha=\frac{\eps(1-\eps^2)}{\sigma_*^2+\eps^2}\qquad\text{and}\qquad
\ell\ge\frac{2(\sigma_*^2+\eps^2)}{\eps^2(1-\eps^2)}\log\frac{2|\mathcal B|}\delta,
\]
then with probability at least $1-\delta$ over $P$, the self-normalized Catoni estimate of \eqref{eq:gg-dist-root} satisfies $\big|\widehat{\|d\|^2}-\|d\|^2\big|\le\eps\|d\|^2$ for all $d\in\mathcal B$.
\item[(b)] \emph{Classical form.} Let $m$ points be fixed independently of $(D,P)$, let $\eps\le\frac12$, and recall $\mu(m^2/2)=2\log(2nm^2/\delta)$ from \eqref{eq:mu-def}. If
\[
r\ge\min\{n,3\,\mu(m^2/2)\},\qquad
\ell\ge\frac9{\eps^2}\log\frac{2m^2}\delta\qquad\text{and}\qquad
\alpha=\frac{\eps(1-\eps^2)}{3+\eps^2},
\]
then with probability at least $1-\delta$, we have $\big|\widehat{\|x-y\|^2}-\|x-y\|^2\big|\le\eps\|x-y\|^2$ for all pairs of points $x,y$.
\end{enumerate}
\end{theorem}

Comparing part~(a) with Lemma~\ref{lem:catoni} shows that self-normalization increases the required sketch length only by the factor $(1-\eps^2)^{-1}=1+O(\eps^2)$, while the row sparsity $r$ enters solely through the variance bound $\sigma_*^2$. (In part~(b), the restriction $\eps\le\frac12$ is used only to simplify $2(3+\eps^2)/(1-\eps^2)\le\frac{26}{3}<9$; for $\frac12<\eps<1$, the same sparsity $r\ge\min\{n,3\,\mu(m^2/2)\}$ suffices with $\ell\ge2(3+\eps^2)\,\eps^{-2}(1-\eps^2)^{-1}\log(2m^2/\delta)$.) In particular, for dense rows ($f_q=0$, $\sigma_*^2=2$), part~(a) requires $\ell\ge(4+O(\eps^2))\,\eps^{-2}\log(2|\mathcal B|/\delta)$, matching the optimal leading constant $4$ of Theorem~\ref{thm:dense-gg}.

\begin{proof}[Proof of Theorem~\ref{thm:gg-dist}]
(a) Fix $d\in\mathcal B$, and set $T_i:=(\Phi d)_i^2/\|d\|^2$ and $\hat\tau:=\widehat{\|d\|^2}/\|d\|^2$, so that our goal is to prove $|\hat\tau-1|\le\eps$. By \eqref{eq:gg-row} and Proposition~\ref{prop:gg-moments} applied to the unit vector $x:=d/\|d\|$, the variables $T_i$ are i.i.d.\ with mean $\E T_i=1$ and variance $\Var T_i=2+3\Var(\sigma_x^2)\le2+3f_q\,k_{\mathcal B}/r\le\sigma_*^2$. Setting $c=\tau\|d\|^2$ rewrites the left-hand side of \eqref{eq:gg-dist-root} as $\Psi(\tau):=\sum_{i=1}^\ell\psi\big((\alpha/\tau)(T_i-\tau)\big)$ for $\tau>0$. Because $\Psi$ is strictly decreasing with unique root $\hat\tau$ whenever $\Phi d\ne0$ (and $\hat\tau=0$, $\Psi(\tau)=\ell\psi(-\alpha)<0$ when $\Phi d=0$), we have the exact equivalence of events
\[
\{\hat\tau>1+\eps\}=\{\Psi(1+\eps)>0\}\qquad\text{and}\qquad
\{\hat\tau<1-\eps\}=\{\Psi(1-\eps)<0\}.
\]

Notice that at $\tau=1\pm\eps$, the sum $\Psi(1\pm\eps)$ is identical to the standard Catoni sum in \eqref{eq:catoni-def} evaluated on the unit-mean samples $T_i$ at the shifted scale parameter $\alpha_\pm:=\alpha/(1\pm\eps)=\eps(1\mp\eps)/(\sigma_*^2+\eps^2)$. Applying \eqref{eq:catoni-mgf} to $W=T_i$ (with $c_0=1$ and variance at most $\sigma_*^2$) using the $+$ sign and scale $\alpha_+$ at $c=1+\eps$ (so $c_0-c=-\eps$), Markov's inequality and row independence yield
\[
\Pr[\hat\tau>1+\eps]
\le\E e^{\Psi(1+\eps)}=\prod_{i=1}^\ell\E e^{\psi(\alpha_+(T_i-1-\eps))}
\le\exp\Big\{-\ell\Big(\alpha_+\eps-\frac{\alpha_+^2}2(\sigma_*^2+\eps^2)\Big)\Big\}.
\]
Likewise, applying \eqref{eq:catoni-mgf} with the $-$ sign and scale $\alpha_-$ at $c=1-\eps$ bounds $\Pr[\hat\tau<1-\eps]\le\E e^{-\Psi(1-\eps)}$ by the same expression with $\alpha_-$ in place of $\alpha_+$. Remarkably, the two exponents are identical:
\[
\alpha_\pm\eps-\frac{\alpha_\pm^2}2(\sigma_*^2+\eps^2)
=\frac{\eps^2}{\sigma_*^2+\eps^2}\,(1\mp\eps)\Big(1-\frac{1\mp\eps}2\Big)
=\frac{\eps^2(1-\eps^2)}{2(\sigma_*^2+\eps^2)} .
\]
The condition on $\ell$ ensures that each tail has probability at most $\delta/(2|\mathcal B|)$, and a union bound over all $2|\mathcal B|$ tails proves~(a).

(b) Let $\mathcal B$ be the set of nonzero difference vectors $x-y$ across all unordered pairs ($|\mathcal B|\le m^2/2$; pairs with $x=y$ have $\widehat{\|x-y\|^2}=0=\|x-y\|^2$ deterministically). Applying Lemma~\ref{lem:flat} with $\delta_0=\delta/2$ to the normalized vectors in $\mathcal B$ ensures $k_{\mathcal B}\le\max_{d\in\mathcal B}w_{d/\|d\|}\le2\log(4n|\mathcal B|/\delta)\le\mu(m^2/2)$ with probability at least $1-\delta/2$ over $D$. On this event, $2+3f_qk_{\mathcal B}/r\le3$, so part~(a) applies with $\sigma_*^2=3$ and failure probability $\delta/2$, requiring $\ell\ge2(3+\eps^2)\,\eps^{-2}(1-\eps^2)^{-1}\log(4|\mathcal B|/\delta)\le 9\,\eps^{-2}\log(2m^2/\delta)$ for $\eps\le\frac12$.
\end{proof}

\subsection{One bit against full precision (BG)}\label{app:bg}

This section proves our guarantees for the asymmetric estimator BG from Section~\ref{sec:bg}, along with Theorem~\ref{thm:dense-bg} for dense projections. Throughout this section (until the proof of Corollary~\ref{cor:bg-classical}), we condition on a fixed sign matrix $D$ and a pair of unit vectors $(u,v)$, let $\capset$ denote the support of a single row, and write $\E$, $\Var$, and $\Pr$ for expectations, variances, and probabilities over the draw of $P$. Writing $\bar\zeta:=\E[Z\mid D]=\E\zeta_\capset$ for the expected single-row score, the estimation error decomposes into sampling noise plus support bias: $\hat\rho_{\rm BG}-\rho=(\hat\rho_{\rm BG}-\bar\zeta)+b_{uv}$, where $b_{uv}=\E\zeta_\capset-\rho$. Lemma~\ref{lem:bg-cond} characterizes the conditional distribution and moment generating function of a single row score $Z$ given its support $\capset$. To bound the support bias $b_{uv}=\E[\sigma_{uv}/\sigma_u]-\rho$, we derive an exact algebraic identity expressing $\zeta_\capset-\rho$ in terms of products of coordinate sampling fluctuations, which we bound via Cauchy--Schwarz on typical supports ($\sigma_u^2\ge t$) and pointwise on the lower tail ($\sigma_u^2<t$) in Proposition~\ref{prop:bg-bias}. To control the sampling noise across rows, Catoni's decoder requires only the unconditional variance bound $\Var(Z\mid D)\le\pi/2$, whereas the dot-product decoder requires bounding the unconditional moment generating function $\E[e^{\lambda(Z-\bar\zeta)}\mid D]$ over the random draw of $\capset$ (Proposition~\ref{prop:bg-mgf}).

Throughout the analysis, we set $\eta:=\sigma_u^2-1$ and $\omega:=\sigma_{uv}-\rho\sigma_u^2=\sqrt{1-\rho^2}\,\sigma_{12}$, and use the shorthand $\E[W;\mathcal A]:=\E[W\,\mathbf 1_{\mathcal A}]$ for any event $\mathcal A$. For any vector $y\in\R^n$, we refer to $T=q^{-1}\sum_{j\in\capset}y_j$ in Lemma~\ref{lem:sampling} as the \emph{captured sum} of $y$, whose expectation is $\sum_jy_j$. Because $\ut_j=e_1^\top\cvec_j$ and $\vt_j=t_\theta^\top\cvec_j$, the definition of $M_\capset$ in \eqref{eq:random-metric} implies that
\begin{equation}
\sigma_u^2,\ \ \sigma_{uv},\ \ \sigma_v^2,\ \ \omega\quad\text{are the captured sums of}\quad
\ut\circ\ut,\ \ \ut\circ\vt,\ \ \vt\circ\vt,\ \ \ut\circ(\vt-\rho\ut),
\label{eq:bg-captured}
\end{equation}
with coordinate totals $1$, $\rho$, $1$, and $0$, respectively. In particular, $\E\omega=0$, and Lemma~\ref{lem:sampling} applies directly to the captured energies $\sigma_u^2$ (with $x=u$) and $\sigma_v^2$ (with $x=v$).

\begin{lemma}[Conditional law of the BG score]\label{lem:bg-cond}
Fix $D$ and the support $\capset$ of a single row, and adopt the convention $\signzero$. Then $|\zeta_\capset|\le\sigma_v$, $\E[Z\mid\capset]=\zeta_\capset$, and every $\lambda\in\R$ satisfies
\begin{equation}
\E\big[e^{\lambda Z}\mid\capset\big]=e^{\pi\lambda^2\sigma_v^2/4}\,2\Phi_0\big(\sqrt{\pi/2}\,\lambda\zeta_\capset\big)\le e^{\lambda\zeta_\capset+\pi\lambda^2\sigma_v^2/4},
\label{eq:bg-exact-cgf}
\end{equation}
where $\Phi_0$ denotes the standard normal cumulative distribution function. Furthermore, $\E[Z^2\mid D]=\pi/2$ and $\Var(Z\mid D)=\pi/2-(\E[Z\mid D])^2$ for every realization of $D$.
\end{lemma}

\begin{proof}[Proof of Lemma~\ref{lem:bg-cond}]
Let $\xi_1,\xi_2$ be independent standard normal random variables. By the Gram identity (Lemma~\ref{lem:random-metric}(iv)), $|\zeta_\capset|\le\sigma_v$, and $\sigma_{uv}=0$ whenever $\sigma_u=0$. Conditioned on $\capset$, the projection pair $(X,Y)$ has the same joint distribution as $\big(\sigma_u\,\xi_1,\ \zeta_\capset\,\xi_1+(\sigma_v^2-\zeta_\capset^2)^{1/2}\xi_2\big)$, since both pairs are centered bivariate normal with variances $\sigma_u^2$ and $\sigma_v^2$ and covariance $\sigma_u\,\zeta_\capset=\sigma_{uv}$ (Lemma~\ref{lem:random-metric}(i)). Noting that $\sign(\sigma_u\,\xi_1)\,\xi_1=|\xi_1|$ when $\sigma_u>0$ and $\zeta_\capset=0$ when $\sigma_u=0$, we obtain in both cases the distributional representation
\[
Z\ \overset{d}{=}\ \sqrt{\pi/2}\,\Big(\zeta_\capset\,|\xi_1|+\big(\sigma_v^2-\zeta_\capset^2\big)^{1/2}\xi_2'\Big),
\qquad \xi_2':=\sign(\sigma_u\,\xi_1)\,\xi_2 .
\]
Because $\sign(\sigma_u\,\xi_1)\in\{\pm1\}$ (by $\signzero$) depends only on $\xi_1$, while $\xi_2$ is symmetric and independent of $\xi_1$, the variable $\xi_2'$ is standard normal and independent of $\xi_1$.

Taking expectations using $\E|\xi_1|=\sqrt{2/\pi}$ yields $\E[Z\mid\capset]=\zeta_\capset$. Similarly, using the independence of $|\xi_1|$ and $\xi_2'$ together with the half-normal moment generating function $\E e^{t|\xi_1|}=2e^{t^2/2}\Phi_0(t)$ gives
\[
\E\big[e^{\lambda Z}\mid\capset\big]
=\E e^{\sqrt{\pi/2}\,\lambda\zeta_\capset|\xi_1|}\;\E e^{\sqrt{\pi/2}\,\lambda(\sigma_v^2-\zeta_\capset^2)^{1/2}\xi_2'}
=2e^{\pi\lambda^2\zeta_\capset^2/4}\Phi_0\big(\sqrt{\pi/2}\,\lambda\zeta_\capset\big)\,e^{\pi\lambda^2(\sigma_v^2-\zeta_\capset^2)/4},
\]
establishing the identity in \eqref{eq:bg-exact-cgf}. To deduce the upper bound, observe that $\log(2\Phi_0(t))$ is concave (by log-concavity of the Gaussian density), vanishes at $t=0$, and has derivative $\Phi_0'(0)/\Phi_0(0)=\sqrt{2/\pi}$ at $t=0$; it therefore lies below its tangent line, giving $2\Phi_0(t)\le e^{t\sqrt{2/\pi}}$ for all $t\in\R$. Substituting $t=\sqrt{\pi/2}\,\lambda\zeta_\capset$ yields the inequality in \eqref{eq:bg-exact-cgf}.

Finally, because $\sign(X)^2=1$ everywhere, we have $Z^2=\frac\pi2Y^2$ identically, so Lemma~\ref{lem:random-metric}(i) and~(ii) give $\E[Z^2\mid D]=\frac\pi2\,\E\sigma_v^2=\frac\pi2$, from which the variance formula follows.
\end{proof}

\begin{proof}[Proof of Theorem~\ref{thm:dense-bg}]
A dense Gaussian row $\Phi=G$ corresponds to the deterministic metric $M_\capset=I_2$ ($\sigma_u^2=\sigma_v^2=1$ and $\sigma_{uv}=\zeta_\capset=\rho$) in Lemma~\ref{lem:bg-cond} for any ambient dimension $n$. Lemma~\ref{lem:bg-cond} therefore yields $\E Z_i=\rho$, $\E Z_i^2=\pi/2$, $\Var Z_i=\pi/2-\rho^2$, and $\E e^{\lambda(Z_i-\rho)}\le e^{\pi\lambda^2/4}$ for all $\lambda\in\R$. For any $t>0$, row independence and Markov's inequality at $\lambda=2t/\pi$ give $\Pr[\pm(\hat\rho_{\rm BG}-\rho)\ge t]\le e^{-\ell(\lambda t-\pi\lambda^2/4)}=e^{-\ell t^2/\pi}$. Taking a union bound over all $2N$ one-sided tail events bounds the failure probability by $2Ne^{-\ell t^2/\pi}$, which equals $\delta$ at $t=\sqrt{\pi\log(2N/\delta)/\ell}$.
\end{proof}

\paragraph{Bounding the support bias.}
Next, we establish the coordinatewise pair certificate for the support bias $b_{uv}$, from which Proposition~\ref{prop:bg-support} will follow by specialization.

\begin{proposition}[BG support bias, certificate form]\label{prop:bg-bias}
Fix $D$ and a pair $(u,v)$ under either support law. Set $V_u:=\Var(\sigma_u^2)=f_q(k_u-\bal)/r$, $V_{uv}:=\Var(\sigma_{uv}-\rho\sigma_u^2)=\frac{f_q}{q}\|\ut\circ(\vt-\rho\ut)\|^2$, and $p_t:=\Pr[\sigma_u^2<t]$. Then for every threshold $t\in(0,1)$,
\begin{equation}
|b_{uv}|\le\frac{|\rho|\,V_u}{2(1+\sqrt t)^2}+\frac{\sqrt{V_uV_{uv}}}{\sqrt t\,(1+\sqrt t)}+\frac{|\rho|}2\,p_t+\sqrt{(1-\rho^2)\,p_t} .
\label{eq:bg-bias-coarse}
\end{equation}
Here $p_t\le e^{-(1-t)^2r/(2k_u)}$, $V_u\le k_u/r\le w_u/r$, and $V_{uv}\le\min\{(1-\rho^2)\,w_u,\ (1+|\rho|)^2f_q\kmax\}/r$.
\end{proposition}

In \eqref{eq:bg-bias-coarse}, the first two terms bound the bias on the high-probability event $\{\sigma_u^2\ge t\}$, while the last two terms control the lower-tail event $\{\sigma_u^2<t\}$ where the denominator $\sigma_u$ of $\zeta_\capset$ becomes small. Notice that as $t\uparrow 1$, the prefactors of the first two terms converge to $\frac18$ and $\frac12$, matching the exact second-order Taylor expansion of Remark~\ref{rem:bg-estimate}. Replacing $p_t$ with its Maurer bound $e^{-(1-t)^2r/(2k_u)}$ allows \eqref{eq:bg-bias-coarse} to be evaluated in $O(n)$ time from $\ut$ and $\vt$, and our numerical experiments in Section~\ref{sec:experiments} optimize this bound over $t\in(0,1)$.

\begin{proof}[Proof of Proposition~\ref{prop:bg-bias}]
\emph{Step 1: Exact bias decomposition.} Writing $\sigma_{uv}=\rho\sigma_u^2+\omega$ gives $\zeta_\capset-\rho=\rho\,(\sigma_u-1)+\omega/\sigma_u$ on every support. Because $M_\capset\succeq0$, its determinant is nonnegative, so $\sigma_{12}^2\le\sigma_1^2\sigma_2^2=\sigma_u^2\sigma_2^2$ and hence $\omega^2\le(1-\rho^2)\,\sigma_u^2\sigma_2^2$ (in particular, $\omega=0$ whenever $\sigma_u=0$, consistent with setting $\omega/\sigma_u:=0$ and $\zeta_\capset:=0$ there). Observing that $(\sigma_u-1)^2=(\sigma_u^2-1)-2(\sigma_u-1)$ and $\E\sigma_u^2=1$ gives $\E(\sigma_u-1)=-\frac12\E(\sigma_u-1)^2$, while $\E\omega=0$ gives $\E[\omega/\sigma_u]=\E[\omega\,(\sigma_u^{-1}-1)]$. Taking expectations of $\zeta_\capset-\rho$ therefore yields the exact identity
\begin{equation}
b_{uv}=-\frac\rho2\,\E\big(\sigma_u-1\big)^2-\E\Big[\omega\,\frac{\sigma_u-1}{\sigma_u}\Big].
\label{eq:bg-bias-identity}
\end{equation}

\emph{Step 2: Truncation at $\{\sigma_u^2\ge t\}$.} Abbreviate $c_1(t):=1/(2(1+\sqrt t)^2)$ and $c_2(t):=1/(\sqrt t\,(1+\sqrt t))$, and recall that $\E\eta^2=V_u$ and $\E\omega^2=V_{uv}$. On the event $\{\sigma_u^2\ge t\}$, we have $\sigma_u\ge\sqrt t$ and $|\sigma_u-1|=|\eta|/(1+\sigma_u)\le|\eta|/(1+\sqrt t)$. On the complementary event $\{\sigma_u^2<t\}$, we have $(\sigma_u-1)^2\le1$ (since $0\le\sigma_u<1$) and, by Step~1, $|\omega|\,|\sigma_u-1|/\sigma_u\le|\omega|/\sigma_u\le\sqrt{(1-\rho^2)\,\sigma_2^2}$. Applying the Cauchy--Schwarz inequality in the second and fourth lines below (together with $\E\sigma_2^2=1$) yields
\[
\begin{aligned}
\E\big[(\sigma_u-1)^2;\,\sigma_u^2\ge t\big]&\le\E\eta^2/(1+\sqrt t)^2=2c_1(t)\,V_u,\\
\E\big[|\omega|\,|\sigma_u-1|/\sigma_u;\,\sigma_u^2\ge t\big]&\le\E|\omega\eta|/\big(\sqrt t\,(1+\sqrt t)\big)\le c_2(t)\sqrt{V_uV_{uv}},\\
\E\big[(\sigma_u-1)^2;\,\sigma_u^2<t\big]&\le p_t,\\
\E\big[|\omega|\,|\sigma_u-1|/\sigma_u;\,\sigma_u^2<t\big]&\le\sqrt{1-\rho^2}\;\E\big[\sigma_2;\,\sigma_u^2<t\big]\le\sqrt{(1-\rho^2)\,p_t}.
\end{aligned}
\]
Combining these four inequalities with \eqref{eq:bg-bias-identity} establishes \eqref{eq:bg-bias-coarse}.

\emph{Step 3: Parameter bounds.} Lemma~\ref{lem:sampling}(i) and~(ii) give the expressions for $V_u$ and the upper bound on $p_t$. By \eqref{eq:bg-captured}, $\omega$ is the captured sum of $y:=\ut\circ(\vt-\rho\ut)$, whose coordinates sum to $0$, so Lemma~\ref{lem:sampling} gives $V_{uv}=\frac{f_q}q\|y\|^2=\frac{f_qn}r\|y\|^2$. The two upper bounds on $V_{uv}$ follow immediately from
\[
\begin{aligned}
\|y\|^2&\le\|\ut\|_\infty^2\,\|\vt-\rho\ut\|^2=\frac{w_u}n\,(1-\rho^2),\\
\|y\|&\le\|\ut\circ\vt\|+|\rho|\,\|\ut\circ\ut\|=\sqrt{k_{uv}/n}+|\rho|\sqrt{k_u/n}\le(1+|\rho|)\sqrt{\kmax/n},
\end{aligned}
\]
using $f_q\le1$ in the first line and $k_{uv}\le\sqrt{k_uk_v}\le\kmax$ \eqref{eq:params} in the second.
\end{proof}

\begin{remark}[Leading-order bias approximation]\label{rem:bg-estimate}
Replacing $\sigma_u-1$ by its linear Taylor approximation $\eta/2=(\sigma_u^2-1)/2$ and $\sigma_u$ in the denominator by $1$ in the exact identity \eqref{eq:bg-bias-identity} yields the second-order approximation
\[
b_{uv}\approx-\rho V_u/8-\Cov(\omega,\sigma_u^2)/2,\qquad\text{where}\quad\Cov(\omega,\sigma_u^2)=\frac{f_q}q\sum_j\ut_j^3(\vt_j-\rho\ut_j)
\]
by Lemma~\ref{lem:sampling}. Here $-\rho V_u/8$ is $\rho$ times the self-pair bias of the stored vector $u$, while $-\Cov(\omega,\sigma_u^2)/2$ captures the cross-correlation between energy fluctuations and angular distortion. Evaluated in $O(n)$ time, this approximation agrees with the true bias to within $1.3\%$ on Gaussian-like pairs at $r\in\{64,256\}$, and to within $14\%$ on cap-saturating pairs with $k_u=w_u\le0.2\,r$ (see the supplementary code documentation; Appendix~\ref{app:experiments}). Because it is an asymptotic approximation rather than a rigorous upper bound, however, it is never used in our theorems.
\end{remark}

\begin{proof}[Proof of Proposition~\ref{prop:bg-support}]
Each bound in Proposition~\ref{prop:bg-support} is obtained by evaluating \eqref{eq:bg-bias-coarse} at a specific threshold $t\in(0,1)$ and substituting the parameter bounds for $V_u$, $V_{uv}$, and $p_t$ from Proposition~\ref{prop:bg-bias}. Throughout the proof, we repeatedly use two elementary inequalities for $s,s'\ge0$ and $c>0$:
\begin{gather}
s\,|\rho|+s'\sqrt{1-\rho^2}\ \le\ \sqrt{s^2+s'^2} ,
\label{eq:bg-rho}\\
e^{-c/\tau}\le\frac\tau{\tau_0}\,e^{-c/\tau_0}\quad\text{for }0<\tau\le\tau_0\le c,
\qquad\text{and}\qquad
e^{-c/\tau}\le\frac\tau{ec}\quad\text{for all }\tau>0 .
\label{eq:bg-elem}
\end{gather}
Here \eqref{eq:bg-rho} is Cauchy--Schwarz applied to $(|\rho|,\sqrt{1-\rho^2})$, while \eqref{eq:bg-elem} follows from the fact that $\tau\mapsto e^{-c/\tau}/\tau$ has derivative $e^{-c/\tau}(c-\tau)/\tau^3$, making it increasing on $(0,c]$ and maximized globally at $\tau=c$.

\emph{Part~(i).} Set $\tau_u:=w_u/r\le\frac1{64}$ and choose $t=\frac9{25}$, so that $c_1(t)=\frac{25}{128}$, $c_2(t)=\frac{25}{24}$, and $(1-t)^2/2=0.2048$. Then $V_u\le\tau_u$, $\sqrt{V_uV_{uv}}\le\sqrt{1-\rho^2}\,\tau_u$, and (since $k_u\le w_u$) $p_t\le p_*:=e^{-0.2048/\tau_u}$. Grouping the terms of \eqref{eq:bg-bias-coarse} by $|\rho|$ and $\sqrt{1-\rho^2}$ and applying \eqref{eq:bg-rho} gives
\[
|b_{uv}|\le|\rho|\Big(\frac{25}{128}\tau_u+\frac{p_*}2\Big)+\sqrt{1-\rho^2}\,\Big(\frac{25}{24}\tau_u+\sqrt{p_*}\Big)
\le\tau_u\Big[\Big(\frac{25}{128}+\frac{p_*}{2\tau_u}\Big)^2+\Big(\frac{25}{24}+\frac{\sqrt{p_*}}{\tau_u}\Big)^2\Big]^{1/2}.
\]
Applying the first inequality in \eqref{eq:bg-elem} with $\tau_0=\frac1{64}$ (at $c=0.2048$ for $p_*/\tau_u$ and at $c=0.1024$ for $\sqrt{p_*}/\tau_u$) shows that the bracketed factor is maximized at $\tau_u=\frac1{64}$, where it evaluates to $1.1495\ldots<1.15$.

\emph{Part~(ii).} Choose $t=\frac12$, giving $c_1(t)=(\sqrt2-1)^2$, $c_2(t)=2(\sqrt2-1)$, and $p_{1/2}\le e^{-r/(8k_u)}$. Substituting $V_u=f_q(k_u-\bal)/r$, the second bound on $V_{uv}$, $|\rho|\le1$, and $k_u-\bal\le\sqrt{(k_u-\bal)\,\kmax}$ into the first two terms of \eqref{eq:bg-bias-coarse}, and applying \eqref{eq:bg-rho} with $p_{1/2}\le1$ to the last two terms yields
\[
\begin{aligned}
c_1(t)\,|\rho|\,V_u+c_2(t)\sqrt{V_uV_{uv}}
&\le\big(c_1(t)+2c_2(t)\big)\,\frac{f_q\sqrt{(k_u-\bal)\,\kmax}}r ,\\
\frac{|\rho|}2\,p_{1/2}+\sqrt{(1-\rho^2)\,p_{1/2}}&\le\sqrt{p_{1/2}\big(1+p_{1/2}/4\big)}\le\frac{\sqrt5}2\sqrt{p_{1/2}}<1.12\,e^{-r/(16k_u)}.
\end{aligned}
\]
Since $c_1(t)+2c_2(t)=2\sqrt2-1<1.83$, summing these two inequalities establishes the first bound of~(ii). When $\kmax\le r/70$, the first term is at most $1.83\,\kmax/r$, while the first inequality in \eqref{eq:bg-elem} with $c=\frac1{16}$, $\tau=k_u/r$, and $\tau_0=\frac1{70}$ bounds the second term by $1.12\cdot70\,e^{-70/16}\,k_u/r<0.99\,\kmax/r$; summing gives $(1.83+0.99)\kmax/r<2.83\,\kmax/r$.

\emph{Part~(iii).} Under balanced supports, $k_u=1$ implies $V_u=0$ and $p_t=0$, so every term in \eqref{eq:bg-bias-coarse} vanishes (matching Corollary~\ref{cor:flat-exact}).

\emph{Unconditional bound $4.3\,w_u/r$ of Remark~\ref{rem:bg-main}.} For arbitrary $\tau_u=w_u/r>0$, set $t=\frac14$, so that $c_1(t)=\frac29$, $c_2(t)=\frac43$, and $p_{1/4}\le e^{-9/(32\tau_u)}$. Grouping terms via \eqref{eq:bg-rho}, applying the triangle inequality in $\R^2$ to separate the $\tau_u$ and $p_{1/4}$ contributions, and using the second inequality of \eqref{eq:bg-elem} with $c=\frac9{64}$ yields
\[
|b_{uv}|\le\Big[\Big(\frac29\Big)^2+\Big(\frac43\Big)^2\Big]^{1/2}\tau_u+\frac{\sqrt5}2\,e^{-9/(64\tau_u)}
\le\Big(\frac{2\sqrt{37}}9+\frac{32\sqrt5}{9e}\Big)\tau_u<4.28\,\tau_u .
\]
\end{proof}

\begin{remark}[Sharpness of the BG bias bounds]\label{rem:bg-bias-sharp}
\emph{(a) The closed form $1.15\,w_u/r$.} Maximizing over all possible unit queries $v$ shows that the constant $1.15$ in Proposition~\ref{prop:bg-support}(i) is within a factor of roughly $4.5$ of the worst-case bias for a given cap $w_u$. Specifically, because $\zeta_\capset=q^{-1}\sum_{j\in\capset}\ut_j\vt_j\,\sigma_u^{-1}$ is linear in $\vt$ and $\Pr[j\in\capset]=q$, the bias can be written as $b_{uv}=\sum_j\ut_j\vt_j\,d_j$ with $d_j:=\E[\sigma_u^{-1}\mid j\in\capset]-1$ (for $\ut_j\ne 0$). Consequently, $\sup_{\|v\|=1}|b_{uv}|=\big(\sum_j\ut_j^2d_j^2\big)^{1/2}$. Evaluating this norm exactly on two-level vectors $\ut$ at $w_u/r=\frac1{64}$ produces biases up to $\approx 0.26\,w_u/r$. Indeed, to second order, no stored vector can exceed $w_u/(\sqrt{15}\,r)\approx 0.258\,w_u/r$: writing $x_j:=n\ut_j^2\in[0,w_u]$ (which has mean $1$, mean square $k_u$, and mean cube $m_3:=n^2\|\ut\|_6^6\le w_uk_u$), the approximation of Remark~\ref{rem:bg-estimate} takes the linear form $b_{uv}\approx\frac{f_q}r\sum_j\vt_j\ut_j\big(\frac{3k_u+\bal}8-\frac{x_j}2\big)$, whose Euclidean coefficient norm satisfies
\[
\sup_{\|v\|=1}|b_{uv}|\approx\frac{f_q}r\Big(\frac{m_3}4-\frac{15k_u^2+\bal\,(2k_u-1)}{64}\Big)^{1/2}
\le\frac1r\Big(\frac{w_uk_u}4-\frac{15k_u^2}{64}\Big)^{1/2}\le\frac{w_u}{\sqrt{15}\,r} ,
\]
where the final step maximizes over $k_u$ at $k_u=\frac8{15}w_u$.

\emph{(b) The certificate tail term.} In \eqref{eq:bg-bias-coarse}, the coefficient $1$ in front of the lower-tail term $\sqrt{(1-\rho^2)\,p_t}$ is exact and cannot be reduced. To see this, consider balanced supports with $r=1$ ($\capset=\{J\}$ for $J$ uniform on $[n]$ and $f_q=1$). Let $Q\subset[n]$ have size $n/4$, let $\ut$ be a positive unit vector taking one constant value on $Q$ and another on $Q^c$, and set $\vt$ proportional to $\mathbf 1-\inner{\mathbf 1}{\ut}\,\ut$, so that $\rho=0$. Then $\sigma_u^2=n\ut_J^2$ and $\zeta_\capset=\sqrt n\,\vt_J$, giving $b_{uv}=\E\zeta_\capset=n^{-1/2}\inner{\mathbf 1}{\vt}=(1-\|\ut\|_1^2/n)^{1/2}$. In the limit as the entries of $\ut$ on $Q$ tend to $0$, the entries of $\ut$ on $Q^c$ approach $(4/(3n))^{1/2}$ and those of $\vt$ on $Q^c$ approach $0$. Thus $\sigma_u^2\to 0$ for $J\in Q$ and $\sigma_u^2\to\frac43$ for $J\notin Q$, so that $b_{uv}\to\frac12$ and $p_t\to\frac14$ for every $t\in(0,1)$, while $V_{uv}=n\|\ut\circ\vt\|^2\to0$ and $V_u$ remains bounded. In this limit, the first three terms of \eqref{eq:bg-bias-coarse} vanish and the fourth term converges to $\sqrt{1/4}=\frac12=\lim b_{uv}$.
\end{remark}

\paragraph{Concentration of the dot-product decoder.}
By Lemma~\ref{lem:bg-cond}, the conditional moment generating function of a BG row score depends on the support $\capset$ solely through $\zeta_\capset$ and $\sigma_v^2$. We first bound the support-level fluctuations of these two quantities in terms of $\tau:=\wmax/r$.

\begin{lemma}[Fluctuations over the supports]\label{lem:bg-fluct}
Under either support law, fix $D$ and a pair $(u,v)$ satisfying $\tau:=\wmax/r\le\frac1{16}$. Then:
\begin{enumerate}
\item[(a)] $\Var(\sigma_u^2)\le\tau$, $\Var(\sigma_v^2)\le\tau$, and $\Var(\sigma_{uv})\le\tau$;
\item[(b)] $\E(\sigma_u^2-1)^4\le(3+\tau)\,\tau^2$;
\item[(c)] $\sqrt{\Var(\zeta_\capset)}\le2.2\sqrt\tau$;
\item[(d)] $\E\exp\{\lambda_0(\sigma_v^2-2)\}\le e^{-0.38/\tau}$ for $\lambda_0:=0.68/\tau$.
\end{enumerate}
\end{lemma}

\begin{proof}[Proof of Lemma~\ref{lem:bg-fluct}]
\emph{(a)} Because $w_u,w_v\le\tau r$, Lemma~\ref{lem:sampling}(i) immediately gives $\Var(\sigma_u^2)\le\tau$ and $\Var(\sigma_v^2)\le\tau$. Similarly, since $\sigma_{uv}$ is the captured sum of $\ut\circ\vt$ with total $\rho$, Lemma~\ref{lem:sampling} with $f_q\le1$ and $qn=r$ yields
\[
\Var(\sigma_{uv})=\frac{f_q}q\Big(\|\ut\circ\vt\|^2-\frac{\bal}n\,\rho^2\Big)\le\frac1q\,\|\ut\circ\vt\|^2=\frac{k_{uv}}r\le\frac{w_u}r\le\tau .
\]

\emph{Domination by independent summands.} For $x\in\{u,v\}$, let $S_*$ denote the sum of independent variables $W\in[0,w_x/r]$ constructed in Lemma~\ref{lem:sampling}(v) (with $\E S_*=1$ and $\sum\E W^2=k_x/r$). Because $t\mapsto(t-1)^4$ and $t\mapsto e^{\lambda_0t}$ are convex, Lemma~\ref{lem:sampling}(v) implies that their expectations under $\sigma_x^2$ are bounded above by their expectations under $S_*$. Writing $S_*-1=\sum \widetilde W$ in terms of the independent centered summands $\widetilde W:=W-\E W$, we have $|\widetilde W|\le w_x/r\le\tau$ and $\sum\E \widetilde W^2\le\sum\E W^2=k_x/r\le\tau$.

\emph{(b)} Taking $x=u$ and expanding $(S_*-1)^4=(\sum \widetilde W)^4$, only products in which no factor $\widetilde W$ appears with exponent $1$ survive in expectation. Using $\widetilde W^4\le\tau^2\widetilde W^2$, we obtain
\[
\E(S_*-1)^4\le\sum\E \widetilde W^4+3\Big(\sum\E \widetilde W^2\Big)^2\le\tau^2\cdot\tau+3\tau^2=(3+\tau)\,\tau^2 .
\]

\emph{(c)} Writing $\sigma_{uv}=\sigma_u\,\zeta_\capset$ (which holds even when $\sigma_u=0$) gives the identity $\zeta_\capset-\rho=(\sigma_{uv}-\rho)-(\sigma_u-1)\,\zeta_\capset$, where $|\sigma_u-1|\le|\eta|$ and $|\zeta_\capset|\le\sigma_v$ (Lemma~\ref{lem:bg-cond}). Since $\E\sigma_{uv}=\rho$, Minkowski's inequality yields $\sqrt{\Var(\zeta_\capset)}\le\sqrt{\E(\zeta_\capset-\rho)^2}\le\sqrt{\Var(\sigma_{uv})}+\sqrt{\E\eta^2\sigma_v^2}$. Applying Cauchy--Schwarz together with parts~(a) and~(b) bounds the second term via
\[
\E\eta^2\sigma_v^2=\E\eta^2+\E\eta^2(\sigma_v^2-1)\le\tau+\big(\E\eta^4\;\E(\sigma_v^2-1)^2\big)^{1/2}\le\tau+\tau\sqrt{\tau\,(3+\tau)}\le\tfrac{23}{16}\,\tau ,
\]
since $\tau\,(3+\tau)\le(\frac7{16})^2$ for $\tau\le\frac1{16}$. Combining this with $\Var(\sigma_{uv})\le\tau$ gives $\sqrt{\Var(\zeta_\capset)}\le(1+\frac{\sqrt{23}}4)\sqrt\tau\le2.2\sqrt\tau$.

\emph{(d)} Taking $x=v$, we apply Bernstein's moment generating function bound \eqref{eq:app-bernstein} to $S_*-1$ with $R=V=\tau$ and $\lambda=\lambda_0=0.68/\tau$ (noting $\lambda_0R=0.68<3$):
\[
\log\E e^{\lambda_0(S_*-1)}\le\frac{\lambda_0^2\,\tau}{2(1-\lambda_0\tau/3)}=\frac{0.68^2}{2(1-0.68/3)}\cdot\frac1\tau\le\frac{0.299}\tau .
\]
Multiplying by $e^{-\lambda_0}$ gives $\E e^{\lambda_0(\sigma_v^2-2)}\le e^{-\lambda_0}\,\E e^{\lambda_0(S_*-1)}\le e^{-(0.68-0.299)/\tau}\le e^{-0.38/\tau}$ (where the constant $c=0.68$ is chosen to nearly maximize $c-\frac{c^2}{2(1-c/3)}$).
\end{proof}

\begin{proposition}[BG row moment generating function]\label{prop:bg-mgf}
Under either support law, fix $D$ and a pair $(u,v)$ with $\wmax\le r/16$, and set $\bar\zeta:=\E[Z\mid D]$. Then for all $|\lambda|\le1/3$,
\[
\log\E e^{\lambda(Z-\bar\zeta)}\le\frac{\lambda^2}2\Big(\frac\pi2+9.25\,\frac{\wmax}r\Big).
\]
\end{proposition}

Here the additive correction $9.25\,\wmax/r$ over the dense variance parameter $\pi/2$ accounts for the support-to-support fluctuations of $\zeta_\capset$ and $\sigma_v^2$. As noted in Remark~\ref{rem:bg-main}, neither the constant $9.25$ nor the dependence on the cap $\wmax$ (rather than the fourth moment $\kmax$) is sharp: by Lemma~\ref{lem:bg-cond}, the exact unconditional variance is $\Var(Z\mid D)=\pi/2-\bar\zeta^2\le\pi/2$, so the true correction to the cumulant generating function is $O(\lambda^3)$ rather than $O(\lambda^2)$, and the variances of $\zeta_\capset$ and $\sigma_v^2$ scale with $\kmax/r$. We bound them via $\tau=\wmax/r$ simply because caps are already required to control the fourth moment and exponential tail in Lemma~\ref{lem:bg-fluct}(b) and~(d).

\begin{proof}[Proof of Proposition~\ref{prop:bg-mgf}]
Set $\tau:=\wmax/r\le\frac1{16}$.

\emph{Step 1: Conditioning on $\capset$.} Applying the tower property and Lemma~\ref{lem:bg-cond} gives
\[
\E e^{\lambda(Z-\bar\zeta)}\le\E e^{\lambda(\zeta_\capset-\bar\zeta)+\pi\lambda^2\sigma_v^2/4}=e^{\pi\lambda^2/4}\,\E e^\Xi,
\qquad\Xi:=\lambda(\zeta_\capset-\bar\zeta)+\tfrac\pi4\lambda^2(\sigma_v^2-1),
\]
where $\E\Xi=0$ since $\E\zeta_\capset=\bar\zeta$ and $\E\sigma_v^2=1$. It therefore suffices to prove $\log\E e^\Xi\le\frac{\lambda^2}2\cdot9.25\,\tau$ for $|\lambda|\le\frac13$.

\emph{Step 2: Typical supports ($\sigma_v^2\le2$).} By Taylor's theorem, $e^s=1+s+s^2I(s)$ where $I(s):=\int_0^1(1-t)\,e^{ts}\,dt$ is increasing and satisfies $I(s)\le\frac12e^{|s|}$ and $I(0.893)<0.69$. Since $|\zeta_\capset|\le\sigma_v$ and $|\bar\zeta|\le\E\sigma_v\le1$, we have
\begin{equation}
|\zeta_\capset-\bar\zeta|\le\sigma_v+1 .
\label{eq:bg-zeta-range}
\end{equation}
On the event $\{\sigma_v^2\le2\}$, this gives $|\zeta_\capset-\bar\zeta|\le\sqrt2+1$ and $|\sigma_v^2-1|\le1$; for $|\lambda|\le\frac13$, we thus have $\Xi\le\frac{1+\sqrt2}3+\frac\pi{36}<0.893$ and $I(\Xi)<0.69$. On $\{\sigma_v^2>2\}$, we use $I(\Xi)\le\frac12e^{|\Xi|}$. Taking expectations using $\E\Xi=0$, Minkowski's inequality, $\frac\pi4\lambda^2\le\frac\pi{12}|\lambda|$, and Lemma~\ref{lem:bg-fluct}(a) and~(c) yields
\[
\begin{aligned}
\E e^\Xi&=1+\E\big[\Xi^2I(\Xi)\big]\le1+0.69\,\E\Xi^2+\tfrac12\mathcal T,
\qquad\mathcal T:=\E\big[\Xi^2e^{|\Xi|}\,;\,\sigma_v^2>2\big],\\
\sqrt{\E\Xi^2}&\le|\lambda|\sqrt{\Var(\zeta_\capset)}+\frac\pi{12}\,|\lambda|\sqrt{\Var(\sigma_v^2)}\le\Big(2.2+\frac\pi{12}\Big)|\lambda|\sqrt\tau .
\end{aligned}
\]

\emph{Step 3: Atypical supports ($\sigma_v^2>2$).} When $\sigma_v^2>2$, we have $\sigma_v\le\sigma_v^2/\sqrt2$ and $1<\sigma_v^2/2$, so \eqref{eq:bg-zeta-range} gives $|\zeta_\capset-\bar\zeta|\le(\frac1{\sqrt2}+\frac12)\,\sigma_v^2$. Combining this with $0<\sigma_v^2-1<\sigma_v^2$ and $\frac\pi4\lambda^2\le\frac\pi{12}|\lambda|$ gives $|\Xi|\le C\,|\lambda|\,\sigma_v^2$ where $C:=\tfrac12+\tfrac1{\sqrt2}+\tfrac\pi{12}<\tfrac32$. For $|\lambda|\le\frac13$, this implies $|\Xi|\le\frac12\sigma_v^2$ and $\Xi^2e^{|\Xi|}\le C^2\lambda^2\,\sigma_v^4\,e^{\sigma_v^2/2}$. Now observe that every $s\ge2$ satisfies $s^2e^{s/2}\le4e\cdot e^{\lambda_0(s-2)}$: the two sides are equal at $s=2$, and the derivative of the logarithm of the left-hand side is $2/s+\frac12\le\frac32$, whereas $\lambda_0=0.68/\tau\ge0.68\cdot16>\frac32$. Applying this inequality at $s=\sigma_v^2$ and invoking Lemma~\ref{lem:bg-fluct}(d) with $4e\,C^2<23.6$ gives
\[
\mathcal T\le4e\,C^2\lambda^2\;\E e^{\lambda_0(\sigma_v^2-2)}\le23.6\,\lambda^2e^{-0.38/\tau} .
\]

\emph{Step 4: Combining the bounds.} Using $\log(1+s)\le s$ together with Steps~2 and~3 yields
\[
\log\E e^\Xi\le0.69\,\E\Xi^2+\tfrac12\mathcal T\le\frac{\lambda^2}2\Big(1.38\,\big(2.2+\tfrac\pi{12}\big)^2\tau+23.6\,e^{-0.38/\tau}\Big).
\]
Here $1.38\,(2.2+\frac\pi{12})^2<8.37$, while the first inequality of \eqref{eq:bg-elem} with $c=0.38$ and $\tau_0=\frac1{16}$ bounds the second term by $23.6\cdot16\,e^{-6.08}\,\tau<0.88\,\tau$. Summing gives $8.37+0.88=9.25$.
\end{proof}

\begin{lemma}[BG dot-product concentration]\label{lem:bg-conc}
Fix $D$ and an admissible pair $(u,v)$ satisfying $\wmax\le r/16$, and set $\bar\zeta=\E[Z\mid D]$. For every $0<t\le\frac12$,
\[
\Pr\big[|\hat\rho_{\rm BG}-\bar\zeta|\ge t\mid D\big]\le2\exp\Big\{-\frac{\ell t^2}{\pi+18.5\,\wmax/r}\Big\}.
\]
\end{lemma}

\begin{proof}[Proof of Lemma~\ref{lem:bg-conc}]
Set $\nu_{uv}:=\frac\pi2+9.25\,\wmax/r$. Because the pair is admissible, conditioned on $D$ the estimate $\hat\rho_{\rm BG}$ is the average of $\ell$ i.i.d.\ copies of $Z$. For any $0\le\lambda\le\frac13$, Markov's inequality and Proposition~\ref{prop:bg-mgf} yield $\Pr[\hat\rho_{\rm BG}-\bar\zeta\ge t\mid D]\le\exp\{-\ell\,(\lambda t-\frac12\lambda^2\nu_{uv})\}$. Optimizing over $\lambda$ selects $\lambda=t/\nu_{uv}$, which satisfies $\lambda\le\frac1\pi<\frac13$ since $t\le\frac12$ and $\nu_{uv}\ge\frac\pi2$. Substituting this choice gives $\exp\{-\ell t^2/(2\nu_{uv})\}$ with $2\nu_{uv}=\pi+18.5\,\wmax/r$, and repeating the argument with $-\lambda$ bounds the lower tail.
\end{proof}

\paragraph{Proofs of Theorem~\ref{thm:bg-param} and Corollary~\ref{cor:bg-classical}.}

\begin{proof}[Proof of Theorem~\ref{thm:bg-param}]
By the triangle inequality, each estimator's error is bounded by its deviation from the conditional mean $\bar\zeta=\rho+b_{uv}$ plus $|b_{uv}|$. Fix $D$ and treat the admissible pairs as deterministic.

\emph{(a)} Let $s_{uv}:=\sqrt{(\pi+18.5\,\wmax/r)\,L/\ell}$ denote the noise term in part~(a). The hypothesis $\ell\ge4(\pi+18.5\,\wmax/r)\,L$ is equivalent to $s_{uv}\le\frac12$. Because $\wmax\le r/64\le r/16$, applying Lemma~\ref{lem:bg-conc} at $t=s_{uv}$ bounds the failure probability for each pair by $2e^{-L}=\delta/N$, and a union bound over all $N$ pairs gives $|\hat\rho_{\rm BG}-\bar\zeta|\le s_{uv}$ with probability at least $1-\delta$. Adding $|b_{uv}|\le1.15\,w_u/r$ from Proposition~\ref{prop:bg-support}(i) completes the proof of~(a).

\emph{(b)} Conditioned on $D$, the row scores $Z_1,\dots,Z_\ell$ of each pair are i.i.d.\ with mean $\bar\zeta$ and variance $\frac\pi2-\bar\zeta^2\le\frac\pi2$ (Lemma~\ref{lem:bg-cond}). Applying the deviation form of Lemma~\ref{lem:catoni} with $\sigma_*^2=\frac\pi2$ guarantees $|\Cat_\alpha(Z_1,\dots,Z_\ell)-\bar\zeta|\le s$ across all $N$ pairs simultaneously with probability at least $1-\delta$, and adding $|b_{uv}|\le1.15\,w_u/r$ proves~(b).

\emph{Fourth-moment variants (Section~\ref{sec:bg}).} Because both proofs invoke Proposition~\ref{prop:bg-support}(i) only in their final step to bound $|b_{uv}|$, we may replace $1.15\,w_u/r$ directly with the fourth-moment bias bound of Proposition~\ref{prop:bg-support}(ii), which imposes no cap restriction. With that replacement, part~(b) requires no cap condition at all, while part~(a) requires only $\wmax\le r/16$ (from Lemma~\ref{lem:bg-conc}), under which $18.5\,\wmax/r\le 18.5/16<1.16$.
\end{proof}

\begin{proof}[Proof of Corollary~\ref{cor:bg-classical}]
\emph{Dense case ($r=n$).} When $r=n$, we have $\capset=[n]$ and $q=1$ under both support laws, so $\Phi=PHD$ has the exact distribution of a dense Gaussian matrix $G$ for every $D$, and $b_{uv}=0$. Part~(i) then follows from Theorem~\ref{thm:dense-bg} (since $4\log(4N/\delta)\ge\pi\log(2N/\delta)$), and part~(ii) follows from Lemma~\ref{lem:catoni}.

\emph{Sparse case ($r<n$).} Set $K:=|\Ufam|$ in~(i) and $K:=m$ in~(ii), so $r\ge16\max\{\eps^{-1},4\}\,\mu(K)$ and $\mu(K)/r\le\min\{\eps,\frac14\}/16\le\frac1{64}$. We split the failure budget $\delta$ equally between $D$ and $P$: applying Lemma~\ref{lem:flat} with $\delta_0=\delta/2$ yields the flatness bound $2\log(4nK/\delta)=\mu(K)$, and replacing $\delta$ with $\delta/2$ in the conditional step replaces $L=\log(2N/\delta)$ with $\log(4N/\delta)$.

\emph{(i)} With probability at least $1-\delta/2$ over $D$, every vector $x\in\Ufam$ satisfies $w_x\le\mu(|\Ufam|)$, so every pair satisfies $\wmax/r\le\frac1{64}$. Conditioned on such a $D$, Theorem~\ref{thm:bg-param}(a) (with failure budget $\delta/2$) gives noise term $s_{uv}\le\eps\,\sqrt{(\pi+18.5/64)/4}<0.927\,\eps<\frac12$ and bias term $1.15\,\mu(|\Ufam|)/r\le1.15\,\eps/16<0.072\,\eps$, whose sum is strictly below $\eps$.

\emph{(ii)} With probability at least $1-\delta/2$ over $D$, all $m$ stored vectors satisfy $w_u\le\mu(m)$, so Proposition~\ref{prop:bg-support}(i) gives $|b_{uv}|\le1.15\,\eps/16<0.1\,\eps$ for every pair $(u,v)$ (allowing $v$ to depend on $D$). For the noise, applying the fixed-accuracy form of Lemma~\ref{lem:catoni} conditioned on $D$ with $\sigma_*^2=\frac\pi2$, target accuracy $0.9\,\eps$, and failure budget $\delta/2$ requires $\ell\ge(\pi/(0.81\,\eps^2)+2)\log(4N/\delta)$, which holds since $\pi/0.81<3.9$; the total error is then at most $0.9\,\eps+0.1\,\eps=\eps$.

\emph{Encoding complexity and budget trade-off.} Since $\eps\le\frac12$, we have $\max\{\eps^{-1},4\}\le2/\eps$, so the minimum allowed sparsity is at most $\lceil32\,\mu(K)/\eps\rceil$, and multiplying by $\ell=O(\eps^{-2}\log(4N/\delta))$ gives $r\ell=O(\eps^{-3}\mu(K)\log(4N/\delta))$. Finally, to obtain the general budget split $\eps=\eps_s+\eps_b$ in Remark~\ref{rem:bg-main}, simply replace the conditional bias bound in~(ii) with the unconditional bound $|b_{uv}|\le4.3\,w_u/r\le4.3\,\mu(m)/r\le\eps_b$ from the proof of Proposition~\ref{prop:bg-support} and apply Lemma~\ref{lem:catoni} at accuracy $\eps_s$.
\end{proof}

\subsection{One bit on both sides (BB)}\label{app:bb}

This section establishes the SimHash (BB) guarantees of Section~\ref{sec:bb}. Fix $D$ and a pair of unit vectors $(u,v)$ at angle $\theta\in[0,\pi]$. Conditioned on its support $\capset$, a row produces a sign disagreement with probability $\theta_\capset/\pi$ (Lemma~\ref{lem:random-metric}(iii)), and the $\ell$ rows are conditionally independent given $D$. Thus $\hat p:=\hat\theta/\pi$ is the average of $\ell$ independent $\mathrm{Bernoulli}(p)$ variables with unconditional success probability $p:=\E\theta_\capset/\pi$, decomposing the angular error into
\begin{equation}\label{eq:bb-split}
\hat\theta-\theta\ =\ \underbrace{\pi\,(\hat p-p)}_{\text{noise}}\ +\ \underbrace{\E\theta_\capset-\theta}_{\text{support bias}} .
\end{equation}
While the sampling noise $\pi(\hat p-p)$ behaves just like dense SimHash, bounding the support bias $\E\theta_\capset-\theta$ requires expanding the distorted angle $\theta_\capset=\vartheta_\theta(M_\capset)$ around $M=I_2$ in powers of the \emph{metric sampling error} $E_\capset:=M_\capset-I_2$. Because the first-order derivative is linear in $E_\capset$ and $\E M_\capset=I_2$ (Lemma~\ref{lem:random-metric}(ii)), the first-order bias vanishes in expectation. On the event that $M_\capset$ has eigenvalues in $[\frac12,2]$, we bound the second-order Taylor remainder by $\frac78\sin\theta\,\|E_\capset\|_F^2$ (Lemma~\ref{lem:bb-expansion}), where the factor $\sin\theta$ makes the bias relative at small angles. Supports whose metric eigenvalues fall outside $[\frac12,2]$ contribute at most $\pi$ to the angular error and occur with probability exponentially small in $r/\wU$ (Lemma~\ref{lem:bb-tail}), producing the cap-tail term.

\paragraph{Derivatives and second-order expansion of the metric angle.}
Recall that in the orthonormal basis $U=[\,u_1\ \ u_2\,]$ of the plane, $u$ and $v$ have local coordinate vectors $e_1=(1,0)^\top$ and $t_\theta=(\cos\theta,\sin\theta)^\top$ (more generally, we write $t_\phi:=(\cos\phi,\sin\phi)^\top$ for any angle $\phi$). For any positive definite matrix $M\succ0$, the angle $\vartheta_\theta(M)\in[0,\pi]$ between $e_1$ and $t_\theta$ under $M$ has cosine $e_1^\top Mt_\theta/\{(e_1^\top Me_1)(t_\theta^\top Mt_\theta)\}^{1/2}$ and, by the Gram identity of Lemma~\ref{lem:random-metric}(iv), sine $\sin\theta\,\sqrt{\det M}/\{(e_1^\top Me_1)(t_\theta^\top Mt_\theta)\}^{1/2}$. Clearing the common positive denominator expresses $\vartheta_\theta(M)$ as a complex argument:
\begin{equation}\label{eq:bb-arg}
\vartheta_\theta(M)=\arg\Big(e_1^\top Mt_\theta+i\,\sin\theta\,\sqrt{\det M}\,\Big),\qquad M\succ0 .
\end{equation}
Three immediate properties follow from \eqref{eq:bb-arg}: first, $\vartheta_\theta(cM)=\vartheta_\theta(M)$ for all $c>0$ (scale invariance); second, when $\theta\in\{0,\pi\}$, we have $t_\theta=\pm e_1$ and $\sin\theta=0$, so $\vartheta_\theta(M)=\theta$ identically; third, when $0<\theta<\pi$, the imaginary part is strictly positive, so $\vartheta_\theta$ is smooth on the cone of positive definite $2\times2$ matrices.

Let $b:=t_{\theta/2}$ denote the unit bisector of $e_1$ and $t_\theta$, and let $b_\perp:=t_{\theta/2+\pi/2}$ denote its unit normal. For any $\theta\in[0,\pi]$ and any symmetric $2\times2$ matrix $A$, define the linear form
\begin{equation}
\vartheta'_\theta[A]:=-\frac{\sin\theta}2\,\big(b^\top Ab-b_\perp^\top Ab_\perp\big)=\frac{\sin\theta}2\big\{\cos\theta\,(A_{22}-A_{11})-2\sin\theta\,A_{12}\big\}.
\label{eq:bb-linear}
\end{equation}
Because $b$ and $b_\perp$ are orthonormal, both Rayleigh quotients $b^\top Ab$ and $b_\perp^\top Ab_\perp$ lie in $[\lambda_{\min}(A),\lambda_{\max}(A)]$, which immediately gives the eigenvalue-spread bound
\begin{equation}\label{eq:bb-linear-bound}
\big|\vartheta'_\theta[A]\big|\le\frac{\sin\theta}2\,\big\{\lambda_{\max}(A)-\lambda_{\min}(A)\big\}\qquad\text{for all }\theta\in[0,\pi].
\end{equation}
The next lemma identifies $\vartheta'_\theta$ as the Fr\'echet derivative of $\vartheta_\theta$ at $I_2$ and bounds the second directional derivative at an arbitrary metric $M\succ 0$ in terms of the \emph{whitened} perturbation $K=M^{-1/2}AM^{-1/2}$.

\begin{lemma}[Whitened derivatives of the angle]\label{lem:bb-whiten}
Let $0<\theta<\pi$, $M\succ0$, and $A$ be symmetric, and set $\phi:=\vartheta_\theta(M)$, $K:=M^{-1/2}AM^{-1/2}$, and $f(t):=\vartheta_\theta(M+tA)$ for small $|t|$.
\begin{enumerate}
\item[(i)] $|\vartheta'_\theta[A]|\le\frac{\sin\theta}2\{\lambda_{\max}(A)-\lambda_{\min}(A)\}$, and $f'(0)=\vartheta'_\theta[A]$ when $M=I_2$.
\item[(ii)] $|f''(0)|\le\frac{1+\sqrt5}4\sin\phi\,\|K\|_F^2$.
\end{enumerate}
\end{lemma}
\begin{proof}[Proof of Lemma~\ref{lem:bb-whiten}]
\emph{Step 1: Reduction to $M=I_2$.} For any invertible matrix $B$, the angle between $x$ and $y$ under the metric $B^\top NB$ equals the angle between $Bx$ and $By$ under $N$, since $x^\top B^\top NBy=(Bx)^\top N(By)$. Factoring $M+tA=M^{1/2}(I_2+tK)M^{1/2}$ with $B=M^{1/2}$ shows that $f(t)$ is the angle between $M^{1/2}e_1$ and $M^{1/2}t_\theta$ under the metric $I_2+tK$. At $t=0$, their Euclidean angle is $f(0)=\phi\in(0,\pi)$. Rotating the plane by an orthogonal matrix $O$ that aligns the directions of $M^{1/2}e_1$ and $M^{1/2}t_\theta$ with $e_1$ and $t_\phi$ (and using the fact that angles are invariant to vector lengths) gives $f(t)=\vartheta_\phi(I_2+t\,OKO^\top)$ for small $|t|$. Since $\|OKO^\top\|_F=\|K\|_F$ (and $M=I_2$ gives $\phi=\theta$, $O=I_2$, $K=A$), it suffices to show that for any $\phi\in(0,\pi)$ and symmetric $K$, the function $g(t):=\vartheta_\phi(I_2+tK)$ satisfies $g'(0)=\vartheta'_\phi[K]$ and $|g''(0)|\le\frac{1+\sqrt5}4\,\sin\phi\,\|K\|_F^2$.

\emph{Step 2: Complex logarithm differentiation.} For small $|t|$, the matrix $I_2+tK$ is positive definite, and \eqref{eq:bb-arg} gives $g(t)=\arg\xi(t)=\operatorname{Im}\log\xi(t)$ where
\[
\xi(t)=\cos\phi+t\,e_1^\top Kt_\phi+i\,\sin\phi\,d(t),\qquad d(t):=\big(1+t\operatorname{tr}K+t^2\det K\big)^{1/2}.
\]
Differentiating twice yields $g'=\operatorname{Im}(\xi'/\xi)$ and $g''=\operatorname{Im}(\xi''/\xi)-\operatorname{Im}\{(\xi'/\xi)^2\}$.

\emph{Step 3: Orthogonal reflection basis.} Let $R$ be the reflection across the bisector $t_{\phi/2}$ of $e_1$ and $t_\phi$, and let $R'$ be the reflection across the line at angle $\pi/4$ to the bisector. We expand $K$ in the orthogonal basis $\{I_2,R,R'\}$:
\[
K=\frac{\kappa_0I_2+\kappa_1R+\kappa_2R'}{\sqrt2},\qquad
R=\begin{pmatrix}\cos\phi&\sin\phi\\ \sin\phi&-\cos\phi\end{pmatrix},\qquad
R'=\begin{pmatrix}-\sin\phi&\cos\phi\\ \cos\phi&\sin\phi\end{pmatrix}.
\]
Because $R$ and $R'$ are symmetric, traceless, and anticommuting ($RR'=-R'R$) with $\|I_2\|_F^2=\|R\|_F^2=\|R'\|_F^2=2$, we have $\|K\|_F^2=\kappa_0^2+\kappa_1^2+\kappa_2^2$ and $\operatorname{tr}K=\sqrt2\,\kappa_0$. Moreover, the traceless part $T:=(\kappa_1R+\kappa_2R')/\sqrt2$ satisfies $T^2=\frac12(\kappa_1^2+\kappa_2^2)I_2$, so $\det T=-\frac12(\kappa_1^2+\kappa_2^2)$ and $\det K=\frac12(\kappa_0^2-\kappa_1^2-\kappa_2^2)$. Differentiating $d(t)^2$ twice at $t=0$ gives $d'(0)=\frac12\operatorname{tr}K=\kappa_0/\sqrt2$ and $d''(0)=\det K-\frac14(\operatorname{tr}K)^2=-\frac12(\kappa_1^2+\kappa_2^2)$. Finally, since $Rt_\phi=e_1$ and $R't_\phi=(0,1)^\top$, we have $e_1^\top Kt_\phi=(\kappa_0\cos\phi+\kappa_1)/\sqrt2$. Evaluating $\xi$, $\xi'$, and $\xi''$ at $t=0$ therefore gives $\xi(0)=e^{i\phi}$, $\xi'(0)=(\kappa_0e^{i\phi}+\kappa_1)/\sqrt2$, and $\xi''(0)=-\frac{i\sin\phi}2\,(\kappa_1^2+\kappa_2^2)$.

\emph{Step 4: Evaluating $g'(0)$ and $g''(0)$.} Dividing by $\xi(0)=e^{i\phi}$ gives $\xi'(0)/\xi(0)=(\kappa_0+\kappa_1e^{-i\phi})/\sqrt2$, whose square has imaginary part $-\sin\phi\,(\kappa_0\kappa_1+\kappa_1^2\cos\phi)$, while $\xi''(0)/\xi(0)$ has imaginary part $-\frac12\sin\phi\cos\phi\,(\kappa_1^2+\kappa_2^2)$. Substituting these into Step~2 yields
\begin{equation}
g'(0)=-\frac{\sin\phi}{\sqrt2}\,\kappa_1,\qquad
g''(0)=\sin\phi\,\Big[\kappa_0\kappa_1+\frac{\cos\phi}2\,(\kappa_1^2-\kappa_2^2)\Big].
\label{eq:bb-derivs}
\end{equation}
Because $R=bb^\top-b_\perp b_\perp^\top$ for $b=t_{\phi/2}$, we have $\sqrt2\,\kappa_1=\operatorname{tr}(KR)=b^\top Kb-b_\perp^\top Kb_\perp$, so $g'(0)=\vartheta'_\phi[K]$ by \eqref{eq:bb-linear}. In the expression for $g''(0)$, the quadratic form inside the brackets has eigenvalue $-\frac12\cos\phi$ along $\kappa_2$ and eigenvalues $\frac14\big(\cos\phi\pm\sqrt{4+\cos^2\phi}\big)$ (the roots of $x^2-\frac12x\cos\phi-\frac14=0$) on the $(\kappa_0,\kappa_1)$-plane. All three eigenvalues are bounded in magnitude by $\frac14\big(|\cos\phi|+\sqrt{4+\cos^2\phi}\big)\le\frac{1+\sqrt5}4$, proving~(ii).
\end{proof}

Integrating the second-derivative bound of Lemma~\ref{lem:bb-whiten}(ii) along the line segment from $I_2$ to $M$ yields our key deterministic Taylor expansion for well-conditioned metrics.

\begin{lemma}[Second-order expansion of the angle]\label{lem:bb-expansion}
Let $\theta\in[0,\pi]$, let $M$ be a symmetric $2\times2$ matrix satisfying $\frac12I_2\preceq M\preceq2I_2$, and set $E:=M-I_2$. Then
\[
\big|\vartheta_\theta(M)-\theta-\vartheta'_\theta[E]\big|\le\tfrac78\,\sin\theta\,\|E\|_F^2 .
\]
\end{lemma}
\begin{proof}[Proof of Lemma~\ref{lem:bb-expansion}]
When $\theta\in\{0,\pi\}$, both sides vanish identically by \eqref{eq:bb-arg} and \eqref{eq:bb-linear}. For $0<\theta<\pi$, define $M(t):=I_2+tE$ and $\phi(t):=\vartheta_\theta(M(t))$ for $t\in[0,1]$. Because the eigenvalues of $E$ lie in $[-\frac12,1]$, those of $M(t)$ lie in $[1-\frac t2,\,1+t]$, so $\phi(t)$ is smooth on $[0,1]$ with $\phi(0)=\theta$, $\phi(1)=\vartheta_\theta(M)$, and $\phi'(0)=\vartheta'_\theta[E]$. By Taylor's theorem with integral remainder, $\vartheta_\theta(M)-\theta-\vartheta'_\theta[E]=\int_0^1(1-t)\,\phi''(t)\,dt$. Applying Lemma~\ref{lem:bb-whiten}(ii) at $M(t)$ in direction $E$ gives $|\phi''(t)|\le\frac{1+\sqrt5}4\,\sin\phi(t)\,\|K(t)\|_F^2$ where $K(t):=M(t)^{-1/2}EM(t)^{-1/2}$. Because $M(t)$ and $E$ commute, each eigenvalue $\lambda\ge-\frac12$ of $E$ maps to $\lambda/(1+t\lambda)$ in $K(t)$, so $\|K(t)\|_F\le\|E\|_F/(1-\frac t2)$, while the Gram identity (Lemma~\ref{lem:random-metric}(iv)) gives $\sin\phi(t)\le\sin\theta\,\{(1+t)/(1-\frac t2)\}^{1/2}$. Using $(1+t)(1-\frac t2)\le\frac98$ on $[0,1]$, we obtain
\[
|\phi''(t)|\le\frac{1+\sqrt5}4\,\sin\theta\,\|E\|_F^2\,\frac{\{(1+t)(1-\frac t2)\}^{1/2}}{(1-\frac t2)^{3}}
\le\frac{3(1+\sqrt5)}{8\sqrt2}\,\sin\theta\,\|E\|_F^2\,\big(1-\tfrac t2\big)^{-3}.
\]
Since $\int_0^1(1-t)(1-\frac t2)^{-3}\,dt=1$ and $\frac{3(1+\sqrt5)}{8\sqrt2}<0.86<\frac78$, integrating over $t\in[0,1]$ completes the proof.
\end{proof}

\paragraph{Bounding the support bias.}
Taking expectations of the expansion in Lemma~\ref{lem:bb-expansion} over the random support $\capset$ yields the coordinatewise pair certificate for SimHash.

\begin{proposition}[BB support bias, certificate form]\label{prop:bb-bias}
Fix $D$ and a pair $(u,v)$ at angle $\theta\in[0,\pi]$, and set $p:=\Pr[\sign X\ne\sign Y\mid D]$, $E_\capset:=M_\capset-I_2$, and
\[
V_M:=\E\|E_\capset\|_F^2=\frac{f_q\,(\kU-2\bal)}{r}\le\frac{2\wU}{r},\qquad \mathcal G:=\{\tfrac12I_2\preceq M_\capset\preceq2I_2\}.
\]
Then $|\pi p-\theta|\le\tfrac78\sin\theta\,V_M+\big(\max\{\theta,\pi-\theta\}+\tfrac18\sin\theta\big)\Pr[\mathcal G^c]$, where the \emph{tail coefficient} multiplying $\Pr[\mathcal G^c]$ is at most $\pi$.
\end{proposition}
\begin{proof}[Proof of Proposition~\ref{prop:bb-bias}]
By Lemma~\ref{lem:random-metric}(ii) and~(iii), $\pi p=\E\theta_\capset$ and $\E E_\capset=0$. Because $\vartheta'_\theta$ is linear, $\E\vartheta'_\theta[E_\capset]=0$, so $\pi p-\theta=\E\big[\theta_\capset-\theta-\vartheta'_\theta[E_\capset]\big]$. We claim that every support $\capset$ satisfies the pointwise bound
\[
\big|\theta_\capset-\theta-\vartheta'_\theta[E_\capset]\big|\le\tfrac78\sin\theta\,\|E_\capset\|_F^2+\big(\max\{\theta,\pi-\theta\}+\tfrac18\sin\theta\big)\mathbf 1_{\mathcal G^c}.
\]
Taking expectations then yields the stated bias bound (noting that $\max\{\theta,\pi-\theta\}+\frac18\sin\theta\le\pi$ since $\sin\theta\le\min\{\theta,\pi-\theta\}$).

\emph{Supports in $\mathcal G$.} On $\mathcal G$, we have $M_\capset\succ0$ and $\theta_\capset=\vartheta_\theta(M_\capset)$, so the pointwise inequality holds by Lemma~\ref{lem:bb-expansion}.

\emph{Supports in $\mathcal G^c$.} Because $\theta_\capset,\theta\in[0,\pi]$ always, $|\theta_\capset-\theta|\le\max\{\theta,\pi-\theta\}$. Let $\lambda_1,\lambda_2$ be the eigenvalues of $E_\capset$ ordered so that $|\lambda_1|\ge|\lambda_2|$. By \eqref{eq:bb-linear-bound}, $|\vartheta'_\theta[E_\capset]|\le\frac{\sin\theta}2(|\lambda_1|+|\lambda_2|)$. On $\mathcal G^c$, at least one eigenvalue of $E_\capset$ lies outside $[-\frac12,1]$, forcing $|\lambda_1|>\frac12$. Because the quadratic $x\mapsto\frac x2-\frac78x^2$ attains its global maximum $\frac1{14}$ at $x=\frac27$ and is decreasing on $[\frac27,\infty)$ with value $\frac1{32}$ at $x=\frac12$, we have $\frac12|\lambda_2|\le\frac78\lambda_2^2+\frac1{14}$ and $\frac12|\lambda_1|\le\frac78\lambda_1^2+\frac1{32}$. Since $\frac1{14}+\frac1{32}<\frac18$ and $\lambda_1^2+\lambda_2^2=\|E_\capset\|_F^2$, summing gives $|\vartheta'_\theta[E_\capset]|\le\sin\theta\,\big(\frac78\|E_\capset\|_F^2+\frac18\big)$, establishing the pointwise bound on $\mathcal G^c$.

\emph{Expected squared Frobenius error $V_M$.} Each of the four entries of $M_\capset$ is a captured sum $q^{-1}\sum_{j\in\capset}y_j$ of the corresponding entry of $\cvec_j\cvec_j^\top$, with expectation equal to the corresponding entry of $I_2$. Summing the variance formula of Lemma~\ref{lem:sampling} across all four entries gives
\[
V_M=\frac{f_q}{q}\Big(\sum_j\|\cvec_j\cvec_j^\top\|_F^2-\frac{\bal}{n}\|I_2\|_F^2\Big)
=\frac{f_q}{q}\Big(\sum_j\|\cvec_j\|^4-\frac{2\bal}{n}\Big)
=\frac{f_q\,(\kU-2\bal)}{r},
\]
since $q=r/n$ and $\kU=n\sum_j\|\cvec_j\|^4$. Finally, $V_M\le2\wU/r$ follows from $f_q\le1$ and $\kU\le2\wU$ (Lemma~\ref{lem:params}(c)).
\end{proof}

Notice why the $-2\bal$ reduction appears in $V_M$ under balanced supports: under Bernoulli supports, the support size $|\capset|\sim\mathrm{Binomial}(n,q)$ has variance $r(1-q)=rf_q$, so the conditional mean metric $(|\capset|/r)I_2$ fluctuates in trace, adding $\|I_2\|_F^2\Var(|\capset|/r)=2f_q/r$ to $V_M$. Although rescaling $M_\capset$ does not alter the angle $\vartheta_\theta(M_\capset)$, Lemma~\ref{lem:bb-whiten}(ii) bounds the second derivative via the full Frobenius norm $\|K\|_F^2$ (including the trace component $\kappa_0^2$).

\begin{remark}[Coordinate cancellation in the leading BB bias term]\label{rem:bb-second-order}
For $0<\theta<\pi$, let $\vartheta''_\theta[A]$ denote the second directional derivative of $t\mapsto \vartheta_\theta(I_2+tA)$ at $t=0$ (given by \eqref{eq:bb-derivs} with $\phi=\theta$ and $K=A$), so that the leading second-order term of the BB support bias is $b_2:=\frac12\E\vartheta''_\theta[E_\capset]$. Applying the covariance formula of Lemma~\ref{lem:sampling} to the quadratic form $\vartheta''_\theta$ (and noting $\vartheta''_\theta[I_2]=0$ by scale invariance) gives $\E\vartheta''_\theta[E_\capset]=\frac{f_q}q\sum_j\vartheta''_\theta[\cvec_j\cvec_j^\top]$ under both support laws. Writing each row vector $\cvec_j=\|\cvec_j\|\,t_{\theta/2+\chi_j}$ in polar coordinates relative to the bisector $t_{\theta/2}$, the rank-one matrix $\cvec_j\cvec_j^\top$ has coordinates $2^{-1/2}\|\cvec_j\|^2(1,\cos2\chi_j,\sin2\chi_j)$ in the basis $\{I_2,R,R'\}$ of Lemma~\ref{lem:bb-whiten}, so \eqref{eq:bb-derivs} yields the exact expression
\[
b_2=\frac{f_q\sin\theta}{4q}\sum_j\|\cvec_j\|^4\Big\{\cos2\chi_j+\tfrac12\cos\theta\,\cos4\chi_j\Big\}.
\]
Crucially, this sum is signed: when the angles $\chi_j$ are distributed nearly uniformly around the circle (as occurs for a generic Gaussian plane), the trigonometric factors $\cos2\chi_j$ and $\cos4\chi_j$ oscillate in sign and nearly cancel, making $|b_2|$ orders of magnitude smaller than the Frobenius bound of Proposition~\ref{prop:bb-bias} (Section~\ref{sec:exp-tight}).

In the worst case, however, where all $\chi_j$ align with the extremal eigendirections of $\vartheta''_\theta$, the constant $\frac78$ in Proposition~\ref{prop:bb-bias} is within a small factor of the sharp second-order constant. Specifically, under Bernoulli supports ($V_M=\frac{f_q}q\sum_j\|\cvec_j\|^4$), bounding the trigonometric bracket by $1+\frac12|\cos\theta|$ gives $|b_2|\le\frac14\big(1+\frac12|\cos\theta|\big)\sin\theta\,V_M$, which is attained in the limit $n_0/n\to0$ by planes with $\cvec_j=n_0^{-1/2}t_{\theta/2}$ on $n_0$ coordinates and $\cvec_j=(n-n_0)^{-1/2}t_{\theta/2+\pi/2}$ on the remaining $n-n_0$ coordinates (swapping the two directions when $\theta>\frac\pi2$). Under balanced supports ($V_M=\frac{f_q}q(\sum_j\|\cvec_j\|^4-\frac2n)$), the eigenvalue bound of Lemma~\ref{lem:bb-whiten} gives $|b_2|\le\tfrac18\big(|\cos\theta|+\sqrt{4+\cos^2\theta}\big)\sin\theta\,V_M$, which is attained by the same two-direction construction when $n_0/n\to\frac14\big(2+|\cos\theta|-\sqrt{4+\cos^2\theta}\big)$ ($\approx 0.19$ as $\theta\to 0$ or $\pi$). Thus the coefficient $\frac78$ exceeds the sharp second-order worst-case constant by a factor of at most $\frac73\approx 2.33$ (Bernoulli) or $7/(1+\sqrt5)\approx 2.16$ (balanced) as $\theta\to 0,\pi$, and $\frac72=3.5$ at $\theta=\frac\pi2$.
\end{remark}

To complete the BB bias bound, we bound the atypical-metric probability $\Pr[\mathcal G^c]$ in two ways: first in terms of the plane cap $\wU$ via matrix Chernoff (used in Theorem~\ref{thm:bb-param}), and second in terms of coordinate variance and range parameters $(V_U,R_U)$ via matrix Bernstein (used in the coordinatewise BB pair certificate of Section~\ref{sec:exp-tight}).

\begin{lemma}[Metric tail]\label{lem:bb-tail}
Fix $D$ and a pair $(u,v)$. Under either support law, $\Pr[\mathcal G^c]\le4\exp\{-(1-\log2)\,r/(2\wU)\}$. In addition, $\Pr[\mathcal G^c]\le4\exp\{-1/(8V_U+4R_U/3)\}$, where, setting $C_j:=n\cvec_j\cvec_j^\top-I_2$,
\begin{alignat*}{3}
&\text{Bernoulli:}\qquad&V_U&=\tfrac{f_q}q\,\big\|{\textstyle\sum_j}\|\cvec_j\|^2\cvec_j\cvec_j^\top\big\|_{\rm op},\qquad&R_U&=\max\{1,\tfrac{f_q}q\}\max\nolimits_j\|\cvec_j\|^2,\\
&\text{balanced:}&V_U&=\tfrac1{rn}\,\big\|{\textstyle\sum_j}C_j^2\big\|_{\rm op},&R_U&=\tfrac1r\max\nolimits_j\|C_j\|_{\rm op}.
\end{alignat*}
\end{lemma}
\begin{proof}[Proof of Lemma~\ref{lem:bb-tail}]
We apply the matrix Chernoff bound \eqref{eq:app-matrix} to $M_\capset$ and the matrix Bernstein bound \eqref{eq:app-matrix-bernstein} to $E_\capset$. Writing $\xi_j:=\mathbf 1\{j\in\capset\}$, we have $M_\capset=\sum_{j=1}^n\frac{\xi_j}q\,\cvec_j\cvec_j^\top$ and $E_\capset=\sum_{j=1}^n\big(\frac{\xi_j}q-1\big)\cvec_j\cvec_j^\top$, whose summands are independent under Bernoulli supports. Under balanced supports, $M_\capset=\sum_{j\in\capset}\frac nr\cvec_j\cvec_j^\top$ and $E_\capset=\sum_{j\in\capset}\frac1rC_j$; by the transfer principle of Appendix~\ref{app:tools}, both matrix concentration inequalities apply with the parameters of $r$ independent draws $\frac nr\cvec_J\cvec_J^\top$ and $\frac1rC_J$ for $J$ uniform on $[n]$.

\emph{Matrix Chernoff bound.} Under both laws, each summand of $M_\capset$ is positive semidefinite with maximum eigenvalue at most $n\max_j\|\cvec_j\|^2/r=\wU/r$, and the summand expectations sum to $I_2$. Because $\mathcal G^c=\{\lambda_{\min}(M_\capset)<\frac12\}\cup\{\lambda_{\max}(M_\capset)>2\}$, applying \eqref{eq:app-matrix} with $R=\wU/r$ bounds the two tail probabilities by $2\exp\{-h(-\frac12)\,r/\wU\}$ and $2\exp\{-h(1)\,r/\wU\}$; since $h(-\frac12)=\frac12(1-\log2)<h(1)=2\log2-1$, summing gives $4\exp\{-(1-\log2)\,r/(2\wU)\}$.

\emph{Matrix Bernstein bound.} On $\mathcal G^c$, at least one eigenvalue of $E_\capset$ lies outside $[-\frac12,1]$, so $\|E_\capset\|_{\rm op}>\frac12$. Under Bernoulli supports, $\xi_j/q-1$ has mean zero, range $[-1,f_q/q]$, and second moment $f_q/q$, so each summand of $E_\capset$ has operator norm at most $R_U$ and their expected squares sum to $\frac{f_q}q\sum_j\|\cvec_j\|^2\cvec_j\cvec_j^\top$ (of operator norm $V_U$). Under balanced supports, each independent draw $\frac1rC_J$ is centered (as $\sum_jC_j=0$) with operator norm at most $R_U$, and the sum of expected squares over $r$ draws is $\frac1{rn}\sum_jC_j^2$ (of operator norm $V_U$). Applying \eqref{eq:app-matrix-bernstein} at $t=\frac12$ yields $\Pr[\|E_\capset\|_{\rm op}\ge\frac12]\le4\exp\{-1/(8V_U+4R_U/3)\}$.
\end{proof}

\paragraph{Row concentration and proofs of Theorem~\ref{thm:bb-param} and Corollary~\ref{cor:bb-classical}.}
Conditioned on $D$, for any admissible pair, the empirical disagreement fraction $\hat p=\hat\theta/\pi$ is the average of $\ell$ i.i.d.\ $\mathrm{Bernoulli}(p)$ variables. Consequently, for every $\tau>0$ and every $V\ge p(1-p)$, Hoeffding's inequality \eqref{eq:app-hoeffding} and Bernstein's inequality \eqref{eq:app-bernstein} yield
\begin{equation}\label{eq:bb-noise}
\Pr\big[|\hat p-p|\ge\tau\,\big|\,D\big]\le2e^{-2\ell\tau^2}
\qquad\text{and}\qquad
\Pr\big[|\hat p-p|\ge\tau\,\big|\,D\big]\le2\exp\Big\{-\frac{\ell\tau^2}{2(V+\tau/3)}\Big\}.
\end{equation}

\begin{proof}[Proof of Theorem~\ref{thm:bb-param}]
By \eqref{eq:bb-split}, $|\hat\theta-\theta|\le|\pi p-\theta|+\pi|\hat p-p|$. Set $L:=\log(2N/\delta)$. Combining Proposition~\ref{prop:bb-bias} (with tail coefficient bounded by $\pi$) and the Chernoff bound of Lemma~\ref{lem:bb-tail} gives
\begin{equation}
|\pi p-\theta|\le\frac78\,\sin\theta\,\frac{f_q\,(\kU-2\bal)}{r}+4\pi\exp\Big(-\frac{(1-\log2)\,r}{2\wU}\Big).
\label{eq:bb-two-terms}
\end{equation}

(a) Evaluating the Hoeffding bound in \eqref{eq:bb-noise} at $\tau=\sqrt{L/(2\ell)}$ bounds each pair's failure probability by $2e^{-L}=\delta/N$. Taking a union bound over all $N$ pairs and adding \eqref{eq:bb-two-terms} establishes the additive bound for $|\hat\theta-\theta|$ (which transfers to $|\hat\rho_{\rm BB}-\rho|$ since cosine is $1$-Lipschitz). When $\wU\le r/64$, using $f_q\le1$ and $\kU\le2\wU$ bounds the first term of \eqref{eq:bb-two-terms} by $1.75\,\wU/r$, while setting $y:=r/\wU\ge64$ and noting that $y\mapsto ye^{-(1-\log2)y/2}$ is decreasing for $y\ge2/(1-\log2)\approx6.52$ bounds the second term by $4\pi\cdot64\,e^{-32(1-\log2)}\,\wU/r\le0.0438\,\wU/r$; summing gives $(1.75+0.0438)\wU/r\le1.8\,\wU/r$.

(b) Fix any pair with $\theta\ge\theta_0$, and split the target error $\eps\theta$ equally between bias and sampling noise.

\emph{Bias.} Using $\sin\theta\le\theta$, $f_q\le1$, and $r\ge3.5\,\kU/\eps$, the first term of \eqref{eq:bb-two-terms} is at most $\frac78\theta\,\kU/r\le\eps\theta/4$. Since $2/(1-\log2)<6.52$, the condition $r\ge6.52\,\wU\log(16\pi/(\eps\theta_0))$ ensures that the second term of \eqref{eq:bb-two-terms} is at most $4\pi\cdot\eps\theta_0/(16\pi)\le\eps\theta/4$. Thus $|\pi p-\theta|\le\eps\theta/2$, which in particular implies $p\le(1+\frac\eps2)\theta/\pi$.

\emph{Noise.} Applying the Bernstein bound in \eqref{eq:bb-noise} with $\tau=\eps\theta/(2\pi)$ and $V=(1+\frac\eps2)\theta/\pi\ge p(1-p)$ gives exponent
\[
\frac{\ell\tau^2}{2(V+\tau/3)}=\frac{\ell\,\eps^2\theta}{8\pi\,(1+\frac23\eps)}\ge\frac{\ell\,\eps^2\theta_0}{42}\ge L ,
\]
since $\eps\le1$ and $8\pi\cdot\frac53\le42$. Consequently, $\pi|\hat p-p|<\eps\theta/2$ except with probability at most $2e^{-L}=\delta/N$. Pairs with $\theta=0$ have $v=u$, so $Y_i=X_i$ in every row and $\hat\theta=0$ deterministically. A union bound over the pairs with $\theta\ge\theta_0$ completes the proof.
\end{proof}

\begin{proof}[Proof of Corollary~\ref{cor:bb-classical}]
\emph{Dense case ($r=n$).} When $r=n$, $\Phi=PHD$ has the exact law of $G$, so Theorem~\ref{thm:dense-bb} applies directly (as its sketch-length constants $\frac{\pi^2}2\le2\pi^2$ and $8.4\le42$ are smaller than ours, including in the general $\eta$-split of Remark~\ref{rem:bb-main}).

\emph{Sparse case ($r<n$).} Applying Lemma~\ref{lem:flat} with $\delta_0=\delta/2$ to the at most $2N$ basis vectors $(u_1,u_2)$ ensures that, with probability at least $1-\delta/2$ over $D$, all of them satisfy $w_x\le2\log(8nN/\delta)=\mu(2N)$, which implies $\wU\le w_{u_1}+w_{u_2}\le2\,\mu(2N)$ and $\kU\le2\wU\le4\,\mu(2N)$ for every pair (Lemma~\ref{lem:params}(c)). Conditioned on such a $D$, we apply Theorem~\ref{thm:bb-param} with failure budget $\delta/2$ (so $L=\log(4N/\delta)$). Part~(ii) follows immediately from Theorem~\ref{thm:bb-param}(b) since $3.5\,\kU/\eps\le14\,\mu(2N)/\eps$ and $6.52\,\wU\le13.04\,\mu(2N)\le13.5\,\mu(2N)$. For part~(i) and the general budget split $\eta\in(0,1)$ of Remark~\ref{rem:bb-main}, substituting $\kU\le4\,\mu(2N)$ and $\frac2{1-\log2}\wU\le\frac4{1-\log2}\mu(2N)<13.04\,\mu(2N)$ into Theorem~\ref{thm:bb-param}(a) bounds its three terms by
\[
\frac78\cdot\frac{4\,\mu(2N)}r,\qquad 4\pi\exp\Big(-\frac r{13.04\,\mu(2N)}\Big),\qquad\text{and}\qquad\pi\sqrt{\frac{\log(4N/\delta)}{2\ell}}.
\]
The first two terms are each at most $\eta\eps/2$ once $r\ge7\,\mu(2N)/(\eta\eps)$ and $r\ge13.04\,\mu(2N)\allowbreak\cdot\allowbreak\log(8\pi/(\eta\eps))$, while the third is at most $(1-\eta)\eps$ once $\ell\ge\frac{\pi^2}2(1-\eta)^{-2}\eps^{-2}\log(4N/\delta)$; setting $\eta=\frac12$ gives part~(i).

\emph{Encoding complexity.} Because $\eps\mapsto\eps\log(16\pi/\eps)$ is increasing on $(0,\pi]$ (with derivative $\log(16\pi/\eps)-1>0$), we have $\log(16\pi/\eps)\le\pi\log16/\eps$, so the minimum allowed sparsity in~(i) scales as $r=O(\mu(2N)/\eps)$ and multiplying by $\ell$ gives $r\ell=O(\eps^{-3}\mu(2N)\log(4N/\delta))$.
\end{proof}

\subsection{Lower bounds}\label{app:lower}

In this section, we prove Theorem~\ref{thm:lower} using the normalized Walsh blocks introduced in Section~\ref{sec:lower}. Lemma~\ref{lem:lb-walsh} characterizes the structure of a Walsh block of size $s$ when the random signs $D$ align with a Walsh function on its support (a ``hit'' block), and Proposition~\ref{prop:lb-bias} computes the resulting support biases for BG and BB. Scored against itself, a hit block of size $s$ has expected BG score $1-\Delta_s$, where the deficit $\Delta_s$ is roughly one eighth of the variance $V_s$ of its captured energy. General pairs constructed by combining a hit block with a standard basis vector inherit this exact deficit because the off-diagonal cross term $\sigma_{12}$ changes sign under a measure-preserving involution of the supports. Finally, Lemma~\ref{lem:lb-hit} selects the largest block size $s$ for which $D$ hits at least one block in a family of size $N$ with probability exceeding $\delta$, and the proof of Theorem~\ref{thm:lower} converts the resulting conditional bias into a high-probability estimation failure via Cantelli's inequality.

\paragraph{Setup and Kronecker structure of the Hadamard matrix.}
Throughout this section, let $n\ge4$ and $2\le r\le n/2$ (so $q=r/n\le\frac12$), and let $s$ be a power of two satisfying $2\le s\le n/2$. By Definition~\ref{def:balanced}, $f_q\ge1-q\ge\frac12$ under both support laws, so the energy variance $V_s:=f_q(s-\bal)/r$ on a hit block satisfies
\begin{equation}
V_s\ \ge\ \frac{s-1}{2r}\ \ge\ \frac s{4r}\qquad\text{and}\qquad f_q\,(s-1)\ \ge\ \frac{s-1}2\ \ge\ \frac s4 .
\label{eq:lb-fq}
\end{equation}

We index the coordinates of $\R^n$ from $0$ to $n-1$ and write each index uniquely as $i=ms+y$ with $0\le m<n/s$ and $0\le y<s$. For any power of two $d$, let $H_d$ denote the normalized $d\times d$ Walsh--Hadamard matrix in standard Sylvester order ($H_1=(1)$ and $H_{2d}=2^{-1/2}\begin{psmallmatrix}H_d&H_d\\H_d&-H_d\end{psmallmatrix}$), which is symmetric and orthogonal with entries $\pm d^{-1/2}$. By recursion, $H=H_n$ factors as the Kronecker product $H_{n/s}\otimes H_s$:
\begin{equation}
H_{ms+y,\,m's+y'}=(H_{n/s})_{mm'}\,(H_s)_{yy'} .
\label{eq:lb-kron}
\end{equation}
(We may assume Sylvester ordering without loss of generality, since permuting the rows of $H$ simply permutes the columns of $P$, whose distribution is permutation-invariant under both support laws.)

For each $0\le b<n/s$, let $B_b:=\{bs,\dots,bs+s-1\}$ denote the $b$th dyadic \emph{block} of size $s$, let $x_b:=s^{-1/2}\mathbf 1_{B_b}$ be its normalized Walsh block vector, and let $d_b\in\{\pm1\}^s$ with entries $(d_b)_y:=D_{bs+y}$ collect the signs of $D$ restricted to $B_b$. By \eqref{eq:lb-kron}, the preconditioned vector $\xt_b=HDx_b$ has entries $\xt_{b,\,ms+y}=(H_{n/s})_{mb}\,s^{-1/2}\,(H_sd_b)_y$. We say that $D$ \emph{hits} block $B_b$---and write $\mathcal E_b$ for this event---if $d_b$ coincides up to a global sign $\varsigma_b\in\{\pm1\}$ with a column of $\sqrt s\,H_s$, meaning $d_b=\varsigma_b\sqrt s\,(H_s)_{\cdot,y_b}$ for some $y_b\in\{0,\dots,s-1\}$. By orthogonality of $H_s$, the vector $H_sd_b$ then equals $\varsigma_b\sqrt s$ in coordinate $y_b$ and vanishes everywhere else, so that
\begin{equation}
\xt_{b,\,ms+y}=\varsigma_b\,(H_{n/s})_{mb}\;\mathbf 1\{y=y_b\}.
\label{eq:lb-flat}
\end{equation}
Geometrically, on the hit event $\mathcal E_b$, the preconditioner $HD$ maps a contiguous block of $s$ coordinates into a regular ``comb'' of $n/s$ coordinates spaced by $s$, with one nonzero in each block (Figure~\ref{fig:lb-comb} in Section~\ref{sec:lower}). Meanwhile, for any standard basis vector $e_j$, its preconditioned image $\tilde e_j:=HDe_j$ is simply $D_j$ times column $j$ of $H$, whose entries all have magnitude $n^{-1/2}$.

\begin{lemma}[Hit blocks]\label{lem:lb-walsh}
In the setup above, every block $B_b$ ($0\le b<n/s$) satisfies the following properties:
\begin{enumerate}
\item[(i)] \emph{Hit probability and independence.} $\Pr[\mathcal E_b]=2s\,2^{-s}$, and the hit events $\mathcal E_0,\dots,\mathcal E_{n/s-1}$ are mutually independent (for $s=2$, $\mathcal E_b$ holds deterministically for every $D$).
\item[(ii)] \emph{Comb profile.} On $\mathcal E_b$, $|\xt_{b,i}|=\sqrt{s/n}\;\mathbf 1\{i\in C_b\}$, where $C_b:=\{ms+y_b:0\le m<n/s\}$ is the comb of $n/s$ coordinates congruent to $y_b$ modulo $s$. Consequently, $w_{x_b}=k_{x_b}=s$, and a row with support $\capset$ captures from $x_b$ the energy $\sigma_b^2:=\|\xt_{b,\capset}\|^2=sJ_b/r$, where $J_b:=|\capset\cap C_b|$ counts the sampled comb coordinates.
\item[(iii)] \emph{Antisymmetry of the cross term.} On $\mathcal E_b$, fix any $j\notin B_b$. The entrywise products $\tilde e_{j,i}\,\xt_{b,i}$ vanish outside $C_b$, equal $\pm\sqrt s/n$ on $C_b$, and sum to zero over $i\in C_b$. Thus $C_b$ partitions into two equal halves $C_b^+$ and $C_b^-$ of size $n/(2s)$ according to the sign of $\tilde e_{j,i}\,\xt_{b,i}$. Writing $J_b^\pm:=|\capset\cap C_b^\pm|$, we have $\sigma_b^2=(s/r)\,(J_b^++J_b^-)$ and $\sigma_{12}:=q^{-1}\sum_{i\in\capset}\tilde e_{j,i}\,\xt_{b,i}=(\sqrt s/r)\,(J_b^+-J_b^-)$. Moreover, under either support law, there is a probability-preserving involution $\capset\mapsto\capset^*$ on the space of supports that swaps $J_b^+$ and $J_b^-$, so that the \emph{mirror support} $\capset^*$ has the same captured energy $\sigma_b^2$ as $\capset$ and negated cross term $-\sigma_{12}$.
\item[(iv)] \emph{Plane parameters.} On $\mathcal E_b$, fix $j\notin B_b$ and $0<\theta<\pi/2$. The pair $u=e_j$, $v=\cos\theta\,e_j+\sin\theta\,x_b$ has orthonormal plane basis $U=[\,e_j\ \ x_b\,]$, $k_u=1$, $\wU=s+1$, $\kU=s+3$, and $k_v=1+4\cos^2\theta\sin^2\theta+(s-1)\sin^4\theta$, which is strictly below $1.01$ whenever $\tan\theta\le1/(33\sqrt s)$.
\end{enumerate}
\end{lemma}
\begin{proof}[Proof of Lemma~\ref{lem:lb-walsh}]
(i) The sign subvector $d_b$ is uniformly distributed on $\{\pm1\}^s$, and the $2s$ vectors $\pm\sqrt s\,(H_s)_{\cdot,y}$ ($0\le y<s$) are distinct by column orthogonality of $H_s$ (exhausting all $4$ vectors when $s=2$). Independence across $b$ follows because disjoint blocks $B_b$ depend on disjoint subsets of the diagonal entries of $D$.

(ii) Since every entry of $H_{n/s}$ has magnitude $\sqrt{s/n}$, equation \eqref{eq:lb-flat} shows that $\xt_{b,i}^2=s/n$ for $i\in C_b$ and $0$ otherwise. Thus $w_{x_b}=n(s/n)=s$, $k_{x_b}=n(n/s)(s/n)^2=s$, and $\sigma_b^2=q^{-1}(s/n)\,J_b=sJ_b/r$.

(iii) Since $|\tilde e_{j,i}|=n^{-1/2}$ everywhere, part~(ii) implies that $\tilde e_{j,i}\,\xt_{b,i}=\pm\sqrt s/n$ on $C_b$ and $0$ outside $C_b$. Because $HD$ is orthogonal and $j\notin B_b$, the sum of these products is $\inner{\tilde e_j}{\xt_b}=\inner{e_j}{x_b}=0$, so exactly half of the $n/s$ coordinates of $C_b$ have positive product ($C_b^+$) and half have negative product ($C_b^-$). Fixing a bijection between $C_b^+$ and $C_b^-$ and letting $\capset^*$ be the image of $\capset$ under the coordinate involution that swaps paired elements of $C_b^\pm$ while fixing all coordinates outside $C_b$ defines a probability-preserving involution on supports that exchanges $J_b^+$ and $J_b^-$.

(iv) Since $e_j\perp x_b$, we have $\rho=\cos\theta$, $u_1=e_j$, $u_2=x_b$, and $k_u=1$ (as $\tilde e_j$ is flat). By~(ii), the plane leverage score $\|\cvec_i\|^2=\tilde e_{j,i}^2+\xt_{b,i}^2$ equals $(s+1)/n$ on the $n/s$ coordinates of $C_b$ and $1/n$ on the remaining $n-n/s$ coordinates, so \eqref{eq:frame} gives $\wU=s+1$ and $\kU=(s+1)^2/s+(1-1/s)=s+3$. By~(iii), $|\vt_i|=n^{-1/2}\,|\cos\theta\pm\sqrt s\sin\theta|$ on $C_b^\pm$ and $n^{-1/2}\cos\theta$ outside $C_b$. Summing $n\vt_i^4$ cancels the odd powers of $\sqrt s\sin\theta$ between $C_b^+$ and $C_b^-$, giving
\[
k_v=\cos^4\theta+6\cos^2\theta\sin^2\theta+s\sin^4\theta=1+4\cos^2\theta\sin^2\theta+(s-1)\sin^4\theta
\ \le\ 1+4\tan^2\theta+s\tan^4\theta .
\]
When $s\tan^2\theta\le 33^{-2}=1/1089$, we have $s\tan^4\theta\le\tan^2\theta\le1/1089$ (as $s\ge2$), so $k_v\le1+5/1089<1.01$.
\end{proof}

\begin{proposition}[Attained biases]\label{prop:lb-bias}
In the setup above (with $r\le n/2$), fix $D$ and use either support law unless specified otherwise.
\begin{enumerate}
\item[(i)] \emph{Self-pairs.} For any unit vector $u$, $\E[\hat\rho_{\rm BG}(u,u)\mid D]=\E\sigma_u=1-\Delta(u)$, where the deficit $\Delta(u):=\frac12\E(\sigma_u-1)^2$ satisfies the exact identity
\[
\Delta(u)=\frac{\Var(\sigma_u^2)}8-\frac{\E(\sigma_u^2-1)^3}{16}+\frac1{16}\,\E\Big[\big(\sigma_u-1\big)^4\big(\sigma_u^2+4\sigma_u+5\big)\Big].
\]
In particular, $\Delta(u)\ge\Var(\sigma_u^2)/8-\E(\sigma_u^2-1)^3/16$ and $\Delta(u)\le\Var(\sigma_u^2)/8+\frac38\,\E(1-\sigma_u^2)_+^3$.
\item[(ii)] \emph{Hit blocks.} On $\mathcal E_b$, the captured energy $\sigma_b^2$ has variance $V_s=f_q(s-\bal)/r$, and $\Delta_s:=\Delta(x_b)$ depends only on $n$, $r$, $s$, and the support law, satisfying
\[
\frac{V_s}8\Big(1-\frac s{2r}\Big)\ \le\ \Delta_s\ \le\ \frac{V_s}8\Big(1+46\sqrt{s/r}\,\Big).
\]
Consequently, $\Delta_s\ge V_s/16$ whenever $s\le r$, and $\Delta_s=(1+o(1))\,V_s/8$ as $s/r\to0$.
\item[(iii)] \emph{Attained BG bias.} On $\mathcal E_b$, for any $j\notin B_b$, $\rho\in[-1,1]$, and $v=\rho\,x_b+\sqrt{1-\rho^2}\,e_j$, we have $\E[\hat\rho_{\rm BG}(x_b,v)\mid D]-\rho=-\rho\,\Delta_s$.
\item[(iv)] \emph{Attained BB bias.} Under balanced supports with $s\le r$, on $\mathcal E_b$, for any $j\notin B_b$, $0<\theta<\pi/2$ with $\tan\theta\le\frac1{10}$, $u=e_j$, and $v=\cos\theta\,e_j+\sin\theta\,x_b$,
\[
\E[\hat\theta\mid D]-\theta\ \le\ -\sin\theta\cos\theta\,\Delta_s+\frac\pi4\,\Pr\big[\sigma_2^2\ge\cot^2\theta\big]
\ \le\ -\theta\,\frac{f_q(s-1)}{17\,r} ,
\]
where $\sigma_2^2=\sigma_b^2$ is the energy captured from $u_2=x_b$ (and $\Pr[\sigma_2^2\ge\cot^2\theta]=0$ whenever $s\tan^2\theta<1$, since $\sigma_b^2\le s$).
\end{enumerate}
\end{proposition}
\begin{proof}[Proof of Proposition~\ref{prop:lb-bias}]
(i) When $v=u$, we have $\sigma_{uv}=\sigma_u^2$ and $\E[Z\mid\capset]=\sigma_u$ (Lemma~\ref{lem:random-metric}(iii)), so $\E[\hat\rho_{\rm BG}(u,u)\mid D]=\E\sigma_u$. Using $\E\sigma_u^2=1$ gives $1-\E\sigma_u=\E\big[\frac12(\sigma_u^2+1)-\sigma_u\big]=\frac12\,\E(\sigma_u-1)^2=\Delta(u)$. Using $\sigma_u^2-1=(\sigma_u-1)(\sigma_u+1)$, direct algebraic expansion verifies the polynomial identity
\[
(\sigma_u-1)^2=\frac{(\sigma_u^2-1)^2}4-\frac{(\sigma_u^2-1)^3}8+\frac{(\sigma_u-1)^4\,(\sigma_u^2+4\sigma_u+5)}8 .
\]
Taking expectations and dividing by $2$ yields the exact identity for $\Delta(u)$, and dropping the nonnegative last term gives the lower bound. For the upper bound, write $(\sigma_u-1)^2=(\sigma_u^2-1)^2/(1+\sigma_u)^2$: when $\sigma_u\ge1$, we have $(1+\sigma_u)^2\ge4$, so $(\sigma_u-1)^2\le\frac14(\sigma_u^2-1)^2$; when $0\le\sigma_u<1$, using $4-(1+\sigma_u)^2=(1-\sigma_u)(3+\sigma_u)$ and $1-\sigma_u^2=(1-\sigma_u)(1+\sigma_u)$ gives
\[
(\sigma_u-1)^2-\frac{(\sigma_u^2-1)^2}4=\frac{(\sigma_u^2-1)^2}4\cdot\frac{(1-\sigma_u)(3+\sigma_u)}{(1+\sigma_u)^2}=\frac{(1-\sigma_u^2)^3}4\cdot\frac{3+\sigma_u}{(1+\sigma_u)^3}
\ \le\ \frac34\,(1-\sigma_u^2)^3 ,
\]
since $(3+\sigma_u)/(1+\sigma_u)^3\le 3$ for $\sigma_u\ge0$. Combining both cases gives $(\sigma_u-1)^2\le\frac14(\sigma_u^2-1)^2+\frac34(1-\sigma_u^2)_+^3$, and taking half the expectation proves the upper bound.

(ii) By Lemma~\ref{lem:lb-walsh}(ii), $\sigma_b^2=sJ_b/r$, where $J_b\sim\mathrm{Binomial}(n/s,q)$ under Bernoulli supports and $J_b\sim\mathrm{Hypergeometric}(n,n/s,r)$ under balanced supports. Neither distribution depends on $D$, and $\Var(\sigma_b^2)=V_s$ by Lemma~\ref{lem:sampling}(i).

\emph{Lower bound.} The third central moment of $J_b$ equals $\Var(J_b)\,(1-2q)$ in the binomial case and
\[
\E(J_b-\E J_b)^3=\Var(J_b)\,\Big(1-\frac2s\Big)\,\frac{n-2r}{n-2}
\]
in the hypergeometric case (see Chapter~6 of Johnson, Kemp, and Kotz, \emph{Univariate Discrete Distributions}). Because $r\le n/2$ and $s\ge2$, we have $\E(J_b-\E J_b)^3\le\Var(J_b)$ under both laws. Scaling by $(s/r)^3$ gives $\E(\sigma_b^2-1)^3\le(s/r)^3\,\Var(J_b)=(s/r)\,V_s$, so part~(i) yields $\Delta_s\ge V_s/8-(s/r)\,V_s/16$.

\emph{Upper bound.} Applying Maurer's lower-tail bound $\Pr[\sigma_b^2<t]\le e^{-(1-t)^2r/(2s)}$ (Lemma~\ref{lem:sampling}(ii) with $k_{x_b}=s$) gives
\[
\E(1-\sigma_b^2)_+^3=\int_0^13(1-t)^2\,\Pr[\sigma_b^2<t]\,dt\ \le\ 3\int_{-\infty}^1(1-t)^2e^{-(1-t)^2r/(2s)}\,dt
=3\sqrt{\pi/2}\,\Big(\frac sr\Big)^{3/2}.
\]
Since $\frac38\cdot3\sqrt{\pi/2}\le1.41$ and $s/r\le4V_s$ by \eqref{eq:lb-fq}, part~(i) yields $\Delta_s\le V_s/8+1.41\,(s/r)^{3/2}\le V_s/8+5.64\,V_s\sqrt{s/r}\le(V_s/8)(1+46\sqrt{s/r})$.

(iii) Here $u=x_b$ is the stored vector, so $\sigma_u=\sigma_b$ and $\sigma_{uv}=\rho\,\sigma_b^2+\sqrt{1-\rho^2}\,\sigma_{12}$ (with $\sigma_{12}$ defined in Lemma~\ref{lem:lb-walsh}(iii)). By Lemma~\ref{lem:random-metric}(iii), $\E[Z\mid\capset]=\rho\sigma_b+\sqrt{1-\rho^2}\;\sigma_{12}\,\sigma_b^{-1}\,\mathbf 1\{\sigma_b>0\}$ (noting that $\sigma_b=0\implies\sigma_{12}=0$). Because replacing $\capset$ with its mirror support $\capset^*$ (Lemma~\ref{lem:lb-walsh}(iii)) preserves $\sigma_b$ and negates $\sigma_{12}$, the second term has expectation zero, leaving $\E[\hat\rho_{\rm BG}\mid D]=\rho\,\E\sigma_b=\rho\,(1-\Delta_s)$.

(iv) Set $T:=\tan\theta$. For the pair $u=e_j$ and $v=\cos\theta\,e_j+\sin\theta\,x_b$, we have $u_1=e_j$ and $u_2=x_b$, so the random metric \eqref{eq:random-metric} has diagonal entries $\sigma_1^2=1$ (deterministically on every balanced support since $\tilde e_j$ is flat) and $\sigma_2^2=\sigma_b^2$, and off-diagonal entry $\sigma_{12}$ from Lemma~\ref{lem:lb-walsh}(iii).

\emph{Complex representation of $\theta_\capset$.} Using $t_\theta=\cos\theta\,(1,T)^\top$, we have $\sigma_{uv}=\cos\theta\,(1+T\sigma_{12})$ and $\sigma_v^2=\cos^2\theta\,(1+2T\sigma_{12}+T^2\sigma_2^2)$. Because $\sigma_2^2\ge\sigma_{12}^2$ (as $M_\capset\succeq0$), we may define the upper half-plane complex number $\xi_\capset:=1+T\sigma_{12}+i\,T\sqrt{\sigma_2^2-\sigma_{12}^2}$, which satisfies $\operatorname{Re}\xi_\capset=\sigma_{uv}/\cos\theta$ and $|\xi_\capset|=\sigma_v/\cos\theta$. Defining $\arg\xi\in[0,\pi]$ for $\xi\ne0$ in the closed upper half-plane and $\arg0:=\pi/2$, Lemma~\ref{lem:random-metric}(iii) (with $\sigma_u=\sigma_1=1$) implies $\theta_\capset=\arg\xi_\capset$ for every support $\capset$.

\emph{Averaging a support and its mirror.} Because the mirror support $\capset^*$ has the same $\sigma_2^2$ and negated cross term $-\sigma_{12}$, multiplying $\xi_\capset$ and $\xi_{\capset^*}$ gives
\[
\xi_\capset\,\xi_{\capset^*}=\big(1+i\,T\sqrt{\sigma_2^2-\sigma_{12}^2}\big)^2-T^2\sigma_{12}^2=1-T^2\sigma_2^2+2i\,T\sqrt{\sigma_2^2-\sigma_{12}^2}.
\]
We claim that every support satisfies
\begin{equation}
\tfrac12\big(\theta_\capset+\theta_{\capset^*}\big)\ \le\ \arctan\big(T\sigma_2\big)+\frac\pi4\,\mathbf 1\{T^2\sigma_2^2\ge1\}.
\label{eq:lb-pair}
\end{equation}
Indeed, if $\xi_\capset=0$ (or $\xi_{\capset^*}=0$), then $T\sigma_{12}=-1$ and $\sigma_2^2=\sigma_{12}^2$, so $T^2\sigma_2^2=1$ and the two angles are $\pi/2$ and $0$, satisfying \eqref{eq:lb-pair}. Otherwise, $\operatorname{Re}\xi_\capset+\operatorname{Re}\xi_{\capset^*}=2>0$, so at least one of $\theta_\capset,\theta_{\capset^*}$ is strictly below $\pi/2$, implying $\theta_\capset+\theta_{\capset^*}<3\pi/2$. Because $\xi_\capset\,\xi_{\capset^*}$ lies in the closed upper half-plane, its unique argument in $[0,3\pi/2)$ is $\arg(\xi_\capset\,\xi_{\capset^*})\in[0,\pi]$, so $\theta_\capset+\theta_{\capset^*}=\arg(\xi_\capset\,\xi_{\capset^*})\le\pi$. When $T^2\sigma_2^2\ge1$, we have $\arctan(T\sigma_2)\ge\pi/4$, so \eqref{eq:lb-pair} holds. When $T^2\sigma_2^2<1$, the product $\xi_\capset\,\xi_{\capset^*}$ has positive real part $1-T^2\sigma_2^2$, and the tangent double-angle identity gives
\[
\theta_\capset+\theta_{\capset^*}=\arctan\frac{2T\sqrt{\sigma_2^2-\sigma_{12}^2}}{1-T^2\sigma_2^2}\ \le\ \arctan\frac{2T\sigma_2}{1-T^2\sigma_2^2}
=2\arctan\big(T\sigma_2\big),
\]
establishing \eqref{eq:lb-pair}.

\emph{Expectation and constants.} Taking expectations of \eqref{eq:lb-pair} over $\capset$ and using concavity of $x\mapsto\arctan(Tx)$ on $[0,\infty)$ (bounding it by its tangent line at $x=1$, which has slope $T/(1+T^2)=\sin\theta\cos\theta$) yields
\begin{align*}
\E[\hat\theta\mid D]=\E\,\tfrac12(\theta_\capset+\theta_{\capset^*})&\le\E\arctan\big(T\sigma_2\big)+\frac\pi4\,\Pr\big[\sigma_2^2\ge T^{-2}\big]\\
&\le\theta+\sin\theta\cos\theta\,\big(\E\sigma_2-1\big)+\frac\pi4\,\Pr\big[\sigma_2^2\ge\cot^2\theta\big],
\end{align*}
proving the first inequality in~(iv) since $\E\sigma_2=\E\sigma_b=1-\Delta_s$. Finally, using $\Delta_s\ge V_s/16$ (from~(ii) as $s\le r$), Chebyshev's inequality $\Pr[\sigma_2^2\ge T^{-2}]\le V_s/(T^{-2}-1)^2=V_s\,T^4/(1-T^2)^2$, and $\sin\theta\cos\theta=T/(1+T^2)$ bounds the bias by
\[
-T\,V_s\,\Big(\frac1{16\,(1+T^2)}-\frac\pi4\cdot\frac{T^3}{(1-T^2)^2}\Big)\ \le\ -\theta\,\frac{f_q(s-1)}{17\,r} ,
\]
since $T\ge\theta$, $V_s=f_q(s-1)/r$ under balanced supports, and the parenthetical factor is decreasing in $T$ with value $>0.061>\frac1{17}$ at $T=\frac1{10}$.
\end{proof}

To prove parts~(b)--(d) of Theorem~\ref{thm:lower}, we pack disjoint blocks $B_1,\dots,B_{N'}$ of size $s$ (reserving $B_0$ for the basis vector $e_0$ in part~(c)) and choose $s$ so that $D$ hits at least one block with probability exceeding $\delta$.

\begin{lemma}[A likely hit]\label{lem:lb-hit}
Let $N\ge1$, $\delta\in(0,\frac12]$, and $\Lambda=\log(\min\{N,n/2\}/\delta)$, and let $2\le s_{\max}\le n/2$. Let $s$ be the largest power of two satisfying
\[
s\le s_{\max}\qquad\text{and}\qquad2^s\le\min\{sN,\,n/2\}/\delta ,
\]
and set $N':=\min\{N,\,n/s-1\}$. Then $N'\ge1$, the union event $\mathcal E_{\rm hit}:=\mathcal E_1\cup\dots\cup\mathcal E_{N'}$ has probability $\Pr[\mathcal E_{\rm hit}]>\delta$, and
\begin{equation}
s>\tfrac12\min\{s_{\max},\ \Lambda/\log2\}.\label{eq:lb-maxk}
\end{equation}
\end{lemma}
\begin{proof}[Proof of Lemma~\ref{lem:lb-hit}]
Because $s=2$ satisfies $2^2=4\le 2/\delta\le\min\{2N,n/2\}/\delta$, the dyadic integer $s\ge2$ is well defined, and $s\le n/2$ implies $N'\ge1$. The second condition ensures $sN'=\min\{sN,n-s\}\ge\min\{sN,n/2\}\ge\delta\,2^s$, so Lemma~\ref{lem:lb-walsh}(i) gives
\[
\Pr[\mathcal E_{\rm hit}]=1-\big(1-2s\,2^{-s}\big)^{N'}\ \ge\ 1-e^{-2sN'2^{-s}}\ \ge\ 1-e^{-2\delta}\ >\ \delta ,
\]
where the final inequality uses the concavity of $\delta\mapsto 1-e^{-2\delta}-\delta$ on $[0,\frac12]$. Finally, by maximality of $s$, the next power of two $2s$ must violate one of the two defining constraints: either $2s>s_{\max}$, or $2^{2s}>\min\{2sN,n/2\}/\delta\ge\min\{N,n/2\}/\delta=e^\Lambda$ (so $2s>\Lambda/\log2$), which proves \eqref{eq:lb-maxk}.
\end{proof}

\begin{proof}[Proof of Theorem~\ref{thm:lower}]
(a) Setting $x:=x_0$ and $\mathcal E:=\mathcal E_0$, Lemma~\ref{lem:lb-walsh} and Proposition~\ref{prop:lb-bias}(i)--(iv) immediately yield all claims in part~(a).

\emph{Reduction via block selection and Cantelli's inequality.} For parts~(b) and~(c), invoke Lemma~\ref{lem:lb-hit} with $s_{\max}=r$ to obtain $s$, $N'\le N$, and $\mathcal E_{\rm hit}$. On $\mathcal E_{\rm hit}$, let $b^*\in\{1,\dots,N'\}$ be the smallest block index for which $\mathcal E_{b^*}$ holds (note that $b^*$ is measurable with respect to $D$).

Suppose a family of $N'$ pairs indexed by $b=1,\dots,N'$ satisfies the following \emph{conditional miss property} with constant $c>0$ for $\eps\le 1/(2c)$: whenever $r<s/(c\,\eps)$ and $D\in\mathcal E_{\rm hit}$, the estimator for pair $b^*$ fails to achieve its target accuracy with conditional probability at least $\frac12$ over $P$. Then any sketch that succeeds simultaneously across all $N'$ pairs with overall probability at least $1-\delta/2$ must satisfy $r\ge s/(c\,\eps)$ (otherwise its failure probability would be at least $\frac12\Pr[\mathcal E_{\rm hit}]>\delta/2$). Moreover, $r\ge s/(c\,\eps)$ and $\eps\le 1/(2c)$ imply $s\le r/2<s_{\max}$, so the active branch in \eqref{eq:lb-maxk} is $\Lambda/\log2$, yielding $r\ge s/(c\,\eps)>\Lambda/(2c\log2\cdot\eps)$. Establishing the conditional miss property with $c=128$ for BG and $c=136$ for BB therefore proves parts~(b) and~(c), since $256\log2<178$ and $272\log2<189$.

To establish the conditional miss property from a bias, we use \emph{Cantelli's inequality}: if, conditioned on $D$, an estimator $W$ of a target parameter $\tau$ has expectation $\E[W\mid D]=\tau-t$ ($t>0$) and variance $\Var(W\mid D)<t^2/4$, then
\[
\Pr\big[W-\E[W\mid D]\ge t/2\ \big|\ D\big]\ \le\ \frac{\Var(W\mid D)}{\Var(W\mid D)+t^2/4}\ <\ \frac12 ,
\]
so $W<\tau-t/2$ with conditional probability strictly exceeding $\frac12$.

(b) Consider the family of $N'$ self-pairs $(x_b,x_b)$ ($1\le b\le N'$), where $\rho=1$. Suppose $r<s/(128\,\eps)$ and $D\in\mathcal E_{\rm hit}$. Since $s\le r$, Proposition~\ref{prop:lb-bias}(ii) and \eqref{eq:lb-fq} give $\Delta_s\ge V_s/16\ge s/(64r)>2\eps$, and Proposition~\ref{prop:lb-bias}(i) gives $\E[Z_1\mid D]=1-\Delta_s$ for pair $b^*$. For the dot-product decoder with $\ell\ge\pi/(2\eps^2)$, Lemma~\ref{lem:bg-cond} gives $\Var(\hat\rho_{\rm BG}\mid D)\le\pi/(2\ell)\le\eps^2<(\Delta_s/2)^2$, so Cantelli's inequality with $\tau=1$ and $t=\Delta_s$ implies $|\hat\rho_{\rm BG}-1|>\Delta_s/2>\eps$ with conditional probability $>\frac12$. Likewise, any decoder $\tilde\rho$ that concentrates within $\eps$ of $\E[Z_1\mid D]=1-\Delta_s$ with conditional probability at least $\frac12$ satisfies $|\tilde\rho-1|\ge\Delta_s-\eps>\eps$ on that event (in particular, by Lemma~\ref{lem:catoni} with $N=1$, $\delta=\frac12$, and $\sigma_*^2=\pi/2$, Catoni's decoder satisfies this condition once $\ell\ge2(\pi/2+\eps^2)\,\eps^{-2}\log4$). Thus the conditional miss property holds with $c=128$.

(c) Consider the family of $N'$ pairs $(e_0,\ \cos\theta\,e_0+\sin\theta\,x_b)$ ($1\le b\le N'$) at angle $\theta\le\arctan(1/10)$. Suppose $r<s/(136\,\eps)$ and $D\in\mathcal E_{\rm hit}$. For pair $b^*$, Proposition~\ref{prop:lb-bias}(iv) and \eqref{eq:lb-fq} give downward bias $t:=\theta-\E[\hat\theta\mid D]\ge\theta\,f_q(s-1)/(17\,r)\ge\theta\,s/(68\,r)>2\eps\theta$. Conditioned on $D$, $\ell\hat\theta/\pi\sim\mathrm{Binomial}(\ell,p)$ with $\pi p=\theta-t\le\theta$, so for $\ell\ge\pi/(\eps^2\theta)$, we have $\Var(\hat\theta\mid D)\le\pi^2p/\ell\le\pi\theta/\ell\le\eps^2\theta^2<(t/2)^2$. Cantelli's inequality with $\tau=\theta$ then gives $|\hat\theta-\theta|>t/2>\eps\theta$ with conditional probability $>\frac12$, establishing the conditional miss property with $c=136$.

(d) Applying Lemma~\ref{lem:lb-hit} with $s_{\max}=n/2$ to the family $x_1,\dots,x_{N'}$, with probability exceeding $\delta$ over $D$ at least one block $x_b$ is hit, giving $w_{x_b}=k_{x_b}=s>\frac12\min\{n/2,\Lambda/\log2\}\ge\frac12\min\{\Lambda,n/2\}$ by Lemma~\ref{lem:lb-walsh}(ii) and \eqref{eq:lb-maxk}.
\end{proof}

\paragraph{Asymptotic exactness of the BG pair certificate on self-pairs.}
On a hit self-pair $(x_b,x_b)$, the true BG bias is $-\Delta_s=-(1+o(1))\,V_s/8$ as $s/r\to0$ (Proposition~\ref{prop:lb-bias}(i) and~(ii)). Notably, the coordinatewise pair certificate \eqref{eq:bg-bias-coarse} recovers this exact asymptotic constant: since $\rho=1$, $V_u=V_s$, and $V_{uv}=0$, \eqref{eq:bg-bias-coarse} simplifies to $V_s/\big(2(1+\sqrt t)^2\big)+p_t/2$; choosing the threshold $t:=1-(s/r)^{1/3}\to 1$ makes the first term converge to $V_s/8$ while Lemma~\ref{lem:sampling}(ii) bounds the second term by $\frac12e^{-(r/s)^{1/3}/2}=o(V_s)$. By contrast, the closed-form cap bound $1.15\,w_{x_b}/r=1.15\,s/r$ of Proposition~\ref{prop:bg-support}(i) (valid for $s\le r/64$, where $\Delta_s\ge\frac{127}{128}\,s/(32r)$) overestimates the self-pair bias by a factor between $\approx 9$ (for small $q$ and $s/r$ under Bernoulli supports, where $\Delta_s\approx s/(8r)$) and $\approx 37$.

\section{Experimental details}\label{app:experiments}

This appendix collects the supplementary tables referenced in Sections~\ref{sec:practice}, \ref{sec:certificates}, and~\ref{sec:experiments}. Full experimental protocols---including random seeds, sample sizes, control variates, pair-certificate optimization details, term-by-term script outputs for Table~\ref{tab:budgets}, and the automated verification scripts (\texttt{check\_A1-tools.py}--\texttt{check\_A6-lower.py}) checking every numerical constant in Appendix~\ref{app:proofs}---are documented in \texttt{code/README.md} in the supplementary material.

\subsection{Benchmark parameters and two-stage cost comparison}\label{app:k4}

Tables~\ref{tab:k4-params}--\ref{tab:composite} document the retrieval benchmark of Section~\ref{sec:practice} and Table~\ref{tab:budgets} ($n=65{,}536$, $m=N=10^9$, $\eps=\delta=0.1$). Table~\ref{tab:k4-params} reports the empirical flatness parameters of Gaussian-like unit vectors and planes used in the certified columns of Table~\ref{tab:budgets} (sample medians across $3000$ simulated realizations for a typical pair, and $\chi^2$ extreme-value extrapolations across $10^9$ vectors for the database maximum). Fixing the sketch length $\ell$ at its worst-case corollary value, Table~\ref{tab:budget-split} shows how the error budget $\eps=0.1$ splits between sampling noise and support bias on a typical pair under balanced supports ($L=\log(2N/\delta)=23.7$). Notice that for BG with the dot-product decoder, the cap correction $18.5\,\wmax/r$ in Theorem~\ref{thm:bg-param}(a) consumes half of the bias budget available to Catoni's query-independent decoder, nearly doubling its certified sparsity ($r=758$ vs.\ $398$). For SimHash (BB), the certified sparsity $r=842$ is governed almost entirely by the matrix Chernoff cap-tail threshold ($r=104$ suffices for the leading fourth-moment bias term alone, and $r\approx 300$ suffices under the matrix Bernstein pair certificate of Lemma~\ref{lem:bb-tail}).

\begin{table}[htbp]\centering\small
\begin{tabular}{@{}lcccc@{}}
\toprule
 & $\wmax$ & $\kmax$ & $\wU$ & $\kU$ \\
\midrule
typical pair (median) & $20.7$ & $3.01$ & $22.9$ & $8.0$ \\
database maximum (extrapolated) & $59$ & $3.4$ & $64$ & $8.2$ \\
\bottomrule
\end{tabular}
\caption{Empirical flatness parameters of Gaussian-like unit vectors and planes in dimension $n=65{,}536$ used to evaluate the certified columns of Table~\ref{tab:budgets} ($m=N=10^9$).}
\label{tab:k4-params}
\end{table}

\begin{table}[htbp]\centering\small
\begin{tabular}{@{}lrccr@{}}
\toprule
row of Table~\ref{tab:budgets} & $\ell$ & noise term & left for the bias & certified $r$ \\
\midrule
BG, dot & $9{,}765$ & $0.0941$ & $0.0059$ & $758$ \\
BG, Catoni & $9{,}570$ & $0.0885$ & $0.0115$ & $398$ \\
BB, popcount & $48{,}188$ & $0.0493$ & $0.0507$ & $842$ \\
\bottomrule
\end{tabular}
\caption{Partition of the target error budget $\eps=0.1$ between sampling noise and support bias for the certified typical-pair column of Table~\ref{tab:budgets} under balanced supports ($n=65{,}536$, $N=10^9$, $\delta=0.1$).}
\label{tab:budget-split}
\end{table}

\begin{table}[htbp]\centering\small
\begin{tabular}{@{}lrrrrrr@{}}
\toprule
 & \multicolumn{3}{c}{composition} & & \multicolumn{2}{c}{signed sparse rows, $\ell r$} \\
\cmidrule(lr){2-4}\cmidrule(l){6-7}
estimator & $n'$ & $\ell_2$ & $n'\ell_2$ & dense $n\ell$ & worst case & typical pair \\
\midrule
BG, dot & 17{,}251 & 30{,}678 & $5.3\cdot10^{8}$ & $4.9\cdot10^{8}$ & $1.1\cdot10^{8}$ & $7.4\cdot10^{6}$ \\
BB, popcount & 17{,}266 & 48{,}188 & $8.3\cdot10^{8}$ & $7.7\cdot10^{8}$ & $4.9\cdot10^{8}$ & $4.1\cdot10^{7}$ \\
\bottomrule
\end{tabular}
\caption{Two-stage dimension reduction followed by dense signing requires slightly more multiply--adds per vector than a single-stage dense projection, whereas signing sparse FJLT rows directly is substantially faster. Operation counts are evaluated in the benchmark regime of Table~\ref{tab:budgets} ($n=65{,}536$, $N=10^9$, $\eps=\delta=0.1$), omitting the shared $n\log_2n\approx10^6$ Hadamard additions. Here ``composition'' grants the first stage a zero-cost, exact Gaussian projection to dimension $n'$ followed by $\ell_2$ dense signed rows, while the remaining columns use the dense $\ell$ and balanced FJLT $(\ell,r)$ from Table~\ref{tab:budgets}.}
\label{tab:composite}
\end{table}

Table~\ref{tab:composite} reports the multiply--add counts underlying the two-stage comparison in Section~\ref{sec:certificates}. Even when the first stage is granted a zero-cost exact Gaussian projection to dimension $n'$ (with the error budget split optimally at roughly $\eps/2$ per stage), preserving $N=10^9$ pairwise cosines to within $\eps/2$ requires $n'\approx 17{,}250$ and $\ell_2\approx 4\ell_{\rm dense}$ second-stage rows, costing $8\%$ more multiply--adds than a single-stage dense projection in $\R^n$. Directly signing sparse FJLT rows reduces this projection cost by $1.7\times$ to $4.8\times$ in the worst case and by $20\times$ to $71\times$ on a typical pair.

\subsection{End-to-end error on the flat one-hot pair}\label{app:e2e}

\begin{table}[htbp]\centering\small
\begin{tabular}{@{}lllrrrrr@{}}
\toprule
 & & & \multicolumn{5}{c}{all entries $\times10^{-3}$} \\
\cmidrule(l){4-8}
 & & & & \multicolumn{4}{c}{RMSE} \\
\cmidrule(l){5-8}
input & estimator & supports & bias & $\ell=2^{9}$ & $2^{11}$ & $2^{13}$ & $2^{15}$ \\
\midrule
one-hot & GG & Bernoulli & $-1.0$ & $62.0$ & $30.4$ & $15.6$ & $8.9$ \\
 &  & balanced & $-0.6$ & $58.7$ & $28.9$ & $14.0$ & $6.9$ \\
\addlinespace[2pt]
 & BG & Bernoulli & $-4.1$ & $39.5$ & $19.8$ & $10.9$ & $6.9$ \\
 &  & balanced & $+0.1$ & $37.8$ & $18.9$ & $9.3$ & $5.0$ \\
\addlinespace[2pt]
 & BB & Bernoulli & $+1.6$ & $21.3$ & $10.7$ & $5.7$ & $3.0$ \\
 &  & balanced & $+1.8$ & $20.4$ & $10.2$ & $5.1$ & $3.0$ \\
\midrule
dense rate $\sigma/\sqrt\ell$ & GG & & & $59.5$ & $29.7$ & $14.9$ & $7.4$ \\
 & BG & & & $38.5$ & $19.3$ & $9.6$ & $4.8$ \\
 & BB & & & $21.2$ & $10.6$ & $5.3$ & $2.7$ \\
\bottomrule
\end{tabular}
\caption{On the perfectly flat one-hot pair ($k_u=w_u=1$), balanced supports eliminate the BG support bias completely, whereas the BB support bias persists under both support laws. Entries report end-to-end empirical bias and RMSE on the one-hot pair of Table~\ref{tab:bias-grid} ($n=4096$, $\rho=0.9$, $r=32$), with BB evaluated on $\hat\rho_{\rm BB}=\cos\hat\theta$.}
\label{tab:e2e-rmse}
\end{table}

Table~\ref{tab:e2e-rmse} repeats the end-to-end sketch-length sweep of Figure~\ref{fig:rmse-panels} on the perfectly flat one-hot pair ($k_u=w_u=1$) discussed in Sections~\ref{sec:exp-sparse} and~\ref{sec:exp-balance}. Under balanced supports, the BG bias vanishes identically (Proposition~\ref{prop:bg-support}(iii)), allowing the BG RMSE to follow the dense curve $\sigma/\sqrt\ell$ all the way to $\ell=2^{15}$, whereas under Bernoulli supports BG saturates at a bias floor of $-4.1\cdot10^{-3}$. By contrast, the BB bias ($+1.6\cdot10^{-3}$ for Bernoulli and $+1.8\cdot10^{-3}$ for balanced supports, five to six standard errors above zero) persists under both support laws: even though a balanced row captures the individual energies $\sigma_1^2=\sigma_2^2=1$ without error, the off-diagonal inner product $\sigma_{12}$ still fluctuates across supports because the plane fourth moment is $\kU=4$, giving a nonzero metric variance $V_M=f_q(\kU-2\bal)/r\ge 2f_q/r$.

\end{document}